\documentclass[10pt,theapa]{article}
\usepackage{graphicx}
\usepackage{covington}
\usepackage{hyperref}
\usepackage{theapa}
\usepackage{tabularx}
\usepackage{subcaption}
\usepackage{multirow}
\usepackage{amsmath}
\usepackage{url}
\usepackage{smartdiagram}
\usesmartdiagramlibrary{additions}
\usepackage{tikz}
\usepackage{tikz-qtree}
\usepackage{afterpage}

\renewcommand{\cite}{\shortcite}
\renewcommand{\citeA}{\shortciteA}

\begin{document}

\title{Survey of the State of the Art in Natural Language Generation: Core tasks, applications and evaluation\thanks{This article appears in {\em Journal of Artificial Intelligence Research} (JAIR), volume 61, pp. 65-170}}

\author{Albert Gatt\\
       Institute of Linguistics and Language Technology (iLLT)\\ 
       University of Malta,\\
       Tal-Qroqq, Msida MSD2080, Malta\\
       {\tt albert.gatt@um.edu.mt}
       \and
       Emiel Krahmer\\
       Tilburg center for Cognition and Communication (TiCC), \\
       Tilburg University, P.O.Box 90153,  NL-5000 LE,\\ Tilburg, The Netherlands\\
       {\tt e.j.krahmer@tilburguniversity.edu}}

\maketitle

\begin{abstract}
This paper surveys the current state of the art in Natural Language Generation ({\sc nlg}), defined as the task of generating text or speech from non-linguistic input. A survey of {\sc nlg} is timely in view of the changes that the field has undergone over the past two decades, especially in relation to new (usually data-driven) methods, as well as new applications of {\sc nlg} technology. This survey therefore aims to (a) give an up-to-date synthesis of research on the core tasks in {\sc nlg} and the architectures adopted in which such tasks are organised; (b) highlight a number of recent research topics that have arisen partly as a result of growing synergies between {\sc nlg} and other areas of artificial intelligence; (c) draw attention to the challenges in {\sc nlg} evaluation, relating them to similar challenges faced in other areas of {\sc nlp}, with an emphasis on different evaluation methods and the relationships between them.
\end{abstract}

\newpage
\tableofcontents
\newpage

\section{Introduction}\label{sec:intro}

In his intriguing story The Library of Babel ({\it La biblioteca de Babel}\/, 1941), Jorge Luis Borges describes a library in which every conceivable book can be found. It is probably the wrong question to ask, but readers cannot help wondering: who wrote all these books? Surely, this could not be the work of human authors? The emergence of automatic text generation techniques in recent years provides an interesting twist to this question. Consider Philip M. Parker, who offered more than 100.000 books for sale via Amazon.com, including for example his {\it The 2007-2012 Outlook for Tufted Washable Scatter Rugs, Bathmats, and Sets That Measure 6-Feet by 9-Feet or Smaller in India}\/. Obviously, Parker did not write these 100,000 books by hand. Rather, he used a computer program that collects publicly available information, possibly packaged in human-written texts, and compiles these into a book. Just like the library of Babel contains many books that are unlikely to appeal to a broad audience, Parker's books need not find many readers. In fact, even if only a small percentage of his books get sold a few times, this would still make him a sizeable profit.

Parker's algorithm can be seen to belong to a research tradition of so-called {\bf text-to-text generation} methods, applications that take existing texts as their input, and automatically produce a new, coherent text as output. Other example applications that generate new texts from existing (usually human-written) text include:

\begin{itemize}
\item  machine translation, from one language to another \cite<e.g.,>{hutchins1992,och2003};
\item  fusion and summarization of related sentences or texts to make them more concise \cite<e.g.,>{Clarke2010};
\item  simplification of complex texts, for example to make them more accessible for low-literacy readers \cite<e.g.,>{Siddharthan2014} or for children \cite{Macdonald2016};
\item automatic spelling, grammar and text correction \cite<e.g.,>{kukich1992techniques,dale2012hoo};
\item automatic generation of peer reviews for scientific papers \cite{bartoli2016your};
\item  generation of paraphrases of input sentences
\cite<e.g.,>{bannard2005paraphrasing,kauchak2006paraphrasing}; and
\item  automatic generation of questions, for educational and other purposes \cite<e.g.,>{brown2005automatic,rus2010overview}.
\end{itemize}

Often, however, it is necessary to generate texts which are not grounded in existing ones. Consider, as a case in point, the minor earthquake that took place close to Beverly Hills, California on March 17, 2014. The Los Angeles Times was the first newspaper to report it, within 3 minutes of the event, providing details about the time, location and strength of the quake. This report was automatically generated by a `robo-journalist', which converted the incoming automatically registered earthquake data into a text, by filling gaps in a predefined template text \cite{Slate2014}.

Robo-journalism and associated practices, such as data journalism, are examples of what is usually referred to as {\bf data-to-text generation}.  They have had a considerable impact in the fields of journalism and media studies \cite{VanDalen2012,Clerwall2014,Hermida2015}. The technique used by the Los Angeles Times was not new; many applications have been developed over the years which automatically generate text from non-linguistic data including, but not limited to, systems which produce:

\begin{itemize}
\item soccer reports \cite<e.g.,>{Theune2001,Chen2008};
\item virtual `newspapers' from sensor data \cite{Molina2011} and news reports on current affairs \cite{Lepp2017};
\item text addressing environmental concerns, such as wildlife tracking \cite{Siddharthan2012,Ponnamperuma2013}, personalised environmental information \cite{Wanner2015}, and enhancing engagement of citizen scientists via generated feedback \cite{VanderWal2016};
\item weather and financial reports \cite{Goldberg1994,Reiter2005,Turner2008a,Ramos-Soto2015,Plachouras2016}; 
\item summaries of patient information in clinical contexts \cite{Huske-Kraus2003,Harris2008,Portet2009,Gatt2009,Banaee2013}; 
\item interactive information about cultural artefacts, for example in a museum context \cite<e.g.,>{ODonnell2001,Stock2007}; and
\item text intended to persuade \cite{Carenini2006}, or motivate behaviour modification \cite{Reiter2003}.
\end{itemize}

These systems may differ considerably in the quality and variety of the texts they produce, their commercial viability and the sophistication of the underlying methods, but all are examples of data-to-text generation. Many of the  systems mentioned above focus on imparting information to the user. On the other hand, as shown by the examples cited above of systems focussed on persuasion or behaviour change, informing need not be the exclusive goal of {\sc nlg}. Nor is it a trivial goal in itself, since in order to successfully impart information, a system needs to select what to say, distinguishing it from what can be easily inferred (possibly also depending on the target user), before expessing it coherently.

Generated texts need not have a large audience. There is no need to automatically generate a report of, say, the Champions League European football final, which is covered by many of the best journalists in the field anyway. However, there are many other games, less important to the general public (but presumably very important to the parties involved). Typically, all sports statistics (who played?, who scored? etc.) for these games are stored, but such statistics are not as a rule perused by sport-reporters. Companies like Narrative Science\footnote{\url{https://www.narrativescience.com}} fill this niche by automatically generating sport reports for these games. Automated Insights\footnote{\url{https://automatedinsights.com}} even generates reports based on user-provided `fantasy football' data. In a similar vein, the automatic generation of weather forecasts for offshore oil platforms \cite{Sripada2003}, or from sensors monitoring the performance of gas turbines \cite{Yu2006}, has proven to be a fruitful application of data-to-text techniques. Such applications are now the mainstay of companies like Arria-NLG.\footnote{\url{http://www.arria.com}}

Taking this idea one step further, data-to-text generation paves the way for tailoring texts to specific audiences.  For example, data from babies in neonatal care can be converted into text differently, with different levels of technical detail and explanatory language, depending on whether the intended reader is a doctor, a nurse or a parent \cite{Mahamood2011}. 
One could also easily imagine that different sport reports are generated for fans of the respective teams; the winning goal of one team is likely to be considered a lucky one from the perspective of the losing team, irrespective of its `objective' qualities \cite{van2017pass}. A human journalist would not dream of writing separate reports about a sports match (if only for lack of time), but for a computer this is not an issue and this is likely to be appreciated by a reader who receives a more personally appropriate report.

\subsection{What is Natural Language Generation?}

Both text-to-text generation and data-to-text generation are instances of {\bf Natural Language Generation} ({\sc nlg}). In the most widely-cited survey of {\sc nlg} methods to date \cite{Reiter1997,Reiter2000}, {\sc nlg} is characterized as `the subfield of artificial intelligence and computational linguistics that is concerned with the construction of computer systems than can produce understandable texts in English or other human languages from some underlying non-linguistic representation of information' \cite[p.1]{Reiter1997}. Clearly this definition fits data-to-text generation better than text-to-text generation, and indeed \citeA{Reiter2000} focus exclusively on the former, helpfully and clearly describing the rule-based approaches that dominated the field at the time. 

It has been pointed out that precisely defining {\sc nlg} is rather difficult \cite<e.g.,>{evans2002nlg}: everybody seems to agree on what the output of an {\sc nlg} system should be (text), but what the exact input is can vary substantially \cite{McDonald1993}. Examples include flat semantic representations, numerical data and structured knowledge bases. More recently, generation from visual input such as image or video has become an important challenge \cite<e.g.,>[among many others]{Mitchell2012,Kulkarni2013,Thomason2014}.

A further complication is that the boundaries between different approaches are themselves blurred. For example, text summarisation was characterized above as a text-to-text application. However, many approaches to text-to-text generation (especially abstractive summarisation systems, which do not extract content wholesale from the input documents) use techniques which are also used in data-to-text, as when opinions are extracted from reviews and expressed in completely new sentences \cite<e.g.,>{labbe2012towards}. Conversely, a data-to-text generation system could conceivably rely on text-to-text generation techniques for learning how to express pieces of data in different or creative ways \cite{Mcintyre2009,Gatt2009,Kondadadi2013}.

Considering other applications of {\sc nlg} similarly highlights how blurred boundaries can get. For example, the generation of spoken utterances in dialogue systems \cite<e.g.,>{Walker2007,Rieser2009,Dethlefs2014} is another application of {\sc nlg}, but typically it is closely related to dialogue management, so that management and realisation policies are sometimes learned in tandem \cite<e.g.,>{Rieser2011}.  

The position taken in this survey is that what distinguishes data-to-text generation is ultimately its input. Although this varies considerably, it is precisely the fact that such input is not -- or isn't exclusively -- linguistic that is the main challenge faced by most of the systems and approaches we will consider. In what follows, unless otherwise specified in context, the terms `Natural Language Generation' and `{\sc nlg}' will be used to refer to systems that generate text from non-linguistic data.

\subsection{Why a Survey on Natural Language Generation?}

Arguably \citeA{Reiter2000} is  still the most complete available survey of {\sc nlg}. However, the field of {\sc nlg} has changed drastically in the last 15 years, with the emergence of successful applications generating tailored reports for specific audiences, and with the emergence of text-to-text as well as vision-to-text generation applications, which also tend to rely more on statistical methods than traditional data-to-text. None of these are covered by \citeA{Reiter2000}. Also notably absent are discussions of applications that move beyond standard, `factual' text generation, such as those that account for personality and affect, or creative text such as metaphors and narratives. Finally, a striking omission by \citeA{Reiter2000} is the lack of discussion of evaluation methodology. Indeed, evaluation of {\sc nlg} output has only recently started to receive systematic attention, in part due to a number of shared tasks that were conducted within the {\sc nlg} community. 

Since \citeA{Reiter2000} published their book, various other {\sc nlg} overview texts have also appeared. \citeA{Bateman2005} cover the cognitive, social and computational dimensions of {\sc nlg}. \citeA{McDonald2010} offers a general characterization of {\sc nlg} as `the process by which thought is rendered into language' (p. 121). \citeA{Wanner2010} zooms in on automatic generation of reports, while \citeA{DiEugenio2010} look at specific applications, especially in education and health-care. Various specialized collections of articles have also been published, including that by \citeA{Krahmer2010a}, which targets data-driven approaches; and by \citeA{Bangalore2014}, which focusses on interactive systems. The web offers various unpublished technical reports, such as the survey by \citeA{Theune2003} on dialogue systems; the reports by \citeA{Piwek2003} and \citeA{Belz2003} on affective {\sc nlg}; and the survey by \citeA{Gkatzia2016} on content selection. While useful, these resources do not discuss recent developments or offer a comprehensive review. This indicates that a new state-of-the-art survey is highly timely. 

\subsection{Goals of this Survey}

The goal of the current paper is to present a comprehensive overview of {\sc nlg} developments since 2000, both in order to provide {\sc nlg} researchers with a synthesis and pointers to relevant research, and to introduce the field to researchers who are less familiar with {\sc nlg}.
Though {\sc nlg} has been a  part of {\sc ai} and {\sc nlp}  from the early days \cite<see e.g.,>{Winograd1972,Appelt1985}, as a field it has arguably not been fully embraced by these broader communities, and has only recently began to take full advantage of  recent advances in data-driven, machine learning and deep learning approaches.

Following \citeA{Reiter2000}, our main focus, especially in the first part of the survey, will be on data-to-text generation. In any case, doing full justice to recent developments in the various text-to-text generation applications is beyond the scope of a single survey, and many of these are covered in other surveys, including those by \citeA{Mani2001} and \citeA{Nenkova2011} for summarisation; \citeA{androutsopoulos2010survey} on paraphrasing; and \citeA{piwek2012varieties} on automatic question generation. However, we will in various places discuss connections between data-to-text and text-to-text generation, both because -- as noted above -- the boundaries are blurred, but  also, and perhaps more importantly, because 
text-to-text systems have long been couched in the data-driven frameworks that are becoming increasingly popular in data-to-text generation, also giving rise to some hybrid systems that combine rule-bused and statistical techniques \cite<e.g.,>{Kondadadi2013}.

Our review will start with an updated overview of the core {\sc nlg} tasks that were introduced by \citeA{Reiter2000}, followed by a discussion of architectures and approaches, where we pay special attention to those not covered in the \citeA{Reiter2000} survey. These two sections constitute the `foundational' part of the survey. Beyond these, we highlight several new developments, including  approaches where the input data is visual; and research aimed at generating more varied, engaging or creative and entertaining texts, taking {\sc nlg} beyond the factual, repetitive texts it is sometimes accused of producing. We believe that these applications are not only interesting in themselves, but may also inform more `utility'-driven text generation application. For example, by including insights from narrative generation we may be able to generate more engaging reports and by including insights from metaphor generation we may be able to phrase information in these reports in a more original manner. Finally, we will discuss recent developments in evaluation of natural language generation applications.

In short, the goals of this survey are:

\begin{itemize}
\item To give an up-to-date synthesis of research on the core tasks in {\sc nlg}, as well as the architectures adopted in the field, especially in view of recent developments exploiting data-driven techniques (Sections \ref{sec:tasks} and \ref{sec:architectures});
\item To highlight a number of relatively recent research issues that have arisen partly as a result of growing synergies between {\sc nlg} and other areas of artificial intelligence, such as computer vision, stylistics and computational creativity (Sections \ref{sec:image}, \ref{sec:style} and \ref{sec:creativity});
\item To draw attention to the challenges in {\sc nlg} evaluation, relating them to similar challenges faced in other areas of {\sc nlp}, with an emphasis on different evaluation methods and the relationships between them (Section \ref{sec:evaluation}).
\end{itemize}

\section{NLG Tasks}\label{sec:tasks}

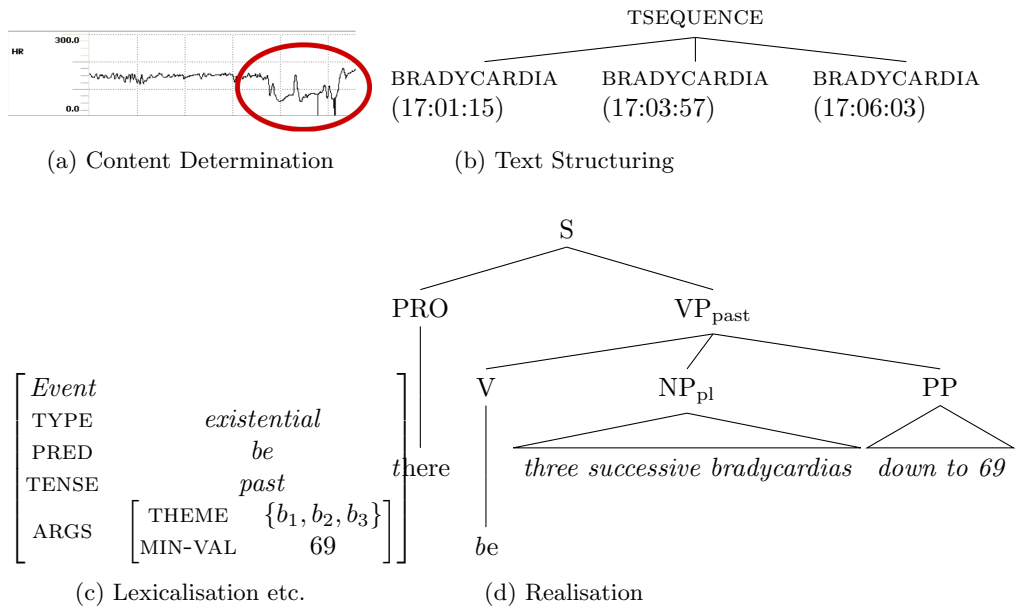
\begin{figure}[!t]
	\makebox[\linewidth][l] {%
	\begin{subfigure}[b]{0.4\textwidth}
        \includegraphics[width=\textwidth]{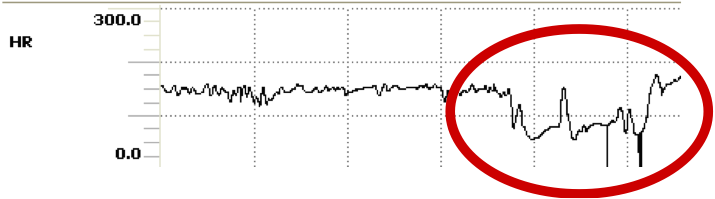}
        \caption{Content Determination }
        \label{fig:cd}
    \end{subfigure}
    \begin{subfigure}[b]{0.4\textwidth}
    	\begin{minipage}{\textwidth}
        	\begin{tikzpicture}
            	\Tree 
            	[.{\sc tsequence} 
                	[.\parbox{2.5cm}{{\sc bradycardia} (17:01:15)} ]
					[.\parbox{2.5cm}{{\sc bradycardia} (17:03:57)} ]
                    [.\parbox{2.5cm}{{\sc bradycardia} (17:06:03)} ]
				]
            \end{tikzpicture}
        \end{minipage}
        \caption{Text Structuring }
        \label{fig:ts}
    \end{subfigure}
    }\\\\
    \makebox[\linewidth][l] {%
    \begin{subfigure}[b]{0.4\textwidth}
    	\begin{minipage}{\textwidth}    	
    	$\begin{bmatrix}
    	\textit{Event} & \\
    	\textsc{type} & \textit{existential}\\
    	\textsc{pred} & \textit{be}\\
    	\textsc{tense} & \textit{past}\\
    	\textsc{args} & \begin{bmatrix}\textsc{theme} & \{b_1,b_2,b_3\} \\ \textsc{min-val} & 69\end{bmatrix}\\ 
		\end{bmatrix}$
		\end{minipage}
        \caption{Lexicalisation etc.}
        \label{fig:agg}
    \end{subfigure}
    \begin{subfigure}[b]{0.4\textwidth}
        \begin{minipage}{\textwidth}    
			\begin{tikzpicture}
            \Tree
            	[.S 
                	[.PRO [{\textit there} ]]
                    [.VP$_{\text{past}}$ 
                    	[.V [{\textit be} ] ]
                        [.NP$_{\text{pl}}$ \edge[roof]; {\em three successive bradycardias} ]
                        [.PP \edge[roof]; {\em down to 69} ] ]
                    ]
            \end{tikzpicture}
        \end{minipage}        
        \caption{Realisation}
        \label{fig:lr}
    \end{subfigure}
    }
\caption{Tasks in {\sc nlg}, illustrated with a simplified example from the neonatal intensive care domain. First the system has to decide what the important events are in the data (a, content determination), in this case, occurrences of low heart rate (bradycardias). Then it has to decide in which order it wants to present data to the reader (b, text structuring) and how to express these in individual sentence plans (c, aggregation, lexicalisation, reference). Finally, the resulting sentences are generated (d, linguistic realisation).}
\label{fig:tasks-example}
\end{figure}

Traditionally, the {\sc nlg} problem of converting input data into output text was addressed by splitting it up into a number of subproblems. The following six are frequently found in many {\sc nlg} systems  \cite{Reiter1997,Reiter2000}; their role is illustrated in Figure \ref{fig:tasks-example}:

\begin{enumerate}
\item {\it Content determination}\/: Deciding which information to include in the text under construction,
\item {\it Text structuring}\/: Determining in which order information will be presented in the text,
\item {\it Sentence aggregation}\/: Deciding which information to present in individual sentences,
\item {\it Lexicalisation}\/: Finding the right words and phrases to express  information,
\item {\it Referring expression generation}\/: Selecting the words and phrases to identify domain objects,
\item {\it Linguistic realisation}\/: Combining all words and phrases into well-formed sentences.
\end{enumerate}

These tasks could be thought of in terms of `early' decision processes (which information to convey to the reader?) to `late' ones (which words to use in a particular sentence, and how to put them in their correct order?). Here, we refer to `early' and `late' tasks by way of distinguishing between choices that are more oriented towards the data (such as what to say) and choices that are of an increasingly linguistic nature (e.g., lexicalisation, or realisation). This characterization reflects a long-running distinction in {\sc nlg} between {\em strategy} and {\em tactics} \cite<a distinction that goes back at least to>{Thompson1977}. This distinction also suggests a temporal order in which the tasks are executed, at least in systems with a modular, pipeline architecture (discussed in Section \ref{sec:modular}): for example, the system first needs to decide which input data to express in the text, before it can order information for presentation. However, such ordering of modules is nowadays increasingly put into question in the data-driven architectures discussed below (Section  \ref{sec:architectures}). 

In this section, we briefly describe these six tasks, illustrating them with examples, and highlight recent developments in each case. As we shall see, while the `early' tasks are crucial for the development of {\sc nlg} systems, they are often intimately connected to the specific application. By contrast, `late' tasks are more often investigated independently of an application, and hence have resulted in approaches that can be shared between applications. 

\subsection{Content Determination}\label{sec:content-det}
As a first step in the generation process, the {\sc nlg} system needs to decide which information should be included in the text under construction, and which should not. Typically, more information is contained in data than we want to convey through text, or the data is more detailed than we care to express in text. This is clear in Figure \ref{fig:cd}, where the input signal -- a patient's heart rate -- only contains a few patterns of interest. Selection may also depend on the target audience (e.g. does it consist of {\it experts}\/ or {\it novices}\/, for example) and on the overall communicative intention (e.g. should the text {\it inform}\/ the reader or {\it convince} him to do something).

Content determination involves choice. In a soccer report, we may not want to verbalise each pass and foul committed, even though the data may contain this information. In the case of neonatal care, data might be collected continuously from sensors measuring heart rate, blood pressure and other physiological parameters. Data thus needs to be filtered and abstracted into a set of {\em preverbal messages}, semantic representations of information which are often expressed in a formal representation language, such as logical or database languages, attribute-value matrices or graph structures. They can express, among other things, which relations hold between which domain entities, for example, expressing that player X scored the first goal for team Y at time T.

Though content determination is present in most {\sc nlg} systems \cite<cf.>{Mellish2006}, approaches are typically closely related to the domain of application. A notable exception is the work of \citeA{Guhe2007}, which offers a cognitively plausible, incremental account of content determination based on studies of speakers' descriptions of dynamic events as they unfold. This work belongs to a strand of research which considers {\sc nlg} first and foremost as a methodology eminently suitable for understanding {\it human}\/ language production.

In recent years, researchers have started exploring data-driven techniques for content determination \cite<see e.g.,>{Barzilay2004,BouayadAgha2013,Kutlak2013,Venigalla2013}. \citeA{Barzilay2004}, for example, used Hidden Markov Models to model topic shifts in a particular domain of discourse (say, earthquake reports), where the hidden states represented `topics', modelled as sentences clustered together by similarity.  A clustering approach was also used by \citeA{Duboue2003} in the biography domain, using texts paired with a knowledge base, from which semantic data was clustered and scored according to its occurrence in text. In a similar vein \citeA{Barzilay2005} use a database of American football records and corresponding text. Their aim was not only to identify bits of information that should be mentioned, but also dependencies between them, since mentioning a certain event (say, a score by a quarterback) may warrant the mention of another (say, another scoring event by a second quarterback). The solution proposed by \citeauthor{Barzilay2005} was to compute both individual preference scores for events, and a link preference score. 

More recently, various researchers have addressed the question of how to automatically learn alignments between data and text, also in the broader context of grounded language acquisition, i.e., modelling how we learn language by looking at correspondences between objects and events in the world and the way we refer to them in language \cite{Roy2002,yu2004multimodal,yu2013grounded}. For example, \citeA{Liang2009}  extended the work by \citeA{Barzilay2005} to multiple domains (soccer and weather), relying on weakly supervised techniques; in a similar vein, \citeA{koncel2014multi} presented a weakly supervised multilevel approach, to deal with the fact that there is no one-to-one correspondence between, for example, soccer events in data and sentences in associated soccer reports. We shall return to these methods as part of a broader discussion of data-driven approaches below (Section \ref{sec:datadriven}).

\subsection{Text Structuring}\label{sec:text-struct}
Having determined what messages to convey, the {\sc nlg} system needs to decide on their order of presentation to the reader. For example, Figure \ref{fig:ts} shows three events of the same type (all bradycardia events, that is, brief drops in heart rate), selected (after abstraction) from the input signal and ordered as a temporal sequence.

This stage is often referred to as text (or discourse or document) structuring. In the case of the soccer domain, for example, it seems reasonable to start with general information (where and when the game was played, how many people attended, etc.), before the goals are described, typically in temporal order. In the neonatal care domain, a temporal order can be imposed among specific events, as in Figure \ref{fig:ts}, but larger spans of text may reflect ordering based on importance, and grouping of information based on relatedness (e.g. all events related to a patient's respiration) \cite{Portet2009}. Naturally, alternative discourse relations may exist between separate messages, such as {\it contrasts}\/ or {\it elaborations}\/. The result of this stage is a discourse, text or document plan, which is a structured and ordered representation of messages. 

These examples again imply that the application domain imposes constraints on ordering preferences. Early approaches, such as \citeA{McKeown1985}, often relied on hand-crafted, domain-dependent structuring rules (which McKeown called {\it schemata}\/).  To account for discourse relations between messages, researchers have alternatively relied on Rhetorical Structure Theory \cite<{\sc rst}; e.g.,>{Mann1988,Scott1990,Hovy1993}, which also typically involved domain-specific rules. For example, \citeA{Williams2008} used {\sc rst} relations to identify ordering among messages that would maximise clarity to low-skilled readers.

Various researchers have explored the possibilities of using machine learning techniques for document structuring \cite<e.g.,>{Dimitromanolaki2003,althaus2004}, sometimes doing this in tandem with content selection \cite{Duboue2003}. General approaches for information ordering \cite{Barzilay2004,Lapata2006} have been proposed, which automatically try to find an optimal ordering of `information-bearing items'. These approaches can be applied to text structuring, where the items to be ordered are typically preverbal messages;  however, they can also be applied in (multidocument) summarisation, where the items to be ordered are sentences from the input documents which are judged to be summary-worthy enough to include  \cite<e.g.,>{Barzilay2002,Bollegala2010}. 

\subsection{Sentence Aggregation}\label{sec:aggregation}

Not every message in the text plan needs to be expressed in a separate sentence; by combining multiple messages into a single sentence, the generated text becomes potentially more fluid and readable \cite<e.g.,>{Dalianis1999,Cheng2000}, although there are also situations where it has been argued that aggregation should be avoided (discussed in Section \ref{sec:affect}). For instance, the three events selected in Figure \ref{fig:ts} are shown as `merged' into a single pre-linguistic representation, which will be mapped to a single sentence. The process by which related messages are grouped together in sentences is known as sentence aggregation.

To take another example, from the soccer domain, one (unaggregated) way to describe the fastest hat-trick in the English Premier League would be:

\begin{examples}\label{ex:aggreg}
\item Sadio Mane scored for Southampton after 12 minutes and 22 seconds.\\
\item Sadio Mane scored for Southampton after 13 minutes and 46 seconds.\\
\item Sadio Mane scored for Southampton after 15 minutes and 18 seconds.
\end{examples}

Clearly, this is rather redundant, not very concise or coherent, and generally unpleasant to read. An aggregated alternative, such as the following, would therefore be preferred:

\begin{example}\label{ex:aggreg2}
Sadio Mane scored three times for Southampton in less than three minutes.
\end{example}

In general, aggregation is difficult to define, and has been interpreted in various ways, ranging from redundancy elimination to linguistic structure combination. \citeA{Reape1999} offer an early survey, distinguishing between aggregation at the semantic level (as illustrated in Figure \ref{fig:agg}) and at the level of syntax, illustrated in the transition from (1-3) to (4) above.

It is probably fair to say that much early work on aggregation was strongly domain-dependent. This work focussed on domain- and application-specific rules (e.g. `if a player scores two consecutive goals, describe these in the same sentence'), that were typically hand-crafted  \cite<e.g.,>{Hovy1988,Dalianis1999,Shaw1998}. Once again, more recent work is gradually moving towards data-driven approaches, where aggregation rules are acquired from corpus data \cite<e.g.,>{Walker2001,Cheng2000}.  \citeA{Barzilay2006} present a system that learns how to aggregate on the basis of a parellel corpus of sentences and  corresponding database entries, by looking for similarities between entries. As was the case with the content selection method of \citeA{Barzilay2005}, \citeA{Barzilay2006} view the problem in terms of global optimisation: an initial classification is done over pairs of database entries which determines whether they should be aggregated or not on the basis of their pairwise similarity. Subsequently, a globally optimal set of linked entries is selected based on transitivity constraints (if $\langle e_{i},e_{j} \rangle$ and $\langle e_{j},e_{k} \rangle$ are linked, then so should $\langle e_{i},e_{k} \rangle$) and global constraints, such as how many sentences should be aggregated in a document. Global optimisation is cast in terms of Integer Linear Programming, a well-known mathematical optimization technique \cite<e.g.,>{nemhauser1988integer}.
 
With syntactic aggregation, it is arguably more feasible to define domain-independent rules to eliminate redundancy \cite{Harbusch2009,Kempen2009}. For example, converting (5) into (6) below

\begin{examples}
\item Sadio Mane scored in the 12th minute and he scored again in the 13th minute.
\item Sadio Mane scored in the 12th minute and again in the 13th.
\end{examples}
could be achieved by identifying the parallel verb phrases in the two conjoined sentences and eliding the subject and verb in the second.  Recent work has explored the possibility of acquiring such rules from corpora automatically. For example, \citeA{Stent2009} describe an approach to the acquisition of sentence-combining rules from a discourse treebank, which are then incorporated into the {\sc sp}a{\sc rk}y sentence planner described by \citeA{Walker2007}. A more general approach to the same problem is discussed by \citeA{White2015}. 

Arguably, aggregation on the syntactic level can only account for relatively small reductions, compared to aggregation at the level of messages. Furthermore, syntactic aggregation assumes that the sentence planning process (which includes lexicalisation) is complete. Hence, while 
traditional approaches to {\sc nlg} view aggregation as part of sentence planning, which occurs prior to syntactic realisation, the validity of this claim depends on the type of aggregation being performed \cite<see also>{theune2006}.

\subsection{Lexicalisation}\label{sec:lexicalisation}

Once the content of the sentence has been finalised, possibly also as a result of aggregation at the message level, the system can start converting it into natural language. In our example (Figure \ref{fig:agg}), the outcome of aggregation and lexicalisation are shown together: here, the three events have been grouped, and mapped to a representation that includes a verb ({\em be}) and its arguments, though the arguments themselves still have to be rendered in a referring expression (see below). This reflects an important decision, namely, which words or phrases to use to express the messages' building blocks. A  complication is that often a single event can be expressed in natural language in many different ways. A scoring event in a soccer match, for example, can be expressed as `to score a goal', `to have a goal noted', `to put the ball in the net', among many others.

The complexity of this lexicalisation process critically depends on the number of alternatives that the {\sc nlg} system can entertain. Often, contextual constraints play an important role here as well: if the aim is to generate texts with a certain amount of variation \cite<e.g.,>{Theune2001}, the system can decide to randomly select a lexicalisation option from a set of alternatives (perhaps even from a set of alternatives not used earlier in the text). However, stylistic constraints come into play: `to score a goal' is an unfortunate way of expressing an own goal, for example. In other applications, lexical choice may even be informed by other considerations, such as the attitude or affective stance towards the event in question \cite<e.g.,>[and the discussion in Section \ref{sec:style}]{Fleischman2002}. Whether or not {\sc nlg} systems aim for variation in their output or not depends on the domain. For example, variation in soccer reports is presumably more appreciated by readers than variation in weather reports \cite<on which see>{Reiter2005}; it may also depend on where in a text the variation occurs. For example, variation in expressing timestamps may be less appreciated than variation in referential forms \cite{castro2016towards}.

One straightforward model for lexicalisation -- the one assumed in Figure \ref{fig:tasks-example} -- is to operate on preverbal messages, converting domain concepts directly into lexical items. This might be feasible in well-defined domains. More often, lexicalisation is harder, for at least two reasons \cite<cf.>{Bangalore2000}: First, it can involve selection between semantically similar, near-synonymous or taxonomically related words \cite<e.g.  {\em animal} vs {\em dog};>{Stede1996,Edmonds2002}. Second, it is not always straightforward to model lexicalisation in terms of a crisp concept-to-word mapping. One source of difficulty is vagueness, which arises, for example, with terms denoting properties that are gradable. For example, selecting the adjectives `wide' or `tall' based on the dimensions of an entity requires the system to reason about the width or height of similar objects, perhaps using some standard of comparison \cite<since a `tall glass' is shorter than a `short man';  cf.>{Kennedy2005a,VanDeemter2012}. A similar issue has been noted in the context of presenting numerical information, such as timestamps and quantities \cite{Reiter2005,power2012generating}. For example, \citeA{Reiter2005} discussed time expressions in the context of weather-forecast generation, pointing out that a timestamp {\em 00:00} could be expressed as {\em late evening}, {\em midnight}, or simply {\em evening} \cite[p. 143]{Reiter2005}. Not surprisingly, humans (including the professional forecasters that contributed to \citeauthor{Reiter2005}'s evaluation), show considerable variation in their lexical choices. 

It is interesting to note that many issues related to lexicalisation have also been discussed in the psycholinguistic literature on lexical access 
\cite{Levelt1989,Levelt1999:lexical}. Among these is the question of how speakers home in on the right word and under what conditions they are liable to make errors, given that the mental lexicon is a densely connected network in which lexical items are connected at multiple levels (semantic, phonological, etc). This has also been a fruitful topic for computational modelling \cite<e.g.,>{Levelt1999:lexical}. In contrast to cognitive modelling approaches, however, research in {\sc nlg} increasingly views lexicalisation as part of surface realisation (discussed below) \cite<a similar observation is made by>[p.351]{Mellish1998a}. A fundamental contribution in this context is by \citeA{Elhadad1997}, who describe a unification-based approach, unifying conceptual representations (i.e., preverbal messages) with grammar rules encoding lexical as well as syntactic choices. 

\subsection{Referring Expression Generation}\label{sec:reg}

Referring Expression Generation ({\sc reg})  is characterised by \citeA[p.11]{Reiter1997} as ``the task of selecting words or phrases to identify domain entities". This characterisation suggests a close similarity to lexicalisation, but \citeA{Reiter2000} point out that the essential difference is that referring expression generation is a ``discrimination task, where the system needs to communicate sufficient information to distinguish one domain entity from other domain entities".  {\sc reg} is among the tasks within the field of automated text generation that has received most attention in recent years \cite{Mellish2006,Siddharthan2011}. Since it can be separated relatively easily from a specific application domain and studied in its own right, various `standalone' solutions for the {\sc reg} problem exist. 

In our running example, the three bradycardia events shown in Figure \ref{fig:ts} are later represented as a set of three entities under the {\sc theme} argument of {\em be}, following lexicalisation (Figure \ref{fig:agg}). How the system refers to them will depend, among other things, on whether they've already been mentioned (in which case, a pronoun or definite description might work) and if so, whether they need to be distinguished from any other similar entities (in which case, they might need to be distinguished by some properties, such as the time when they occurred). 

The first choice is therefore related to {\em referential form}: whether entities are referred to using a pronoun, a proper name or an (in)definite description, for example. This depends partly on the extent to which the entity is `in focus' or `salient' \cite<see e.g.,>{Poesio2004} and indeed such notions underlie many computational accounts of pronoun generation \cite<e.g.,>{McCoy1999,Callaway2002,Kibble2004}. Choosing referential forms has recently been the topic of a series of shared tasks on the Generation of Referring Expressions in Context \cite<{\sc grec};>{Belz2010},  using data from Wikipedia articles, which included choices such as reflexive pronouns and proper names. Many systems participating in this challenge framed the problem in terms of classification among these many options. Still, it is probably fair to say that much work on referential form has focussed on when to use pronouns. Forms such as proper names remain understudied, although recently various researchers have highlighted the problems of proper name generation \cite{Siddharthan2011,deemter2016designing,ferreira2017generating}. 

\begin{figure}[t]
\centering
	\begin{subfigure}[b]{0.45\textwidth}
		\includegraphics[width=\textwidth,height=0.2\textheight]{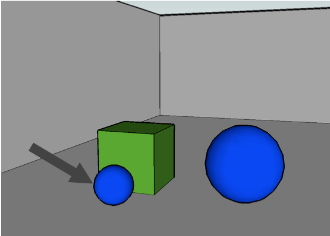}
		\caption{Visual domain from the {\sc gre3d} corpus \cite{Viethen2008}}
		\label{fig:example-domain}
	\end{subfigure}
	\hfill
	\begin{subtable}[b]{0.45\textwidth}
	\begin{tabular}{l | lll}
	\hline
		& \multicolumn{3}{c}{Domain objects}\\
			Attr              & $d_1$              & $d_2$              &  $d_3$\\ \hline
				Color     & blue                 & green               & blue \\
				Shape   & ball                  & cube                 & ball \\
				Size      & small                & large                & large \\
				Rel & bef($d_2$) & beh($d_1$) & nt($d_2$)\\ \hline
	 \end{tabular}
	 \caption{Table of objects and attributes. {\em beh}: `behind'; {\em bef}: `before'; {\em nt}: `next to'}
	 \label{table:example-domain}
	\end{subtable}
	\caption{Visual domain and corresponding tabular representation}
\end{figure}

Determining the {\it referential content}\/ usually comes into play when the chosen form is a  description. Typically, there are multiple entities which have the same referential category or type in a domain (more than one player, for example, or several bradycardias). As a result, other properties of the entity will need to be mentioned if it is to be identified by the reader or hearer. Earlier {\sc reg} research often worked with simple visual domains, such as Figure \ref{fig:example-domain} or its corresponding tabular representation, taken from the {\sc gre3d} corpus \cite{Viethen2008}. In this example, the {\sc reg} content selection problem is to find a set of properties for a target (say $d_1$) that singles it out from its two distractors ($d_2$ and $d_3$). 

{\sc reg} content determination algorithms can be thought of as performing a search through the known properties of the referent for the `right' combination that will distinguish it in context. What constitutes the `right' combination depends on the underlying theory. Too much information in the description (as in {\it the small blue ball before the large green cup}\/) might be misleading or even boring; too little ({\em the ball}) might hinder identification. Much work on {\sc reg} has appealed to the Gricean maxim stating that speakers should make sure that their contributions are sufficiently informative for the purposes of the exchange, but not more so \cite{Grice1975}. How this is interpreted has been the subject of a number of algorithmic interpretations, including:

\begin{itemize}
\item Conducting an exhaustive search through the space of possible descriptions and choosing the smallest set of properties that will identify the target referent, the strategy incorporated by the Full Brevity procedure \cite{Dale1989}. In our example domain, this would select size.
\item Selecting properties incrementally, but choosing the one which rules out most distractors at each step, thereby minimising the possibility of including information that isn't directly relevant to the identification task. This is the underlying idea of the Greedy Heuristic algorithm \cite{Dale1989,Dale1992}, and it has more recently been revived in stochastic utility-based models such as \citeA{Frank2009}. In our example scene, such an algorithm would once again consider size first. 
\item Selecting properties incrementally, but based on domain-specific preference or cognitive salience. This is the strategy incorporated in the Incremental Algorithm \cite{Dale1995}, which would predict that color should be preferred over size in our example. 
\end{itemize}

While these heuristics focus exclusively on the requirement that a referent be unambiguously identified, research on reference in dialogue \cite<e.g.,>{Jordan2005} has shown that under certain conditions, referring expressions may also include `redundant' properties in order to achieve other communicative goals, such as confirmation of a previous utterance by an interlocutor. Similarly, \citeA{White-Clark-Moore:2010} present a system which generates user-tailored descriptions in spoken dialogue, arguing that, for example, a frequent flyer would prefer different descriptions of flights than a student who only flies occasionally. 

These various algorithms compute (possibly different) distinguishing descriptions for target referents (more precisely: they select sets of properties that distinguish the target, but that still need to be expressed in words; see Section \ref{sec:lr} below). Various strands of more recent work can be distinguished \cite<surveyed in>{Krahmer2012}. Some researchers have focussed on extending the expressivity of the `classical' algorithms, to include plurals ({\it the two balls}\/) and relations ({\it the ball in front of a cube}\/)  \cite<e.g.,>[among many others]{Horacek1997,Stone2000,Gardent2002,Kelleher2006,Viethen2008}. Other work has cast the problem in probabilistic terms; for example, \citeA{Fitzgerald2013} frame {\sc reg} as a problem of estimating a log-linear distribution over a space of logical forms representing expressions for sets of objects. 
Other work has concentrated on evaluating the performance of different {\sc reg} algorithms, by collecting controlled human references and comparing these with the references predicted by various algorithms \cite<e.g.,>[again among many others]{Belz2008,Gatt2010,Jordan2005}. In a similar vein, researchers have also started exploring the relevance of {\sc reg} algorithms as psycholinguistic models of human language production \cite<e.g.,>{VanDeemter2012a}. 

A different line of work has moved away from the separation between content selection and form, performing these tasks jointly. For example, \citeA{Engonopoulos2014} use a synchronous grammar that directly relates surface strings to target referents, using a chart to compute the possible expressions for a given target. This work bears some relationship to planning-based approaches we discuss in Section \ref{sec:planning} below, which exploit grammatical formalisms as planning operators \cite<e.g.>{Stone1998,Koller2007}, solving realisation and content determination problems in tandem (including {\sc reg} as a special case). 

Finally, in earlier work visual information was typically `simplified' into a table (as we did above), but there has been substantial progress on {\sc reg} in more complex scenarios. For example, the {\sc give} challenge \cite{koller2010report}, provided impetus for the exploration of situated reference to objects in a virtual environment \cite<see also>{Stoia2006,Garoufi2013}. More recent work has started exploring the interface between computer vision and {\sc reg} to produce descriptions of objects in complex, realistic visual scenes, including photographs \cite<e.g.,>{Mitchell2013,Kazemzadeh2014,Mao2016}. This forms part of a broader set of developments focussing on the relatonship between vision and language, which we turn to in Section \ref{sec:image}.

\subsection{Linguistic Realisation}\label{sec:lr}
Finally, when all the relevant words and phrases have been decided upon, these need to be combined to form a well-formed sentence. The simple example in Figure \ref{fig:lr} shows the structure underlying the sentence {\em there were three successive bradycardias down to 69}\/, the linguistic message corresponding to the portion selected from the original signal in Figure \ref{fig:cd}.

Usually referred to as linguistic realisation, this task involves ordering constituents of a sentence, as well as generating the right morphological forms (including verb conjugations and agreement, in those languages where this is relevant). Often, realisers also need to insert function words (such as auxiliary verbs and prepositions) and punctuation marks.  An important complication at this stage is that the output needs to include various linguistic components that may not be present in the input (an instance of the `generation gap' discussed in Section \ref{sec:modular} below); thus, this generation task can be thought of in terms of projection between non-isomorphic structures \cite<cf.>{Ballesteros2015}.  Many different approaches have been proposed, of which we will discuss

\begin{enumerate}
\item human-crafted templates;
\item human-crafted grammar-based systems;
\item statistical approaches.
\end{enumerate}

\subsubsection{Templates} 
When application domains are small and variation is expected to be minimal, realisation is a relatively easy task, and outputs can be specified using templates \cite<e.g.,>{Reiter1995,mcroy2003augmented}, such as the following.

\begin{example}
\$\mbox{\tt{player}} scored for \$\mbox{\tt{team}} in the \$\mbox{\tt{minute}} minute.
\end{example}
This template has three variables, which can be filled with the names of a player, a team, and the minute in which this player scored a goal. It can thus serve to generate sentences like:

\begin{examples}
\item Ivan Rakitic scored for Barcelona in the 4th minute.
\end{examples}

An advantage of templates is that they allow for full control over the quality of the output and avoid the generation of ungrammatical structures. 
 Modern variants of the template-based method include syntactic information in the templates, as well as possibly complex rules for filling the gaps \cite{Theune2001}, making it difficult to distinguish templates from more sophisticated methods \cite{VanDeemter2005}. The disadvantage of templates is that they are labour-intensive if constructed by hand \cite<though templates have recently been learned automatically from corpus data, see e.g.,>[ and the discussion in Section \ref{sec:datadriven} below]{angeli2012parsing,Kondadadi2013}. They also do not scale well to applications which require considerable linguistic variation.
 
\subsubsection{Hand-Coded Grammar-Based Systems}
An alternative to templates is provided by general-purpose, domain-independent realisation systems. Most of these systems are {\em grammar-based}, that is, they make some or all of their choices on the basis of a grammar of the language under consideration. This grammar can be manually written, as in many classic off-the-shelf realisers such as {\sc fuf/surge} \cite{Elhadad1996}, {\sc mumble} \cite{Meteer1987}, {\sc kpml} \cite{Bateman1997}, {\sc nigel} \cite{Mann1983}, and RealPro \cite{Lavoie1997}.  Hand-coded grammar-based realisers tend to require very detailed input. For example, {\sc kpml} \cite{Bateman1997} is based on Systemic-Functional Grammar \cite<{\sc sfg}; >{Halliday2004}, and realisation is modelled as a traversal of a network in which choices depend on both grammatical and semantico-pragmatic information. This level of detail makes these systems difficult to use as simple `plug-and-play' or `off the shelf' modules \cite<e.g.,>{Kasper1989}, something which has motivated the development of simple realisation engines which provide syntax and morphology {\sc api}s, but leave choice-making up to the developer \cite{Gatt2009,Vaudry2013,Bollmann2011,DeOliveira2014,Mazzei2016}. 

One difficulty for grammar-based systems is how to make choices among related options, such as the following, where hand-crafted rules with the right sensitivity to context and input are difficult to design:

\begin{examples}
\item Ivan Rakitic scored for Barcelona in the 4th minute.\\
\item For Barcelona, Ivan Rakitic scored in minute four.\\
\item Barcelona player Ivan Rakitic scored after four minutes.
\end{examples}

\subsubsection{Statistical Approaches}
Recent approaches have sought to acquire probabilistic grammars from large corpora, cutting down on the amount of manual labour required, while increasing coverage. Essentially, two approaches have been taken to include statistical information in the realisation process. One approach, introduced by the seminal work of Langkilde and Knight \cite{Langkilde2000,Langkilde-Geary2002} on the {\sc halogen/nitrogen} systems, relies on a two-level approach, in which a small, hand-crafted grammar is used to generate alternative realisations represented as a forest, from which a stochastic re-ranker selects the optimal candidate. Langkilde and Knight rely on corpus-based statistical knowledge in the form of n-grams, whereas others have experimented with more sophisticated statistical models to perform reranking \cite<e.g.,>{Bangalore2000,Ratnaparkhi2000,cahill2007stochastic}. The second approach does not rely on a computationally expensive generate-and-filter approach, but uses statistical information directly at the level of generation decisions. An example of this approach is the p{\sc cru} system developed by \citeA{Belz2008}, which generates the most likely derivation of a sentence, given a corpus, using a context-free grammar. In this case, the statistics are exploited to control the generator's choice-making behaviour as it searches for the optimal solution.

In both approaches, the base generator is hand-crafted, while statistical information is used to filter outputs. An obvious alternative would be to {\it also}\/ rely on statistical information for the base-generation system. Fully data-driven grammar-based approaches have been developed by acquiring grammatical rules from treebanks. For example, the Open{\sc ccg} framework \cite{hypertagging:acl08,white-rajkumar:2009:EMNLP,deplen:2012:EMNLP} presents a broad coverage English surface realizer, based on Combinatory Categorial Grammar \cite<{\sc ccg}; >{Steedman2000}, relying on a corpus of {\sc ccg} representations derived from the Penn Treebank \cite{Hockenmaier2007} and using statistical language models for re-ranking. There are several other approaches to realisation that adopt a similar rationale, based on a variety of grammatical formalisms, including Head-Driven Phrase Structure Grammar \cite<{\sc hpsg}; >{Nakanishi2005,Carroll2005}, Lexical-Functional Grammar \cite<{\sc lfg}; >{Cahill2006} and Tree Adjoining Grammar \cite<{\sc tag}; >{Gardent2015}. In many of these  systems, the base generator uses some variant of the chart generation algorithm \cite{Kay1996} to iteratively realise parts of an input specification and merge them into one or more final structures, which can then be ranked \cite<see>[for further discussion]{Rajkumar2014}. The existence of stochastic realisers with wide-coverage grammars has motivated a greater focus on subtle choices, such as how to avoid structural ambiguity, or how to handle choices such as explicit complementiser insertion in English \cite<see e.g.,>{Rajkumar2011}. In a somewhat similar vein, the statistical approach to microplanning proposed by \citeA{gardent2017statistical} focuses on interactions between surface realization, aggregation, and sentence segmentation in a joint model.

Other approaches to realisation also rely on one or more classifiers to improve outputs.  For example, \citeA{Filippova2007,Filippova2009} describe an approach to linearisation of  constituents using a two-step approach with Maximum Entropy classifiers, first determining which constituent should occupy sentence-initial position, then ordering the constituents in the remainder of the sentence. \citeA{Bohnet2010} present a realiser using underspecified dependency structures as input, in a framework based on Support Vector Machines, where classifiers are organised in a cascade. An initial classifier decodes semantic input into the corresponding syntactic features, while two subsequent classifiers first linearise the syntax and then render the correct morphological realisation for the component lexemes. 

Modelling choices using classifier cascades is not restricted to realisation alone. Indeed, in some cases, it has been adopted as a model for the {\sc nlg} process as a whole, a topic we will return to in Section \ref{sec:classification}. One outcome of this view of {\sc nlg} is that the nature of the input representation also changes: the more decisions that are made within the statistical generation system, the less linguistic and more abstract the input representation becomes, paving the way for integrated, end-to-end stochastic generation systems, such as \citeA{Konstas2013}, which we also discuss in the next section.

\subsection{Discussion}\label{sec:tasks-disc}
This section has given an overview of some classic tasks that are found in most {\sc nlg} systems. One of the common trends that can be identified in each case is the steady move from early, hand-crafted approaches based on rules, to the more recent stochastic approaches that rely on corpus data, with a concomitant move towards more domain-independent approaches. Historically, this was the case already for individual tasks, such as referring expression generation or realisation, which became topics of intensive research in their own right. However, as more and more approaches to all {\sc nlg} tasks begin to take a statistical turn, there is increasing emphasis on learning techniques; the domain-specific aspect is, as it were, incidental, a property of the training data itself. As we shall see in the next section, this trend has also influenced the way different {\sc nlg} tasks are organised, that is, the architecture of systems for text generation from data.
\section{NLG Architectures and Approaches}\label{sec:architectures}
Having given an overview of the most common sub-tasks that {\sc nlg} systems incorporate, we now turn to the way such tasks can be organised. Broadly speaking, we can distinguish between three dominant approaches to {\sc nlg} architectures:

\begin{enumerate}
\item {\it Modular}\/ architectures: 
By design, such architectures involve fairly crisp divisions among sub-tasks, though with significant variations among them;
\item {\it Planning}\/ perspectives: Viewing text generation as planning links it to a long tradition in {\sc ai} and affords a more integrated, less modular perspective on the various sub-tasks of {\sc nlg};
\item {\it Integrated} or {\it global} approaches: Now the dominant trend in {\sc nlg} (as it is in {\sc nlp} more generally), such approaches cut across task divisions, usually by placing a heavy reliance on statistical learning of correspondences between (non-linguistic) inputs and outputs.
\end{enumerate}

The above typology of {\sc nlg} is based on architectural considerations. An orthogonal question concerns the extent to which a particular approach relies on symbolic or knowledge-based methods, as opposed to stochastic, data-driven methods. It is important to note that none of the three architectural types listed above is inherently committed to one or the other of these. Thus, it is possible for a system to have a modular design but incorporate stochastic methods in several, or even all, sub-tasks. Indeed, our survey of the various tasks in Section \ref{sec:tasks} included several examples of stochastic approaches. Below, we will also discuss a number of data-driven systems whose design is arguably modular. Similarly, it is possible for a system to take a non-modular perspective, but eschew the use of data-driven models (this is a feature of some planning-based {\sc nlg} systems discussed in Section \ref{sec:planning} below, for instance). 

The fact that many modular {\sc nlg} systems are not data-driven is largely due to historical reasons since, of the three designs outlined above, the modular one is the oldest. As we will show below, however, challenges to the classical modular pipeline architecture -- once designated by \citeA{Reiter1994} as the consensus at the time -- have included blackboard and revision-based architectures that were not stochastic. At the same time, it must be acknowledged that the large-scale adoption of integrated, non-modular approaches has been impacted significantly by the uptake of data-driven techniques within the {\sc nlg} community and the development of repositories of data to support training and evaluation. 

In summary, there are at least two orthogonal ways of classifying {\sc nlg} systems, based on their {\em design} or on the {\em methods} adopted in their development. Our survey in this section follows the typology outlined above for convenience of exposition. The caveats raised here should, however, be borne in mind by the reader, and will in any case be brought up repeatedly in what follows, as we discuss different approaches under each heading.

\subsection{Modular Approaches}\label{sec:modular}
\begin{figure}[t]
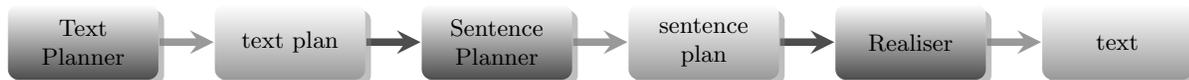

\centering
\begin{minipage}[t][2cm]{\textwidth}
 \begin{center}
 \smartdiagramset{
   border color=none,
   set color list={gray!60!black,white!60!black, gray!60!black, white!60!black, gray!60!black, white!60!black},
   back arrow disabled=true}
	\smartdiagramadd[flow diagram:horizontal]{%
		Text Planner, text plan, Sentence Planner, sentence plan, Realiser, text%
	}{}
\end{center}
\end{minipage}
\caption{Classical three-stage NLG architecture, after \citeA{Reiter2000}. Darker segments illustrate the three main modules; lighter segments show the outputs.}
\label{fig:modular-arch}
\end{figure}

Existing surveys of {\sc nlg}, including those by \citeA{Reiter1997,Reiter2000} and \citeA{Reiter2010}, typically refer to some version of the pipeline architecture displayed in Figure \ref{fig:modular-arch} as the `consensus' architecture in the field. Originally introduced by \citeA{Reiter1994}, the pipeline was a generalisation based on actual practice and was claimed to have the status of a `de facto standard'. This, however, has been contested repeatedly, as we shall see. 

Different modules in the pipeline incorporate different subsets of the tasks described in Section \ref{sec:tasks}. The first module, the {\it Text Planner} (or Document Planner, or Macroplanner), combines content selection and text structuring (or document planning). Thus, it is concerned mainly with strategic generation \cite{McDonald1993}, the choice of `what to say'. The resulting text plan, a structured representation of messages, is the input to the {\it Sentence Planner}\/ (or microplanner), which typically combines sentence aggregation, lexicalisation and referring expression generation \cite{Reiter2000}. If text planning amounts to {\it deciding what to say}\/, sentence planning can be understood as deciding {\it how to say it}. All that remains then is to actually say it, i.e., generate the final sentences in a grammatically correct way, by applying syntactic and morphological rules. This task is performed by the {\it Linguistic Realiser}. Together, sentence planning and realisation encompass the set of tasks traditionally referred to as {\em tactical generation}.

The pipeline architectures shares some characteristics with a widely-used architecture in text summarisation \cite{Mani2001,Nenkova2011}, where the process is sub-divided into (a) analysis of source texts and selection of information; (b) transformation of the selected information to enhance fluency; and (c) synthesis of the summary. 

A second related architecture, which was also noted by \citeA{Reiter1994}, is that proposed in psycholinguistics for human speech production, where the most influential psycholinguistic model of language production, proposed by \citeA{Levelt1989,Levelt1999}, makes a similar distinction between deciding what to say and determining how to say it. Levelt's model allows for a limited degree of self-monitoring through feedback loops, a feature that is absent in Reiter's {\sc nlg} pipeline, but continues to play an important role in psycholinguistics \cite<cf.>{Pickering2013}, though here too there has been increasing emphasis on more integrated models. 

A hallmark of the architecture in Figure \ref{fig:modular-arch} is that it represents clear-cut divisions among tasks that are traditionally considered to belong to the `what' (strategic) and the `how' (tactical). However, this does not imply that this division is universally accepted in practice. In an earlier survey, \citeA{Mellish2006} concluded that while several {\sc nlg} systems incorporate many of the core tasks outlined in Section \ref{sec:tasks}, their organisation varies considerably from system to system. Indeed, some tasks may be split up across modules. For example, the content determination part of referring expression generation might be placed in the sentence planner, but decisions about form (such as whether to use an anaphoric {\sc np}, and if so, what kind of {\sc np} to produce) may have to wait until at least some realisation-related decisions have been taken. Based on these observations, \citeauthor{Mellish2006} proposed an alternative formalism, the `objects-and-arrows' framework, within which different types of information flow between {\sc nlg} sub-tasks can be accommodated. Rather than offering a specific architecture, this framework was intended as a formalism within which high-level descriptions of different architectures can be specified. However, it retains the principle that the tasks, irrespective of their organisation, are relatively well-defined and distinguished.

Another recent development in relation to  the pipeline architecture in Figure \ref{fig:modular-arch} is a proposal by \citeA{Reiter2007} to accommodate systems in which input consists of raw (often numeric) data that requires some preprocessing before it can undergo the kind of selection and planning that the Text Planner is designed to execute. The main characteristic of these systems is that input is unstructured, in contrast to systems which operate over logical forms, or database entries. Examples of application domains where this is the case include weather reporting \cite<e.g.,>{Goldberg1994,Buseman1997,Coch1998,Turner2008a,Sripada2003,Ramos-Soto2015}, where the input often takes the form of numerical weather predictions; and generation of summaries from patient data \cite<e.g.,>{Hueske-kraus2003,Harris2008,Gatt2009,Banaee2013}. In such cases, {\sc nlg} systems  often need to perform some form of {\em data abstraction} (for example, identifying broad trends in the data), followed by {\em data interpretation}. The techniques used to perform these tasks range from extensions of  signal processing techniques \cite<e.g.,>{Portet2009} to the application of reasoning formalisms based on fuzzy set theory \cite<e.g.,>{Ramos-Soto2015}. \citeauthor{Reiter2007}'s \citeyear{Reiter2007} proposal accommodates these steps by extending the pipeline `backwards', incorporating stages prior to Text Planning.

Notwithstanding its elegance and simplicity, there are challenges associated with a pipeline {\sc nlg} architecture, of which two are particularly worth highlighting:

\begin{itemize}
\item The {\em generation gap}\/ \cite{Meteer1991} refers to mismatches between strategic and tactical components, so that early decisions in the pipeline have unforeseen consequences further downstream. To take an example from \citeA{Inui1992}, a generation system might determine a particular sentence ordering during the sentence planning stage, but this might turn out to be ambiguous once sentences have actually been realised and orthography has been inserted;
\item {\em Generating under constraints}: Itself perhaps an instance of the generation gap, this problem can occur when the output of a system has to match certain requirements, for example, it cannot exceed a certain length \cite<see>[for discussion]{Reiter2000a}. Formalising this constraint might appear possible at the realisation stage -- by stipulating the length constraint in terms of number of words or characters, for instance -- but it is much harder at the earlier stages, where the representations are pre-linguistic and their mapping to the final text are potentially unpredictable.
\end{itemize}

These, and related problems, motivated the development of alternative architectures. For instance, some early {\sc nlg} systems were based on an interactive design, in which a module's initially incomplete output could be fleshed out based on feedback from a later module \cite<the {\sc pauline} system is an example of this;>{Hovy1988}. An even more flexible stance is taken in blackboard architectures, in which task-specific procedures are not rigidly pre-organised, but perform their tasks reactively as the output, represented in a data structure shared between tasks, evolves \cite<e.g.,>{Nirenburg1989}. Finally, revision-based architectures allow a limited form of feedback between modules under monitoring, with the possibility of altering choices which prove to be unsatisfactory \cite<e.g.,>{Mann1981,Inui1992}. This has the advantage of not requiring `early' modules to be aware of the consequences of their choices for subsequent modules, since something that goes wrong can always be revised \cite{Inui1992}. Revision need not be carried out exclusively to rectify shortcomings. For instance, \citeA{Robin1993} used revision in the context of sports summaries; an initial draft was revised to add historical background information that was made relevant by the events reported in the draft, also taking decisions as to where to place them in relation to the main text. The price that all of these alternatives potentially incur is, of course, a reduction in efficiency, as noted by \citeA{Smedt1996}. 

Despite early criticisms of the modular approach, the strategic versus tactical division continues to influence recent data-driven approaches to {\sc nlg}, including a number of those discussed in Sections \ref{sec:datadriven} and \ref{sec:deep-learning} below \cite<e.g.>[among others]{Dusek2015,Dusek2016}.

However, other alternatives to pipelines often end up blurring the boundaries between modules in the {\sc nlg} system. This is a feature that is more evident in some planning-based and integrated approaches proposed in recent years. It is to these that we now turn.

\subsection{Planning-Based Approaches}\label{sec:planning}
In {\sc ai}, the planning problem can be described as the process of identifying a sequence of one or more actions to satisfy a particular goal. An initial goal can be decomposed into sub-goals, satisfied by actions each of which has its preconditions and effects. In the classical planning paradigm \cite<{\sc strips};>{Fikes1971}, actions are represented as tuples of such preconditions and effects. 

The connection between planning and {\sc nlg} lies in that text generation can be viewed as the execution of planned behaviour to achieve a communicative goal, where each action leads to a new state, that is, a change in a context that includes both the linguistic interaction or discourse history to date, but also the physical or situated context and the user's beliefs and actions \cite<see>[for some recent perspectives on this topic]{Lemon2008,Rieser2009,Dethlefs2014,Garoufi2013,Garoufi2014}. This perspective on {\sc nlg} is therefore related to the view of `language as action' \cite{Clark1996a}, itself rooted in a philosophical tradition inaugurated by the work of \citeA{Austin1962} and \citeA{Searle1969}. Indeed, some of the earliest {\sc ai} work in this tradition \cite<especially>{Cohen1979,Cohen1985} sought an explicit formulation of preconditions (akin to Searle's felicity conditions) for speech acts and their consequences.

Given that there is in principle no restriction on what types of actions can be incorporated in a plan, it is possible for plan-based approaches to {\sc nlg} to cut across the boundaries of many of the tasks that are normally encapsulated in the classic pipeline architecture, combining both tactical and strategic elements by viewing the problems of what to say and how to say it as part and parcel of the same set of operations. Indeed, there are important precedents in early work for a unified view of {\sc nlg} as a hierarchy of goals, the {\sc kamp} system \cite{Appelt1985} being among the best known examples. For instance, to generate referring expressions in {\sc kamp}, the starting point was reasoning about interlocutors' beliefs and mutual knowledge, whereupon the system generated sub-goals that percolated all the way down to property choice and realisation, finally producing a referential {\sc np} whose predicted effect was to alter the hearer's belief state about the referent \cite<see>[for a similar approach to the generation of referring expressions in dialogue]{Heeman1995}. 

One problem with these perspectives, however, is that deep reasoning about beliefs, desires and intentions \cite<or {\sc bdi}, as it is often called following>{Bratman1987} requires highly expressive formalisms and incurs considerable computational expense. One solution is to avoid general-purpose reasoning formalisms and instead adapt a linguistic framework to the planning paradigm for {\sc nlg}. 

\subsubsection{Planning through the Grammar}
The idea of interpreting linguistic formalisms in planning terms is again prefigured in early {\sc nlg} work. For example, some early systems \cite<e.g. {\sc kpml}, which we briefly discussed in the context of realisation in Section \ref{sec:lr};>{Bateman1997} were based on Systemic-Functional Grammar \cite<{\sc sfg; }>{Halliday2004}, which can be seen as a precursor to contemporary planning-based approaches, since {\sc sfg} models linguistic constructions as the outcome of a traversal through a decision network that extends backwards to pragmatic intentions. In a similar vein, both \citeA{Hovy1991} and \citeA{Moore1993} interpreted the relations of Rhetorical Structure Theory \cite{Mann1988} as  operators for text planning.

Some recent approaches integrate much of the planning machinery into the grammar itself, viewing linguistic structures as planning operators. This requires grammar formalisms which integrate multiple levels of linguistic analysis, from pragmatics to morpho-syntax. It is common for contemporary planning-based approaches to {\sc nlg} to be couched in the formalism of Lexicalised Tree Adjoining Grammar \cite<{\sc ltag}; >{Joshi1997}, though other formalisms, such as Combinatory Categorial Grammar \cite{Steedman2000} have also been shown to be adequate to the task \cite<see especially>[for an approach to generation using Discourse Combinatory Categorial Grammar]{Nakatsu2010}.

In an {\sc ltag}, pieces of linguistic structure (so-called elementary trees in a lexicon) can be coupled with semantic and pragmatic information that specify (a) what semantic preconditions need to obtain in order for the item to be felicitously used; and (b) what pragmatic goals the use of that particular item will achieve \cite<see>[for planning-based work using {\sc ltag}]{Stone1998,Garoufi2013,Koller2002}. As an example of how such a formalism could be deployed in a planning framework, let us focus on the task of referring to a target entity. \citeA{Koller2007} formulated the task in a way that obviates the need to distinguish between the content determination and realisation phases \cite<an approach already taken in>{Stone1998}. Furthermore, they do not separate sentence planning, {\sc reg} and realisation, as is done in the traditional pipeline.  Consider the sentence {\em Mary likes the white rabbit}. Simplifying the formalism for ease of presentation, we can represent the lexical item {\em likes} as follows \cite<this example is based on>[albeit with some simplifications]{Garoufi2014}:

\begin{example}
{\bf likes($u$, $x$, $y$)} action:\\
{\sc preconditions}:
\begin{itemize}
\item The proposition that $x$ likes $y$ is part of the knowledge base (i.e. the statement is supported);
\item $x$ is animate;
\item The current utterance $u$ can be substituted into the derivation $S$ under construction;
\end{itemize}
 {\sc effects}:
 \begin{itemize}
 \item $u$ is now part of $S$
 \item New {\sc np} nodes for $x$ in agent position and $y$ in patient position have been set up (and need to be filled).
 \end{itemize}
\end{example}

As in {\sc strips}, an operator consists of preconditions and effects. Note that the preconditions associated with the lexical item require support in the knowledge base (thus making reference to the input {\sc kb}, which normally would not be accessible to the realiser), and include semantic information (such as that the agent needs to be animate). Having inserted {\em likes} as the sentence's main verb, we have two noun phrases which need to be filled by generating {\sc np}s for the arguments $x$ and $y$. Rather than deferring this task to a separate {\sc reg} module, \citeauthor{Koller2007} build referring expressions by associating further pragmatic preconditions on the linguistic operators (elementary trees) that will be incorporated in the referential {\sc np}. First, the entity must be part of the hearer's knowledge state, since an identifying description (say, to $x$) presupposes that the hearer is familiar with it. Second, an effect of adding words to the {\sc np} (such as the predicates {\em rabbit} or {\em white}) is that the phrase excludes distractors, i.e. entities of which those properties are not true. In a scenario with one human being and two rabbits, only one of which (the $y$ in our example) is white, the derivation would proceed by first updating the {\sc np} corresponding to $y$ with {\em rabbit}, thereby excluding the human from the distractor set, but leaving the goal to distinguish $y$ unsatisfied (since $y$ is not the only rabbit). The addition of another predicate to the {\sc np} ({\em white}) does the trick.

A practical advantage to planning-based approaches is the availability of a significant number of off-the-shelf planners. Once the {\sc nlg} task is formulated in an appropriate plan description language, such as the Planning Domain Definition Language \cite<{\sc pddl}; >{McDermott2000}, it becomes possible in principle to use any planner to generate text. However, planners remain beset by problems of efficiency. In a set of experiments on {\sc nlg} tasks of differing complexity, \citeA{Koller2011} noted that planners tend to spend significant amounts of time on preprocessing, though solutions could often be found efficiently once preprocessing was complete. 

\subsubsection{Stochastic Planning under Uncertainty using Reinforcement Learning}\label{sec:reinforcement}
The approaches to planning we have discussed so far are largely rule-based and tend to view the relationship between a planned action and its consequences (that is, its impact on the context), as fixed \cite<though exceptions exist, as in {\em contingency planning}, which generates multiple plans to address different possible outcomes;>{Steedman2007}. 

As \citeA{Rieser2009} note, this view is unrealistic. Consider a system that generates a restaurant recommendation. The consequences of its output (that is, the new state it gives rise to) are subject to noise arising from several sources of uncertainty. In part, this is due to trade-offs, for example, between needing to include the right amount of information while avoiding excessive prolixity. Another source of uncertainty is the user, whose actions may not be the ones predicted by the system. An instance of Meteer's \citeyear{Meteer1991} generation gap can rear its head, for instance if a stochastic realiser renders the content of a message in an ambiguous, or excessively lengthy utterance \cite{Rieser2009}, a problem that could be addressed by allowing different sub-tasks to share knowledge sources and be guided by overlapping constraints \cite[discussed below]{Dethlefs2015}. 

In short, planning a good solution to reach a communicative goal could be viewed as a stochastic optimisation problem (a theme we revisit in Section \ref{sec:classification} below). This view is shared by many recent approaches based on Reinforcement Learning \cite<{\sc rl};>{Lemon2008,Rieser2009,Rieser2011}, especially those that tackle {\sc nlg} within a dialogue context. In this framework, generation can be modelled as a Markov decision process where states are associated with possible actions and each state-action pair is associated with a probability of moving from a state at time $t$ to a new state at $t+1$ via action $a$. Crucially for the learning algorithm, transitions are associated with a reinforcement signal, via a reward function that quantifies the optimality of the generated output. Learning usually involves simulations in which different generation strategies or `policies' -- essentially, plans corresponding to possible paths through the state space -- come to be associated with different rewards. The {\sc rl} framework has been argued to be better at handling uncertainty in dynamic environments than supervised learning or classification, since these do not enable adaptation in a changing context \cite{Rieser2009}. \citeA{Rieser2011a} showed that this approach is effective in optimising information presentation when generating restaurant recommendations. \citeA{Janarthanam2014} used it to optimise the choice of information to select in a referring expression, given a user's knowledge. The system learns to adapt its user model as the user acquires new knowledge in the course of a dialogue. 

An important contribution of this work has been in exploring joint optimisation, where the policy learned satisfies multiple constraints arising from different sub-tasks of the generation process, by sharing knowledge across the sub-tasks. \citeA{Lemon2011a} showed that joint optimisation can learn a policy that determines when to generate informative utterances or queries to seek more information from a user. Similarly, \citeA{Cuay2011} used hierarchical {\sc rl} to jointly optimise the problem of finding and describing a short route description, while adapting to a user's prior knowledge, giving rise to a strategy whereby the user is guided past landmarks that they are familiar with, while avoiding potentially confusing junctions. Also in a route-finding setting, \citeA{Dethlefs2015} develop a hierarchical model comprising a set of learning agents whose tasks range from content selection through realisation. They show that a joint framework in which agents share knowledge, outperforms an isolated learning framework in which each task is modelled separately. For example, the joint policy learns to give high-level navigation instructions, but switches to low-level instructions if the user goes off-track. Furthermore, utterances produced by the joint policy are less verbose and lead to shorter interactions overall.

The joint optimisation framework is of course not unique to Reinforcement Learning and planning-based approaches. A number of approaches to content determination discussed in earlier sections, including the work of \citeA{Marciniak2005} and \citeA{Barzilay2005}, also use joint optimisation in their approach to content determination and realisation (see Sections \ref{sec:content-det}), as does the work of \citeA{Lampouras2013}. We return to optimisation in Section \ref{sec:classification} below.

In summary, {\sc nlg} research within the planning paradigm has highlighted the desirability of developing unified formalisms to represent constraints on the generation process at multiple levels, whether this is done using {\sc ai}-based planning formalisms \cite{Koller2011}, or stochastically via Reinforcement Learning. Among its contributions, the latter line of work has shed light on the value of (a) hierarchical relationships among sub-problems; and (b) joint optimisation of different sub-tasks. Indeed, the latter trend belongs to a much broader range of research on integrated approaches to {\sc nlg}, to which we turn our attention immediately below.

\subsection{Other Stochastic Approaches to NLG}\label{sec:datadriven}
As we noted at the start of this section, whether a system is data-driven or not is independent of its architectural organisation. Indeed, some of the earliest challenges to a modular or pipeline approach described in Section \ref{sec:modular} above, including revision-based and blackboard architectures, were symbolic in their methodological orientation. At the same time, the shift towards data-driven methods and the availability of data sources has given greater impetus to integrated approaches to {\sc nlg}, although this shift began somewhat later that in other areas of {\sc nlp}. As a result, a discussion of integrated approaches will necessarily tend to emphasise statistical methods.

In the remainder of this section, we start with an overview of methods used to acquire training data for {\sc nlg} -- in particular, pairings of inputs (data) and outputs (text) -- before turning to an overview of techniques and frameworks. One of the themes that will emerge from this overview is that, as in the case of planning, statistical methods often take a unified or `global', rather than a modularised, view of the {\sc nlg} process.

\subsubsection{Acquiring Data}\label{sec:data}
As  noted in Section \ref{sec:tasks}, some {\sc nlg} tasks support the transition to a stochastic approach fairly easily. For example, research on  realisation often exploits the existence of treebanks from which input-output correspondences can be learned. Similarly, the emergence of corpora of referring expressions representing both input domains and output descriptions \cite<e.g.,>{Gatt2007a,Viethen2011a,Kazemzadeh2014,Gkatzia2015} has facilitated the development of probabilistic {\sc reg} algorithms. Shared tasks have also contributed to the development of both data sources and methods (see Section \ref{sec:evaluation}). As we show in Section \ref{sec:image} below, recent work on image-to-text generation has also benefited from the availability of large datasets. For statistical, end-to-end generation in other domains, there is less of an embarrassment of riches. However, this situation is improving as methods to automatically align input data with output text are developed. Still, it is worth emphasising that many of these alignment approaches use data which is semi-structured, rather than the raw, numerical input (e.g., signals) used by the data-to-text systems that \citeA{Reiter2007}, among others, drew attention to.

Currently, there are a number of data-text corpora in specific domains, notably weather forecasting \cite{Reiter2005,Belz2008,Liang2009} and sports summaries \cite{Barzilay2005,Chen2008}. These usually consist of database records paired with free text. A promising recent trend is the introduction of statistical techniques that seek to automatically segment and align such data and text \cite<e.g.,>{Barzilay2005,Liang2009,Konstas2013}. In an influential paper, \citeA{Liang2009} described this framework in terms of a generative model that defines a distribution $p(w\|s)$, for sequences of words $w$ and input states $s$, with latent variables specifying the correspondence between $w$ and $s$ in terms of three main components: (i) the likelihood of database records being selected, given $s$; (ii) the likelihood of certain fields being chosen for some record; (iii) the likelihood that a string of a certain length is generated given the records, fields and states. The parameters of the model can be found using the Expectation Maximization ({\sc em}) algorithm. An example alignment is shown in Figure \ref{fig:liang-example}.

\begin{figure}[t]
\centering
\begin{minipage}{\textwidth}
\begin{tabular}{|l|l|lllll|}
\hline
{\em Events:} & {\sc skycover$_1$} & \multicolumn{5}{c|}{{\sc temperature}$_1$}  \\
{\em Fields:} & {\tt percent=0-25} & & \multicolumn{2}{c}{\tt time=6am-9pm} & {\tt min=9} & {\tt max=21} \\
{\em Text:} & cloudy, & with & \multicolumn{2}{l}{temperatures between} & 10 & 20 degrees. [\textellipsis] \\
\hline
\end{tabular}
\\
\begin{tabular}{|l|lll|ll|}
\hline
{\em Events:} & & \multicolumn{2}{c|}{{\sc winddir}$_1$} & \multicolumn{2}{c|}{{\sc windspeed}$_1$} \\
{\em Fields:} & & {\tt mode=S} & & & {\tt mean=20} \\
{\em Text:} & [\textellipsis] & south & wind & around & 20mph. \\
\hline
\end{tabular}
\end{minipage}
\caption{Database records aligned with text using minimal supervision \cite<after>{Liang2009}.}
\label{fig:liang-example}
\end{figure}

These models perform alignment by identifying regular co-occurrences of segments of data and text. \citeA{koncel2014multi} go beyond this by proposing a model that exploits linguistic structure to align at varying resolutions. For example, (\ref{ex:kedziorski}) below is related to two observations in a soccer game log (an aerial pass and a miss), but can be further analysed into two sub-parts (indicated by indices 1 and 2 in our example), which individually map to these two sub-events.

\begin{example}\label{ex:kedziorski}
(Chamakh rises highest)$_{1}$ and (aims a header towards goal which is narrowly wide)$_{2}$.
\end{example}

A different approach to data acquisition is described by \citeA{Mairesse2014}, who use crowd-sourcing techniques to elicit realisations for semantic/pragmatic inputs describing dialogue acts in the restaurant domain \cite<see>[for another recent approach to crowd-sourcing in a similar domain]{Novikova2016}. The key to the success of this technique is the development of a semantics that is sufficiently transparent for use with non-specialists. In an earlier paper, \citeA{MairesseEtAl2010} describe a method to cut down on the amount of training data required for generation by using {\em uncertainty sampling} \cite{Lewis1994}, whereby a system can be trained on a relatively small amount of input data; subsequently, the learned model is applied to new data, from which the system samples the cases of which it is least certain, forwarding these to a (possibly human) oracle for feedback, which potentially leads to a new training cycle. 

Many of the stochastic end-to-end systems we discuss below rely on well-defined formalisms and typically need fairly precise alignments between inputs and portions of the output. One of the limitations of these approaches is that the reliance on alignment makes such systems highly domain-specific, as noted by \citeA{Angeli2010}.

More recent stochastic methods obviate the need for alignment between input data and output strings. This is the case for many systems based on neural networks \cite<e.g.,>[discussed in Section \ref{sec:deep-learning}]{Wen2015,Dusek2016,Lebret2016,Mei2016} as well as other machine-learning approaches \cite<e.g.,>{Dusek2015,Lampouras2016}. For example, \citeA{Dusek2015} use the dialogue acts from the {\sc bagel} dataset \cite{MairesseEtAl2010} as meaning representations; the {\sc bagel} reference texts are parsed using an off-the-shelf deep syntactic analyser. They define a stochastic sentence planner, a variant of the $A^{*}$ algorithm, which builds optimal sentence plans using a base generator and a scoring function to rank candidates. Realisation is conducted using a rule-based realiser. The approach of \citeA{Lampouras2016}, also uses unaligned {\sc mr}-text pairs from {\sc bagel}, as well as the related {\sc sf} hotel and restaurant dataset by \citeA{Wen2015}. Here, content determination and realisation are both conceived as classification problems (choosing an attribute from the {\sc mr}, or choosing a word for the output), but are optimised jointly in an iterative training algorithm using imitation learning.

\subsubsection{NLG as a Sequential, Stochastic Process}\label{sec:lm}
Given an alignment between data and text, one way of modelling the {\sc nlg} process is to remain faithful to the division between strategic and tactical choices, using the statistical alignment to inform content selection, while deploying {\sc nlp} techniques to acquire rules, templates or schemas \cite<\'{a} la >{McKeown1985} to drive sentence planning and realisation. 

Recall that the generative model of \citeA{Liang2009} pairs data to text based on a sequential, Markov process, combining strategic choices (of {\sc db} records and fields) with tactical choices (of word sequences) into a single probabilistic model. In fact, Markov-based language modelling approaches continue to feature prominently in data-driven {\sc nlg}. One of the earliest examples is the work of \citeA{Oh2002} in the context of a dialogue system in the travel domain, where the input takes the form of a dialogue act (e.g. a query that the system needs to make to obtain information about the user's travel plans) with the attributes to include (e.g. the departure city). \citeauthor{Oh2002}'s approach encompasses both content planning and realisation. It relies on dialogue corpora annotated with utterance classes, that is, the type of dialogue act that each utterance is intended to fulfil. On this basis, they construct separate $n$-gram language models for each utterance class, as well as for word-classes that can appear in the input (for example, words corresponding to {\tt departure city}).  Content planning is handled by a model that predicts which attributes should be included in an utterance on the basis of recent dialogue history. Realisation is handled using a combination of templates and $n$-gram models. Thus, generation is conceived as a two-step (planning followed by realisation) process. 

The reliance on standard language models has one potential drawback, in that such models are founded on a local history assumption, limiting the extent to which prior selections can influence current choices. An alternative, discriminative model \cite<known to the {\sc nlp} community at least since>{Ratnaparkhi96} is logistic regression (Maximum Entropy). The foundations for this approach in {\sc nlg} can be found in the work of \citeA{Ratnaparkhi2000}, who focussed primarily on realisation (albeit combined with elements of sentence planning). He compared two stochastic {\sc nlg} systems based on a maximum entropy learning framework, to a baseline {\sc nlg} system. The first of these ({\sc nlg2} in Ratnaparkhi's paper) uses a conditional language model that generates sentences in an incremental, left-to-right fashion, by predicting the best word given both the preceding history (as in standard n-gram models) and the semantic attributes that remain to be expressed. The second ({\sc nlg3}) augments the model with syntactic dependency relations, performing generation by recursively predicting the left and right children of a given constituent. In an evaluation based on judgements of correctness, \citeauthor{Ratnaparkhi2000} found that the system augmented with dependencies was generally preferred.

In later work, \citeA{Angeli2010} describe an end-to-end {\sc nlg} system that maintains a separation between content selection, sentence planning and realisation, modelling each process as a sequence of decisions in a log-linear framework, where choices can be conditioned on arbitrarily long histories of previous decisions. This enables them to handle long-range dependencies, such as coherence relations, more flexibly \cite<e.g., a model can incorporate the information that a weather report which describes wind speed should do so after mentioning wind direction; see>[for similar insights based on global optimisation]{Barzilay2005}. The separation of tasks is maintained insofar as a different set of features can be used to inform decisions at each stage of the process. Sentence planning and realisation decisions are based on templates acquired from corpus texts: a template is selected based on its likelihood given the database fields selected during content selection. 

\begin{figure}[t]
\centering
\includegraphics[width=0.9\textwidth,height=0.2\textheight]{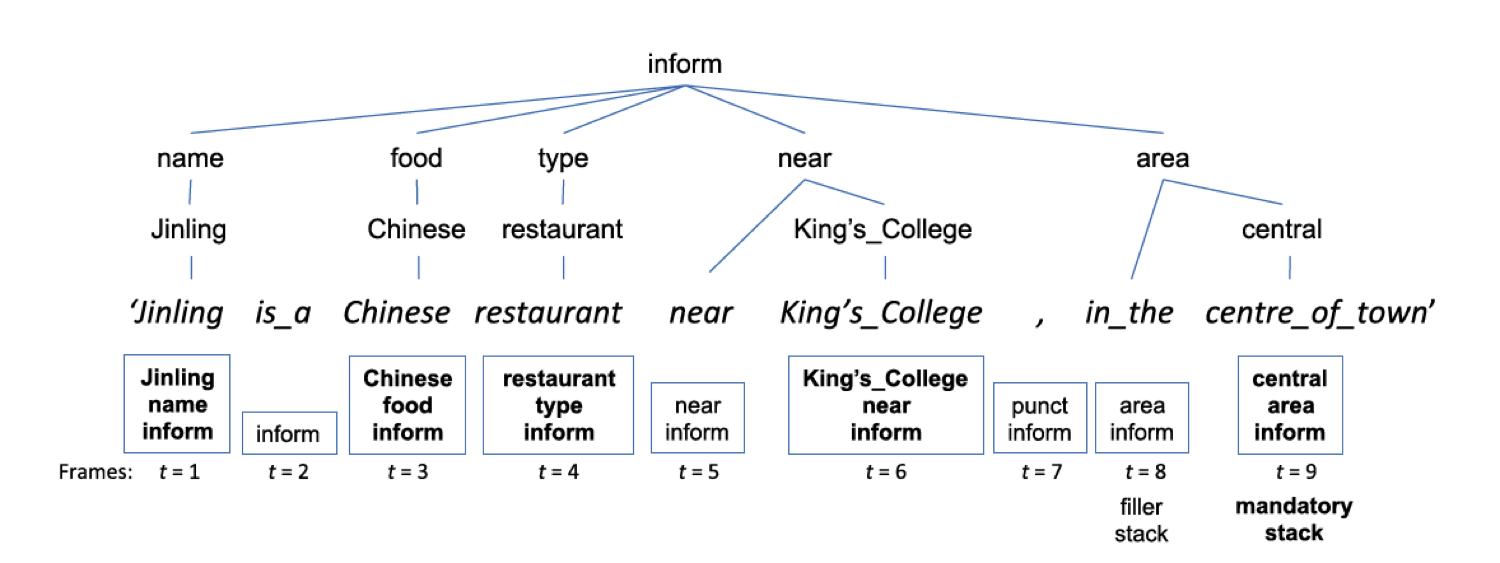}
\caption{Tree structure for a dialogue act, after \citeA{Mairesse2014}. Leaves correspond to word sequences. Non-terminal nodes are semantic attributes, shown at the bottom as semantic stacks. Stacks in bold represent mandatory content.}
\label{fig:mairesseyoung}
\end{figure}

\citeA{Mairesse2014} describe a different approach, which also relies on alignments between database records and text, and seeks a global solution to generation, without a crisp distinction between  strategic and tactical components. In this case, the basic representational framework is a tree of the sort shown in Figure \ref{fig:mairesseyoung}. The root indicates a dialogue act type (in the example, the dialogue act seeks to {\tt inform}). Leaves in the tree correspond to words or word sequences, while nonterminals are {\em semantic stacks}, that is, the pieces of input to which the words correspond. In this framework, content selection and realisation can be solved jointly by searching for the optimal stack sequence for a given dialogue act, and the optimal word sequence corresponding to that stack sequence. \citeauthor{Mairesse2014} use a factored language model ({\sc flm}), which extends n-gram models by conditioning probabilities on different utterance contexts, rather than simply on word histories. Given an input dialogue act, generation works by applying a Viterbi search through the {\sc flm} at each  of the following stages: (a) mandatory semantic stacks are identified for the dialogue act; (b) these are enriched with possible non-mandatory stacks (those which are not in boldface in Figure \ref{fig:mairesseyoung}), usually corresponding to function words; (c) realisations are found for the stack sequence. The approach is also extended to deal with $n-$best realisations, as well as to handle variation, in the form of paraphrases for the same input.

\subsubsection{NLG as Classification and Optimisation}\label{sec:classification}
An alternative way to think about {\sc nlg} decisions at different levels is in terms of classification, already encountered in the context of specific tasks, such as content determination \cite<e.g.,>{Duboue2003} and realisation \cite<e.g.,>{Filippova2007}. Since generation is ultimately about choice-making at multiple levels, one way to model the process is by using a cascade of classifiers, where the output is constructed incrementally, so that any classifier $C_{i}$ uses as (part of) its input the output of a previous classifier $C_{i-i}$. Within this framework, it is still possible to conceive of {\sc nlg} in terms of a pipeline. As \citeA{Marciniak2005} note, an alternative way of thinking about it is in terms of a weighted, multi-layered lattice, where generation amounts to a best-first traversal: at any stage $i$, classifier $C_{i}$ produces the most likely output, which leads to the next stage $C_{i+1}$ along the most probable path. This generalisation is conceptually related to the view of {\sc nlg} in terms of policies in the Reinforcement Learning framework (see Section \ref{sec:reinforcement} above), which define a traversal through sequences of states which may be hierarchically organised \cite<as in the work of>[for example]{Dethlefs2015}.

\citeA{Marciniak2004} start from a small corpus of manually annotated texts of route descriptions, dividing generation into a series of eight classification problems, from determining the linear precedence of discourse units, to determining the lexical form of verbs and the type of their arguments. Generation decisions are taken using the instance-based KStar algorithm, which is shown to outperform a majority baseline on all classification decisions. Instance-based approaches to {\sc nlg} are also discussed by \citeA{Varges2010}, albeit in an overgenerate-and-rank approach where rules overgenerate candidates, which are then ranked by comparison to the instance base.

A similar framework was recently adopted by \citeA{Zarriess2013}, once again taking as their starting point textual data annotated with a dependency representation, as shown in (14) below, where referents are marked {\em v} and {\em p} and the implicit head of the dependency is underlined. 

\begin{examples}\label{ex:zarriess}
\item \gll Junge Familie$_{v:0}$ auf \underline{dem Heimweg}$_{poss:v}$ \underline{ausgeraubt}$_{ag:p}$
      Young family on {the way home} robbed
      \glt `A young family was robbed on their way home.'
      \glend
\end{examples}

These authors use a sequence of classifiers to perform referring expression generation and realisation. They use a ranking model based on Support Vector Machines which, given an input dependency representation extracted from annotated text such as (14), performs two tasks in either order: (a) mapping the input to a shallow syntactic tree for linearisation; and (b) inserting referring expressions. Interestingly, \citeA{Zarriess2013} observe that the performance of either task is order-dependent, in that both classification tasks perform worse when they are second in the sequence. They observe a marginal improvement when the tasks are performed in parallel, but achieve the best performance in a {\em revision-based} architecture, where syntactic mapping is followed by referring expression insertion, followed by a revision of the syntax.

Classification cascades for {\sc nlg} maintain a clean separation between tasks, but research in this area has echoed earlier concerns about pipelines in general (see Section \ref{sec:modular}), the main problem being error propagation. Infelicitous choices will of course impact classification further downstream, a situation analogous to the problem of the generation gap. The conclusion by \citeA{Zarriess2013} in favour of a revision-based architecture, brings our account full circle, in that a well-known solution is shown to yield improvements in a new framework.

Our discussion so far has repeatedly highlighted the fact that a sequential organisation of {\sc nlg} tasks is susceptible to error propagation, whether this takes the form of classifier errors, or decisions in a rule-based module that have a negative impact on downstream components. A potential solution is to view generation as an optimisation problem, where the best combination of decisions is sought in an exponentially large space of possible combinations. We have encountered the use of optimisation techniques, such as Integer Linear Programming ({\sc ilp}) in the context of aggregation and content determination (Section \ref{sec:aggregation}). For example, \citeA{Barzilay2006} group content units based on their pairwise similarity, with an optimisation step to identify a set of pairs that are maximally similar. {\sc ilp} has also been exploited by \citeA{Marciniak2004,Marciniak2005}, as a means to counteract the error propagation problem in their original classification-based approach. Similar solutions have been undertaken by \citeA{Lampouras2013}, in the context of generating text from {\sc owl} ontologies. \citeauthor{Lampouras2013} show that joint optimization using Integer Linear Programming to jointly determine content selection, lexicalisation and aggregation produces more compact verbalisations of ontology facts, compared to a pipeline system \cite<which the authors presented earlier in>{Androtsopoulos2013}. 

Conceptually, the optimisation framework is simple:

\begin{enumerate}
\item Each {\sc nlg} task is once again modelled as classification or label-assignment, but this time, labels are modelled as binary choices (either a label is assigned or not), associated with a cost function, defined in terms of the probability of a label in the training data;
\item Pairs of tasks which are strongly inter-dependent \cite<for example, syntactic choices and {\sc reg} realisations, in the example from>{Zarriess2013} have a cost based on the joint probability of their labels;
\item An {\sc ilp} model seeks the global labelling solution that minimises the overall cost, with the added constraint that if one of a pair of correlated labels $\langle l_{i},l_{j} \rangle$ is selected, the other must be too.
\end{enumerate}

Optimisation solutions have been shown to outperform different versions of the classification pipeline \cite<e.g., that of>{Marciniak2004}, much as the results of \citeA{Dethlefs2015}, discussed above, showed that reinforcement learning of a joint policy produces better dialogue interactions than learning isolated policies for separate {\sc nlg} tasks. The imitation learning framework of \citeA{Lampouras2016} (discussed earlier in Section \ref{sec:data}), which seeks to jointly optimise content determination and realisation, was also shown to achieve competitive results, approaching the performance of the systems of \citeA{Wen2015} on {\sc sf} and of \citeA{Dusek2015} on {\sc bagel}.

\subsubsection{NLG as `Parsing'}\label{sec:parsing}

In recent years, there has been a resurgence of interest in viewing generation in terms of probabilistic context-free grammar ({\sc cfg}) formalisms, or even as the `inverse' of semantic parsing. For example, \citeA{Belz2008} formalises the {\sc nlg} problem entirely in terms of {\sc cfg}s: a base generator expands inputs (bits of weather data in this case) by applying {\sc cfg} rules; corpus-derived probabilities are then used to control the choice of which rules to expand at each stage of the process. The base generator in this work is hand-crafted. However, it is possible to extract rules or templates from corpora, as has been done for aggregation rules \cite[and Section \ref{sec:aggregation}]{Stent2009,White2015}, and also for more general statistical approaches to sentence planning and realisation in a text-to-text framework \cite<e.g.,>{Kondadadi2013}. Similarly, approaches to {\sc nlg} from structured knowledge bases, expressed in formalisms such as {\sc rdf}, have described techniques to extract lexicalised grammars or templates from such inputs paired with textual descriptions \cite{Ell2012,Duma2013,Gyawali2014}.

The work of Mooney and colleagues \cite{Wong2007,Chen2008,Kim2010} has compared a number of different generation strategies inspired by the {\sc wasp} semantic parser \cite{Wong2007}, which uses probabilistic synchronous {\sc cfg} rules learned from pairs of utterances and their semantic representations using statistical machine translation techniques. \citeA{Chen2008} use this framework for generation both by adapting {\sc wasp} in a generation framework, and by further adapting it to produce a new system, {\sc wasper-gen}. While {\sc wasp} seeks to maximise the probability of a meaning representation ({\sc mr}) given a sentence,  {\sc wasper-gen} does the opposite, seeking the maximally probable sentence given an input {\sc mr}, as it were, learning a translation model from meaning to text. When trained on a dataset of sportscasts (the {\sc robocup} dataset), {\sc wasper-gen} outperforms {\sc wasp} on corpus-based evaluation metrics, and is shown to achieve a level of fluency and semantic correctness which approaches that of human text, based on subjective judgements by experimental participants. Note, however, that this framework focusses mainly on tactical generation. Content determination is performed separately, using a variant of the {\sc em}-algorithm to converge on a probabilistic model that predicts which events or predicates should be mentioned.

By contrast, the work of \citeA{Konstas2012,Konstas2013}, which also relies on {\sc cfg}s, uses a unified framework throughout. The starting point is an alignment of text with database records, extending the proposal by \citeA{Liang2009}. The process of converting input data to output text is modelled in terms of rules which implicitly incorporate different types of decisions. For example, given a database of weather records, the rules might take the (somewhat simplified) form shown below:

\begin{examples}
\item $R(\textit{windSpeed}) \rightarrow FS(\textit{temperature}), R(\textit{rain})$\label{ex:konstas1}
\item $FS(\textit{windSpeed,min}) \rightarrow F(\textit{windSpeed,max}) FS(\textit{windSpeed,max})$\label{ex:konstas2}
\item $FS(\textit{windSpeed,min}) \rightarrow W(\textit{windSpeed,min})$\label{ex:konstas3}
\end{examples}
where $R$ stands for a database record, $FS$ is a set of fields, $F(x,y)$ stands for field $y$ in record $x$, $W$ is a word sequence, and all rules have associated probabilities that condition the {\sc rhs} on the {\sc lhs}, akin to the {\sc pcfg}s used in parsing. These rules specify that a description of {\em windSpeed} (\ref{ex:konstas1}) should be followed in the text by a temperature and a rain report. According to rule (\ref{ex:konstas2}), minimum windspeed should be followed by a mention of the maximum windspeed with a certain probability. Rule (\ref{ex:konstas3}) expands the minimum windspeed rule to a sequence of words according to a bigram language model \cite{Konstas2012}. \citeA{Konstas2012} pack the set of rules  acquired from the alignment stage into a hypergraph, and treat generation as decoding to find the maximally likely word sequence. 

Under this view, generation is akin to inverted parsing. Decoding proceeds using an adaptation of the {\sc cyk} algorithm. Since the model defining the mapping from input to output does not incorporate fluency heuristics, the decoder is interleaved with two further sources of linguistic knowledge by \citeA{Konstas2013}: (a) a weighted finite-state automaton (representing an n-gram language model); and (b) a dependency model \cite<cf.>[, also discussed above]{Ratnaparkhi2000}. 

\subsubsection{Deep Learning Methods}\label{sec:deep-learning}
We conclude our discussion of data-driven {\sc nlg} with an overview of applications of deep neural network ({\sc nn}) architectures. The decision to dedicate a separate section is warranted by the recent, renewed interest in these models \cite<see>[for an {\sc nlp}-focussed overview]{Goldberg2016}, as well as the comparatively small (but steadily growing) range of {\sc nlg} models couched within this framework to date. We will also revisit {\sc nn} models for {\sc nlg} under more specific headings in the following sections, especially in discussing stylistic variation (Section \ref{sec:style}) and the image captioning (Section \ref{sec:image}), where they are now the dominant approach.

As a matter of fact, applications of {\sc nn}s in {\sc nlg} hark back at least to \citeA{Kukich1987}, though her work was restricted to small-scale examples. Since the early 1990s, when interest in neural approaches waned in the {\sc nlp} and {\sc ai} communities, cognitive science research has continued to explore their application to syntax and language production \cite<e.g.,>{Elman1990,Elman1993,Chang2006}. The recent resurgence of interest in {\sc nn}s is in part due to advances in hardware that can support resource-intensive learning problems \cite{Goodfellow2016}. More importantly, {\sc nn}s are designed to learn representations at increasing levels of abstraction by exploiting backpropagation \cite{LeCun2015,Goodfellow2016}. Such representations are dense, low-dimensional, and distributed, making them  well-suited to capturing grammatical and semantic generalisations \cite<see>[{\em inter alia}]{Mikolov2013,Luong2013,Pennington2014}. {\sc nn}s have also scored notable successes in sequential modelling using feedforward networks \cite{Bengio2003,Schwenk2005}, log-bilinear models \cite{Mnih2007} and recurrent neural networks \cite<{\sc rnn}s, >{Mikolov2010}, including {\sc rnn}s with long short-term memory units \cite<{\sc lstm}, >{Hochreiter1997}. The latter are now the dominant type of {\sc rnn} for language modelling tasks. Their main advantage over standard language models is that they handle sequences of varying lengths, while avoiding both data sparseness and an explosion in the number of parameters through the projection of histories into a low-dimensional space, so that similar histories share representations. 

A demonstration of the potential utility of recurrent networks for {\sc nlg} was provided by \citeA{Sutskever2011}, who used a character-level {\sc lstm} model for the generation of grammatical English sentences. This, however, focussed exclusively on their potential for realisation. Models that generate from semantic or contextual inputs cluster around two related types of models, descibed below.

\subsubsection{Encoder-Decoder Architectures} 
An influential architecture is the Encoder-Decoder framework \cite{Sutskever2014}, where an {\sc rnn} is used to encode the input into a vector representation, which serves as the auxiliary input to a decoder {\sc rnn}. This decoupling between encoding and decoding makes it possible in principle to share the encoding vector across multiple {\sc nlp} tasks in a multi-task learning setting \cite<see>[for some recent case studies]{Dong2015,Luong2016}. Encoder-Decoder architectures are particularly well-suited to Sequence-to-Sequence ({\sc seq2seq}) tasks such as Machine Translation, which can be thought of as requiring the mapping of variable-length input sequences in the source language, to variable-length sequences in the target \cite<e.g.,>{Kalchbrenner2013,Bahdanau2015}. It is easy to adapt this view to data-to-text {\sc nlg}. For example, \citeA{CastroFerreira2017} adapt {\sc seq2seq} models for generating text from abstract meaning representations ({\sc amr}s).

A further important development within the Encoder-Decoder paradigm is the use of attention-based mechanisms, which force the encoder, during training, to weight parts of the input encoding more when predicting certain portions of the output during decoding \cite<cf.>{Bahdanau2015,Xu2015}. This mechanism obviates the need for direct input-output alignment, since attention-based models are able to learn input-output correspondences based on loose couplings of input representations and output texts \cite<see>[for discussion]{Dusek2016}.  

In {\sc nlg}, many approaches to response generation in an interactive context (such as dialogue or social media posts) adopt this architecture. For example, \citeA{Wen2015} use semantically-conditioned {\sc lstm}s to generate the next act in a dialogue; a related approach is taken by \citeA{Sordoni2015}, who use {\sc rnn}s to encode both the input utterance and the dialogue context, with a decoder to predict the next word in the response \cite<see also>{Serban2016}. \citeA{Goyal2016} found an improvement in the quality of generated dialogue acts when using a character-based, rather than a word-based {\sc rnn}. 

\citeA{Dusek2016} also use a {\sc seq2seq} model with attention for dialogue generation, comparing an end-to-end model where content selection and realisation are jointly optimised (so that outputs are strings), to a model which outputs deep syntax trees, which are then realised using an off-the-shelf realiser \cite<as done in>{Dusek2015}. Like \citeA{Wen2015}, they use a reranker during decoding to rank beam search outputs, penalising those that omit relevant information or include irrelevant information. Their evaluation, on {\sc bagel}, shows that the joint optimisation setup is superior to the {\sc seq2seq} model that generates trees for subsequent realisation. \citeA{Mei2016} also explicitly address the division into content selection and realisation, using {\sc weathergov} data \cite{Angeli2010}. 
They use a bidirectional {\sc lstm} encoder to map input records to a hidden state, followed by an attention-based aligner which models content selection, determining which records to mention as a function of their prior probability and the likelihood of their alignment with words in the vocabulary; a further refinement step weights the outcomes of the alignment with the priors, making it more likely that more important records will be verbalised. In this approach, {\sc lstm}s are able to learn long-range dependencies between records and descriptors, which the log-linear model of Angeli factored in explicitly (see Section \ref{sec:lm} above). Comparable approaches are now also use for automatic generation of poetry \cite<see e.g., >{zhang2014chinese}, a topic to which we will return below.

\subsubsection{Conditioned Language Models} 
A related view of the data-to-text process views the generator  as a conditioned language model, where output is generated by sampling words or characters from a distribution conditioned on input features, which may include semantic, contextual or stylistic attributes. For example, \citeA{Lebret2016} restricts generation to the initial sentence of wikipedia biographies from the corresponding wiki fact table and models content selection and realisation jointly in a feedforward {\sc nn} \cite{Bengio2003}, conditioning output word probabilities on both local context and global features obtained from the input table. This biases the model towards full coverage of the contents of a field. For example, a field in the table containing a person's name typically consists of more than one word and the model should concatenate the words making up the entire name. While simpler than some of the models discussed above, this model can also be thought of as incorporating an attentional mechanism. \citeA{Lipton2016} use character-level {\sc rnn}s conditioned on semantic information and sentiment, to generate product reviews, while \citeA{Tang2016} generate such reviews using an {\sc lstm} conditioned on input `contexts', where contexts incorporate both discrete (user, location etc) and continuous information. Similar approaches have been adopted in a number of models for stylistic and affective generation \cite<see>[and the discussion in Section \ref{sec:style} below]{Li2016,Herzig2017,Ashgar2017,Hu2017,Ficler2017}.

\subsection{Discussion}
An important theme that has emerged from recent work is the blurring of boundaries between tasks that are encapsulated in traditional architectures. This is evident in planning-based approaches, but perhaps the most radical break from this perspective arises in stochastic data-to-text systems which capitalise on alignments between input data and output text, combining content-oriented and linguistic choices within a unified framework. Among the open questions raised by research on stochastic {\sc nlg} is the extent to which sub-tasks need to be jointly optimised and, if so, which knowledge sources should be shared among them. This is also seen in recent work using neural models, where joint learning of content selection and realisation has been claimed to yield superior outputs, compared to models that leave the tasks separate \cite<e.g.,>{Dusek2016}.

An outstanding issue is the balancing act between achieving adequate textual output versus doing so efficiently and robustly. Early approaches that departed from a pipeline architecture tended to sacrifice the latter in favour of the former; this was the case in revision-based and blackboard architectures. The same is to some extent true of planning-based approaches which are rooted in paradigms with a long history in {\sc ai}: As recent empirical work has shown \cite{Koller2011}, these too are susceptible to considerable computational cost, though this comes with the advantage of a unified view of language generation that is also compatible with  well-understood linguistic formalisms, such as {\sc ltag}. 

Stochastic approaches present a different problem, namely, that of acquiring the right data to construct the necessary statistical models. While plenty of datasets have become available, for tasks such as recommendations in the restaurant or hotel domains, brief weather reports, or sports summaries, it remains to be seen whether data-driven {\sc nlg} models can be scaled up to domains where large volumes of heterogeneous data (numbers, symbols etc) are the norm, and where longer texts need to be generated. While such data is not easy to come by, crowd-sourcing techniques can presumably be exploited \cite{Mairesse2014,Novikova2016}.

As we have seen, systems vary in whether they require aligned data (by which we mean data where strings are paired with the portion of the input to which they correspond), or not. 
As deep learning approaches become more popular -- and, as we shall see in the next section, they are now the dominant approach in certain tasks, such as generating image captions -- the need for alignment is becoming less acute, as looser input-output couplings can constitute adequate training data, especially in models that incorporate attentional mechanisms. As these techniques become better understood, they are likely to feature more heavily in a broader range of {\sc nlg} tasks, as well as end-to-end {\sc nlg} systems. 

A second possible outcome of the renewed interest in deep learning is its impact on representation learning and architectures. In a recent opinion piece, \citeA{Manning2015} suggested that the contribution of deep learning to {\sc nlp} has to date been mainly due to the power of distributed representations, rather than the exploitation of the `depth' of multi-layered models. Yet, as Manning also notes, greater depth can confer representational advantages. As researchers begin to define complex architectures that `self-organise' during training by minimising a loss function, it might turn out that different components of such architectures acquire core representations pertaining to different aspects of the problem at hand. This raises the question whether such representations could be reusable, in the same way that the layers of deep convolutional networks in computer vision learn representations at different levels of granularity which turn out to be reusable in a range of tasks \cite<not just object recognition, for instance, even though networks such as {\sc vgg} are typically trained for such tasks; see>{Simonyan2015}. A related aim, suggested by recent attempts at transfer learning, especially in the {\sc seq2seq} paradigm, is to attempt to learn domain-invariant representations that carry over from one task to another.

Could {\sc nlp}, and the field of {\sc nlg} in particular, be about to witness a renewed emphasis on multi-levelled approaches to {\sc nlg}, with `deep' architectures whose components learn optimal representations for different sub-tasks, perhaps along the lines detailed in Section \ref{sec:tasks} above? And to what extent would such representations be reusable? As a number of other commentators have pointed out, the prospect of learning domain-invariant linguistic representations that facilitate transfer learning in {\sc nlp}, remains somewhat elusive, despite certain notable successes, not least those scored in the development of distributed word representations.\footnote{For some remarks on this topic, see for example the blog entry by \citeA{Ruder2017}. A recent note of caution against unrealistic claims of success of neural methods in {\sc nlg} was sounded by \citeA{Goldberg2017}.} This could well be the next frontier in research on statistical {\sc nlg}.

In the following sections, we turn our attention away from standard tasks and the way they are organised, focussing on three broad topics -- image-to-text generation, stylistic variation and computational creativity -- in which {\sc nlg} research has also intersected with research in other areas of Artificial Intelligence and {\sc nlp}.

\section{The Vision-Language Interface: Image Captioning and Beyond}\label{sec:image}
Over the past few years, there has been an explosion of interest in the task of automatically generating captions for images, as part of a broader endeavour to investigate the interface between vision and language \cite{Barnard2016}. Image captioning is arguably a paradigm case of data-to-text generation, where the input comes in the form of an image. The task has become a research focus not only in the {\sc nlg} community but also in the computer vision community, raising the  possibility of more effective synergies between the two groups of researchers. Apart from its practical applications, the grounding of language in perceptual data has long been a matter of scientific interest in {\sc ai} \cite<see>[for a variety of theoretical views on the computational challenges of the perception-language interface]{Winograd1972,Harnad1990,Roy2005}.  

\begin{figure}[!h]
\centering
	\begin{subfigure}[b]{0.3\textwidth}
        \includegraphics[width=\textwidth]{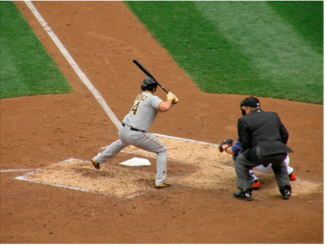}
        \caption{The man at bat readies to swing at the pitch while the umpire looks on  \cite<Human-authored caption from the {\sc ms-coco} dataset>{Lin2014}}
 	\end{subfigure}
 	\begin{subfigure}[b]{0.3\textwidth}
        \includegraphics[width=\textwidth]{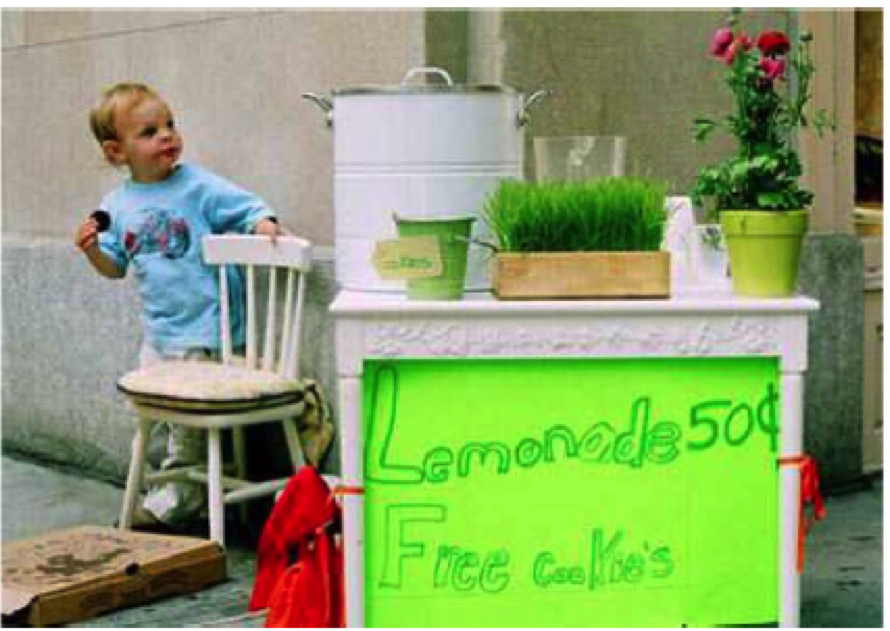}
        \caption{This picture shows one person, one grass, one chair, and one potted plant. The person is near the green grass, and in the chair. The green grass is by the chair, and near the potted plant \cite{Kulkarni2011}}
        \label{fig:kulkarni2011}
 	\end{subfigure}
 	\begin{subfigure}[b]{0.3\textwidth}
        \includegraphics[width=\textwidth]{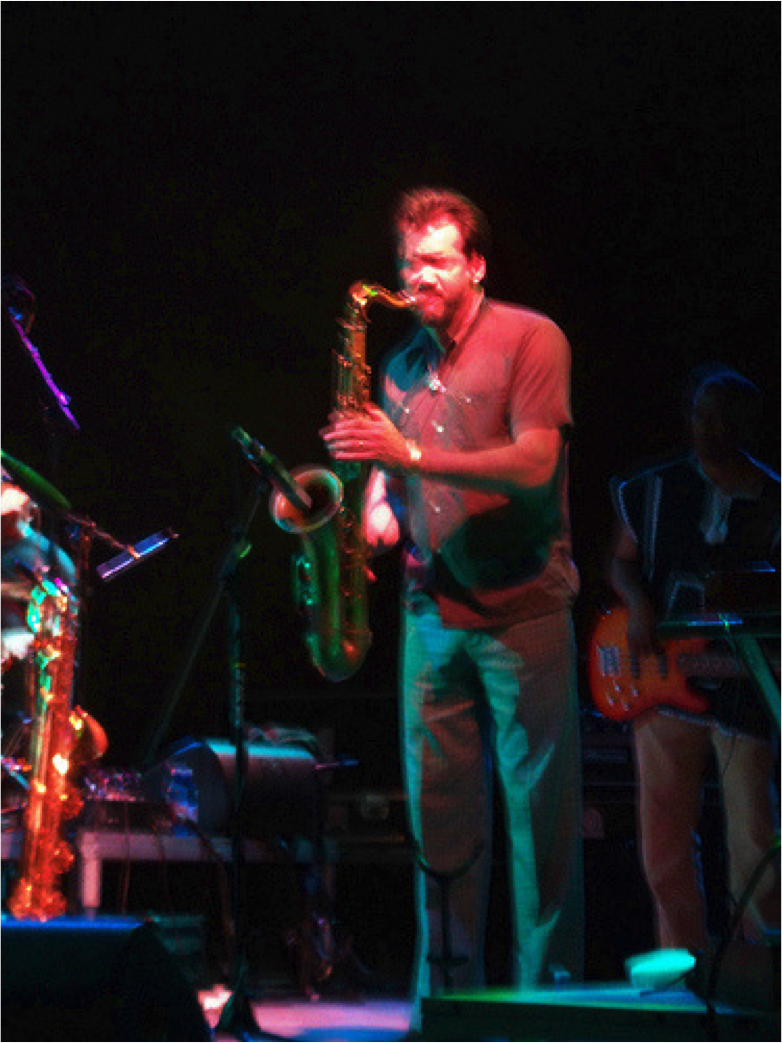}
        \caption{A person is playing a saxophone \cite{Elliott2015}}
        \label{fig:elliott2015}
 	\end{subfigure}
 	\begin{subfigure}[b]{0.3\textwidth}
        \includegraphics[width=\textwidth]{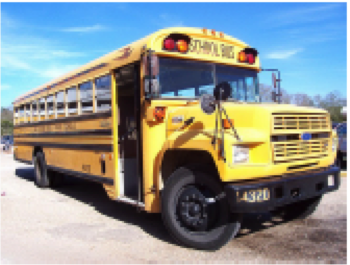}
        \caption{A bus by the road with a clear blue sky \cite{Mitchell2012}}
        \label{fig:mitchell2012}
 	\end{subfigure}
	\begin{subfigure}[b]{0.3\textwidth}
        \includegraphics[width=\textwidth]{img/mao2015.png}
        \caption{A bus is driving down the street in front of a building \cite{Mao2015}}
        \label{fig:mao2015}
 	\end{subfigure}
 	\begin{subfigure}[b]{0.3\textwidth}
        \includegraphics[width=\textwidth]{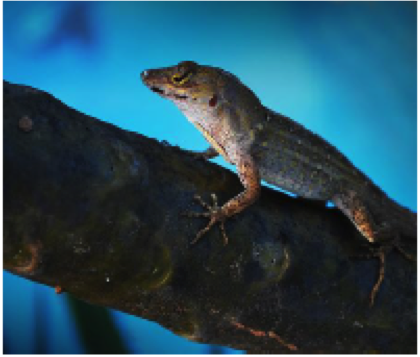}
        \caption{A gecko is standing on a branch of a tree  \cite{Hendricks2016}}
        \label{fig:hendricks2016}
 	\end{subfigure}
\caption{Some caption generation examples}
\label{fig:img2text}
\end{figure}

Figure \ref{fig:img2text} shows some examples of caption generation, sampled from publications spanning approximately 6 years. 
Current caption generation research focusses mainly on what \citeA{Hodosh2013} refer to as {\em concrete conceptual} image descriptions of elements directly depicted in a scene. As \citeA{Donahue2015} put it, image captioning is a task whose input is static and non-sequential (an image, rather than, say, a video), whereas the output is sequential (a multi-word text), in contrast to non-sequential outputs such as object labels \cite<e.g.>[among others]{Duygulu2002,Ordonez2016}. 

Our discussion will be brief, since image captioning has recently been the subject of an extensive review by \citeA{Bernardi2016}, and has also been discussed against the background of broader issues in research on the vision-language interface by \citeA{Barnard2016}. While the present section draws upon these sources, it is organised in a somewhat different manner, also bringing out the connections with {\sc nlg} more explicitly.

\subsection{Data}\label{sec:im-data}
A detailed overview of datasets is provided by \citeA{Bernardi2016}. \citeA{Ferraro2015} offer a systematic comparison of datasets for both caption generation and visual question answering with an accompanying online resource.\footnote{The resource provided by \citeA{Ferraro2015} can be found at \url{http://visionandlanguage.net}.} 

Datasets typically consist of images paired with one or more human-authored captions (mostly in English) and vary from artificially created scenes \cite{Zitnick2013} to real photographs. Among the latter, the most widely used are Flickr8k \cite{Hodosh2013}, Flickr30k \cite{Young2014} and {\sc ms-coco} \cite{Lin2014}.
Datasets such as the {\sc sbu1m} Captioned Photo Dataset \cite{Ordonez2011} include naturally-occurring captions of user-shared photographs on sites such as Flickr; hence the captions included therein are not restricted to the concrete conceptual. There are also a number of specialised, domain-specific datasets, such as the Caltech {\sc ucsd} Birds datast \cite<{\sc cub}; >{Wah2011}.

There have also been a number of shared tasks in this area, including the {\sc coco} (`Common Objects in Context') Captioning Challenge\footnote{\url{http://mscoco.org/dataset/\#captions-challenge2015}}, organised as part of the Large-Scale Scene Understanding Challenge ({\sc lsun})\footnote{\url{http://lsun.cs.princeton.edu/2016/}} and the Multimodal Machine Translation Task \cite{Elliott2016}. We defer discussion of evaluation of image captioning systems to Section \ref{sec:evaluation} of this paper, where it is discussed in the context of {\sc nlg} evaluation as a whole.

\subsection{The Core Tasks}

There are two logically distinguishable sub-tasks in an image captioning system, namely, image analysis and text generation. This is not to say that they need to be organised separately or sequentially. However, prior to discussing architectures as such, it is worth briefly giving an overview of the methods used to deal with these two tasks.

\subsubsection{Image Analysis}
 There are three main groups of approaches to treating visual information for captioning purposes.

\paragraph{Detection} Some systems rely on computer vision methods for the detection and labelling of objects, attributes, `stuff' (typically mapped to mass nouns, such as {\em grass}), spatial relations, and possibly also action and pose information. This is usually followed by a step mapping these outputs to linguistic structures (`sentence plans' of the sort discussed in Section \ref{sec:tasks} and \ref{sec:architectures}), such as trees or templates \cite<e.g.>{Kulkarni2011,Yang2011,Mitchell2012,Elliott2015,Yatskar2014,Kuznetsova2014a}. Since performance depends on the coverage and accuracy of detectors
 \cite{Kuznetsova2014a,Bernardi2016}, some work has also explored generation from gold standard image annotations \cite{Elliott2013,Wang2015,Muscat2015} or artificially created scenes in which the components are known in advance \cite{Ortiz2015}. 

\paragraph{Holistic scene analysis} Here, a more holistic characterisation of a scene is used, relying on features that do not typically identify objects, attributes and the like. Such features include  {\sc rgb} histograms, scale-invariant feature transforms \cite<{\sc sift};>{Lowe2004}, or low-dimensional representations of spatial structure \cite<as in  {\sc gist};>{Oliva2001}, among others. This kind of image processing is often used by systems that frame the task in terms of retrieval, rather than caption generation proper. Such systems either use a unimodal space to compare a query image to training images before caption retrieval
\cite<e.g.>{Ordonez2011,Gupta2012}, or exploit a multimodal space representing proximity between images and captions 
\cite<e.g.>{Hodosh2013,Socher2014}. 

\paragraph{Dense image feature vectors} Given the success of convolutional neural networks ({\sc cnn})  for computer vision tasks \cite<cf. e.g.,>{LeCun2015}, many deep learning approaches use features from a pre-trained {\sc cnn} such as AlexNet \cite{Krizhevsky2012}, {\sc vgg} \cite{Simonyan2015} or Caffe \cite{Jia2014}. Most commonly, caption generators use an activation layer from the pre-trained network as their input features \cite<e.g.>{Kiros2014,Karpathy2014,Karpathy2015,Vinyals2015,Mao2015,Xu2015,Yagcioglu2015,Hendricks2016}. 

\subsubsection{Text Generation or Retrieval}
Depending on the type of image analysis technique, captions can be generated using a variety of different methods, of which the following are well-established.

\paragraph{Using templates or trees} Systems relying on detectors can map the output to linguistic structures in a sentence planning stage. For example, objects can be mapped to nouns, spatial relations to prepositions, and so on.
\citeA{Yao2010} use semi-supervised methods to parse images into graphs and then generate text via a simple grammar. 
Other approaches rely on sequence classification algorithms, such as Hidden Markov Models \cite{Yang2011} and conditional random fields \cite{Kulkarni2011,Kulkarni2013}. \citeA[see the example in Figure \ref{fig:kulkarni2011}]{Kulkarni2013} experiment with both templates and web-derived $n$-gram language models, finding that the former are more fluent, but suffer from lack of variation, an issue we also addressed earlier, in connection with realisation (Section \ref{sec:lr}). 

In the Midge system \cite[see Figure \ref{fig:mitchell2012} for an example caption]{Mitchell2012}, input images are represented as triples consisting of object/stuff detections, action/pose detections and spatial relations. These are subsequently mapped to  $\langle \textit{noun, verb, preposition} \rangle$ triples and realised using a tree substitution grammar. This is further enhanced with the ability to `hallucinate' likely words using a probabilistic model, that is, to insert words which are not directly grounded in the detections performed on the image itself, but have a high probability of occurring, based on corpus data. 
In a human evaluation, Midge was shown to outperform both the system by \citeA{Kulkarni2011} and \citeA{Yang2011} on a number of criteria, including humanlikeness and correctness.

\citeA{Elliott2013} use  {\em visual dependency representations} ({\sc vdr}), a dependency grammar-like formalism to describe spatial relations between objects based on physical features such as proximity and relative position. Detections from an image are mapped to their corresponding {\sc vdr} relations prior to generation \cite<see also>[and the example in Figure \ref{fig:elliott2015}]{Elliott2015}. \citeA{Ortiz2015} use {\sc ilp} to identify pairs of  objects in abstract scenes \cite{Zitnick2013a} before mapping them to a {\sc vdr}. Realisation is framed as a machine translation task over {\sc vdr}-text pairs. A similar concern with identifying spatial relations is found in the work of \citeA{Lin2015}, who  use scene graphs as input to a grammar-based realiser. \citeA{Muscat2015} propose a naive Bayes model to predict spatial prepositions based on image features such as object proximity and overlap. 

\paragraph{Using language models} 
Using language models has the potential advantage of facilitating joint training from image-language pairs. It may also yield more expressive or creative captions if it is used to overcome the limitations of grammars or templates \cite<as shown by the example of Midge;>{Mitchell2012}.
In some cases, n-gram models are trained on out-of-domain data,  the approach taken by \citeA{Li2011} using web-scale $n$-grams, and by \citeA{Fang2015}, who used a maximum entropy language model. Most deep learning architectures use language models in the form of vanilla {\sc rnn}s or long short-term memory networks
 \cite<e.g.>{Kiros2014,Vinyals2015,Donahue2015,Karpathy2015,Xu2015,Hendricks2016,Hendricks2016a,Mao2016}. These architectures model caption generation as a process of predicting the next word in a sequence. Predictions are biased both by the caption history generated so far (or the start symbol for initial words) and by the image features which, as noted above, are typically features extracted from a {\sc cnn} trained on the object detection task.

\paragraph{Caption retrieval and recombination} Rather than generate captions, some systems retrieve them based on training data. The advantage of this is that it guarantees fluency, especially if retrieval is of whole, rather than partial, captions.
\citeA{Hodosh2013} used a multimodal space to represent training images and captions, framing retrieval as a process of identifying the nearest caption to a query image.
The idea of `wholesale' caption retrieval has a number of precedents. For example \citeA{Farhadi2010} use Markov random fields to parse images into $\langle \textit{object,action, scene} \rangle$ triples, paired with parsed captions. A caption for a query image is retrieved by comparing it to the parsed images in the training data, finding the most similar based on WordNet.
Similarly, the {\tt Im2Text} \cite{Ordonez2011} system ranks candidate captions for a query image. \citeA{Devlin2015} use a $k$ nearest neighbours approach, with caption similarity quantified using {\sc bleu} \cite{Papineni2002} and {\sc cide}r \cite{Vedantam2015}. A different view of retrieval is proposed by \citeA{Feng2010}, who use extractive summarisation techniques to retrieve descriptions of images and associated narrative fragments from their surrounding text in news articles.

A potential drawback of wholesale retrieval is that captions in the training data may not be well-matched to a query image. For instance,
\citeA{Devlin2015} note that the less similar a query is to training images, the more generic the caption returned by the system. A possible solution is to use partial matches, retrieving and recombining caption fragments.
\citeA{Kuznetsova2014a} use detectors to match query images to training instances, retrieving captions in the form of parse tree fragments which are then recombined.
\citeA{Mason2014} use a domain-specific dataset to extract descriptions and adapt them to a query image using a joint visual and textual bag-of-words model.
In the deep learning paradigm, both \citeA{Socher2014} and \citeA{Karpathy2014} use word embeddings derived from dependency parses, which are projected, together with {\sc cnn} image features, into a multimodal space. Subsequent work by \citeA{Karpathy2015} showed that this fine-grained pairing works equally well with word sequences, eschewing the need for dependency parsing. 

Recently, \citeA{Devlin2015a} compared nearest-neighbour retrieval approaches to different types of language models for caption generation, specifically, the Maximum Entropy approach of \citeA{Fang2015}, an {\sc lstm}-based approach and {\sc rnn}s which are coupled with a {\sc cnn} for image analysis \cite<e.g.>{Vinyals2015,Donahue2015,Karpathy2015}. A comparison of the linguistic quality of captions suggested that there was a significant tendency for all models to reproduce captions observed in the training set, repeating them for different images in the test set. This could be due to a lack of diversity in the data, which might also explain why the nearest neighbour approach compares favourably with language model-based approaches.

\subsection{How is Language Grounded in Visual Data?}
As the foregoing discussion suggests, views on the relationship between visual and linguistic data depend on how each of the two sub-tasks is dealt with. Thus, systems which rely on detections tend to make a fairly clear-cut distinction between input processing and content selection on the one hand, and sentence planning and realisation on the other \cite<e.g.>{Kulkarni2011,Mitchell2012,Elliott2013}. 
The link between linguistic expressions and visual features is mediated by the outcomes of the detectors. For example, Midge \cite{Mitchell2012} uses the object detections to determine which nouns to mention, before fleshing out the caption with attributes (mapped to adjectives) and verbs. Similarly, \citeA{Elliott2013} uses {\sc vdr}s to determine spatial expressions.

Retrieval-based systems relying on unimodal or multimodal similarity spaces represent the link between linguistic expressions and image features more indirectly. Here, similarity plays the dominant role. In a unimodal space  \cite{Ordonez2011,Gupta2012,Mason2014,Kuznetsova2012,Kuznetsova2014a}, it is images which are compared, with (partial) captions retrieved based on image similarity. A number of deep learning approaches also broadly conform to this scheme. For example, both \citeA{Yagcioglu2015} and \citeA{Devlin2015} retrieve and rank captions for a query image, using a {\sc cnn} for the representation of the visual space. By contrast, multimodal spaces involve a direct mapping between visual and linguistic features \cite<e.g.>{Hodosh2013,Socher2014,Karpathy2014}, enabling systems to map from images to `similar' -- that is, related or relevant -- captions.

Much interesting work on vision-language integration is being carried out with deep learning models.
 \citeA{Kiros2014} introduced multimodal neural language models ({\sc mrnn}), experimenting with two main architectures. Their Modality-Biased Log-Bilinear Model ({\sc mlbl-b}) uses an additive bias to predict the next word in a sequence based on both the linguistic context and {\sc cnn} image features. The Factored 3-way Log-Bilinear Model ({\sc mlbl-f}) also gates the representation matrix for a word with image features.
  In a related vein, \citeA{Donahue2015} propose a combined {\sc cnn} $+$ {\sc lstm} architecture \cite<also used in>[for video captioning]{Venugopalan2015,Venugopalan2015a} where the next word is predicted as a function of both previous words and image features. In one version of the architecture, they inject {\sc cnn} features into the {\sc lstm} at each time-step. In a second version, they use two stacked {\sc lstm}s, the first of which takes {\sc cnn} features and produces an output which constitutes the input to the next {\sc lstm} to predict the word. Finally, \citeA{Mao2015} experiment with various {\sc mrnn} configurations, obtaining their best results with an architecture in which there are two word embedding layers preceding the recurrent layer, which is in turn projected into a multimodal layer where linguistic features are combined with {\sc cnn} features. An example caption is shown in Figure \ref{fig:mao2015} above.

These neural network models shed light on the consequences of combining the two modalities at different stages, reflecting the point made by \citeA[cf. Section \ref{sec:deep-learning}]{Manning2015} that this paradigm encourages a focus on architectures and design. In particular, image features can be used to bias the recurrent, language generation layer  -- at the start, or at each time-step of the {\sc rnn} -- as in the work of \citeA{Donahue2015}. Alternatively, the image features can be combined with linguistic features at a stage following the {\sc rnn}, as in the work of \citeA{Mao2015}.

\subsection{Vision and Language: Current and Future Directions for NLG}
Image to text generation is one area of {\sc nlg} where there is a clear dominance of deep learning methods. Current work focusses on a number of themes:

\begin{enumerate}
\item Generalising beyond training data is still a challenge, as shown by the work of \citeA{Devlin2015a}.
 More generally, dealing with novel images remains difficult, though experiments have been performed on using out-of-domain training data to expand vocabulary \cite{Ordonez2013}, learn novel concepts \cite{Mao2015a} or transfer features from image regions containing known labels, to similar, but previously unattested ones \cite[from which an example caption is shown in Figure \ref{fig:hendricks2016}]{Hendricks2016}.
Progress in zero-shot learning, where the aim is to identify or categorise images for which little or no training data is available, is likely to contribute to the resolution of data sparseness problems \cite<e.g.>{Antol2014,Elhoseiny2015}.

\item Attention is also being paid to what \citeA{Barnard2016} refers to as {\em localisation}, that is, the association of linguistic expressions with parts of images, and the ability to generate descriptions of specific image regions. Recent work includes that of \citeA{Karpathy2015}, \citeA{Johnson2016} and \citeA{Mao2016}, who focus on unambiguous descriptions of specific image regions and/or objects in images (see Section \ref{sec:reg} above for some related work). Attention-based models  are a further development on this front. These have been exploited in various {\sc seq2seq} tasks, notably for machine translation \cite{Bahdanau2015}. In the case of image captioning, the idea is to allocate variable weights to portions of captions in the training data, depending on the current context, to reflect the `relevance' of a word given previous words and an image region \cite{Xu2015}.
 
\item Recent work has also begun to explore generation from images that goes beyond the concrete conceptual, for instance, 
producing explanatory descriptions \cite{Hendricks2016a}. A further development is work on Visual Question Answering, where rather than descriptive captions, the aim is to produce responses to specific questions about images \cite{Antol2015,Geman2015,Malinowski2015,Barnard2016,mostafazadeh2016}. Recently, a new dataset was proposed providing both concrete conceptual and `narrative' texts coupled with images \cite{Huang2016}, a promising new direction for this branch of {\sc nlg}. 

\item There is a growing body of work that generalises the task from static inputs to sequential ones, especially videos \cite<e.g.>{Kojima2002,Regneri2013,Venugopalan2015,Venugopalan2015a}. Here, the challenges include handling temporal dependencies between scenes, but also dealing with redundancy.
\end{enumerate}
\section{Variation: Generating Text with Style, Personality and Affect}\label{sec:style}
Based on the preceding sections, the reader could be excused for thinking that {\sc nlg} is mostly concerned with delivering factual information, whether this is in the form of a summary of weather data, or a description of an image. This bias was also flagged in the Introduction, where we gave a brief overview of some domains of application, and noted that informing was often, though not always, the goal in {\sc nlg}.
 
Over the past decade or so, however, there has been a growing trend in the {\sc nlg} literature to also focus on aspects of textual information delivery that are arguably non-propositional, that is, features of text that are not strictly speaking grounded in the input data, but are related to the manner of delivery. In this section, we focus on these trends, starting with the broad concept of `stylistic variation', before turning to generation of affective text and politeness.

\subsection{Generating with Style: Textual Variation and Personality}
What does the term `linguistic style' refer to?
Most work on what we shall refer to as `stylistic {\sc nlg}' shies away from a rigorous definition, preferring to operationalise the notion in the terms most relevant to the problem at hand. 

`Style' is usually understood to refer to features of lexis, grammar and semantics that collectively contribute to the identifiability of an instance of language use as pertaining to a specific author, or to a specific situation (thus, one distinguishes between levels of stylistic formality, or speaks of the distinctive characteristics of the style of William Faulkner). This implies that any investigation of style must concern itself, at least in part, with {\em variation} among the features that mark such authorial or situational variables. In line with this usage, this section reviews developments in {\sc nlg} in which variation is the key concern, usually at the {\em tactical}, rather than the strategic, level, the idea being that a given piece of information can be imparted in  linguistically distinct, ways \cite<cf.>{Sluis2010}. This strategy was, for example, explicitly adopted by \citeA{Power2003}. 

Given its emphasis on linguistic features, controlling style (however it is defined) is a problem of great interest for {\sc nlg} since it directly addresses issues of {\em choice}, which are arguably the hallmark of any {\sc nlg} system \cite<cf.>{Reiter2010}. Early contributions in this area defined stylistic features using rules to vary generation according to pragmatic or stylistic goals. For example, \citeA{McDonald1985} argued that ``prose style is a consequence of what decisions are made during the transition from the conceptual representation level to the linguistic level'' (p. 61), thereby placing the problem within the domain of sentence planning and realisation. This stance was also adopted by \citeA{Dimarco1993}, who focus on syntactic variation, proposing a stylistic grammar for English and French. \citeA{Sheikha2011} proposed an adaptation of the SimpleNLG realiser \cite{Gatt2009} to handle formal versus informal language, via specific features, such as contractions ({\em are not} vs. {\em aren't}) and lexical choice.

A related perspective on stylistic variation was adopted by \citeA{Walker2002}, in their description of how the {\sc sp}o{\sc t} sentence planner was adapted to learn strategies for different communicative goals, as reflected in the rhetorical and syntactic structures of the sentence plans. The planner was trained using a boosting technique to learn correlations between features of sentence plans and human ratings of the adequacy of a sample of outputs for different communicative goals.

Like \citeA{Walker2002}, contemporary approaches to stylistic variation have tended to eschew rules in favour of data-driven methods to identify relevant features and dimensions of variation from corpora, in what might be thought of as an {\em inductive} view of style, where variation is characterised by the distribution of whatever linguistic features are considered relevant. An important precedent for this view is Biber's corpus-based multidimensional approach to style and register variation \cite{Biber1988}, roughly a contemporary of the grammar-inspired approach of \citeA{Dimarco1993}. 

Biber's model was at the heart of work by \citeA{Paiva2005}, which exhibits some characteristics in common with the `global' statistical approaches to {\sc nlg} discussed in Section \ref{sec:datadriven}, insofar as it exploits statistics to inform decision-making at the relevant choice points, rather than to filter the outputs of an overgeneration module. 
\citeA{Paiva2005} used a corpus of patient information leaflets, conducting factor analysis on their linguistic features to identify two stylistic dimensions. They then allowed their system to generate a large number of texts, varying its decisions at a number of choice points (e.g. choosing a pronoun versus a full {\sc np}) and maintaining a trace. Texts were then scored on the two stylistic dimensions, and a linear regression model was developed to predict the score on a dimension based on the choices made by the system. This model was used during testing to predict the best choice at each choice point, given a desired style.
Style, however, is a global feature of a text, though it supervenes on local decisions. These authors solved the problem by using a best-first search algorithm to identify the {\em series} of local decisions as scored by the linear models, that was most likely to maximise the desired stylistic effect, yielding variations such as the following \cite<examples from>[p. 61]{Paiva2005}: 

\begin{examples}
\item The dose of the patient's medicine is taken twice a day. It is two grams.
\item The two-gram dose of the patient's medicine is taken twice a day.
\item The patient takes the two-gram dose of the patient's medicine twice a day.
\end{examples}

Some authors \cite<e.g.,>[on which more below]{Mairesse2011} have noted that certain features, once selected, may `cancel' or obscure the stylistic effect of other features. This raises the question whether style can in fact be modelled as a linear, additive phenomenon, in which each feature contributes to an overall perception of style independently of others (modulo its weight in the regression equation). 

A second question is whether stylistic variation could be modelled in a more specific fashion, for example, by tailoring style to a specific author, rather than to generic dimensions related to `formality', `involvement' and so on. For instance a corpus-based analysis of human-written weather forecasts by \citeA{Reiter2005} found that lexical choice varies in part based on the author.
One line of work has investigated this using corpora of referring expressions, such as the {\sc tuna} Corpus \cite{Deemter2012}, in which multiple referring expressions by different authors are available for a given input domain. For instance, \citeA{Bohnet2008} and \citeA{DiFabbrizio2008} explore statistical methods to learn individual preferences for particular attributes, a strategy also used by \citeA{Viethen2010}. \citeA{Hervas2013} use case-based reasoning to inform lexical choice when realising a set of semantic attributes for a referring expression, where the case base  differentiates between authors in the corpus to take individual lexicalisation preferences into account \cite<see also>{Hervas2016}. 

A more ambitious view of individual variation is present in the work of \citeA{Mairesse2010,Mairesse2011}, in the context of {\sc nlg} for dialogue systems. Here, the aim is to vary the output of a generator so as to project different personality traits. Similar to the model of \citeA{Biber1988}, personality is here given a multidimensional definition, via the classic `Big 5' model  \cite<e.g.,>{John1999}, where personality is a combination of five major traits (e.g. introversion/extraversion).
However, while stylistic variation is usually defined as a linguistic phenomenon, the linguistic features of personality are only indirectly reflected in speaking or writing \cite<a hypothesis underlying much work on detection of personality and other features in text, including>{Oberlander2006,Argamon2007,Schwartz2013a,Youyou2015}.

\citeauthor{Mairesse2011}'s {\sc personage} system, originally based on rules derived from an exhaustive review of psychological literature \cite{Mairesse2010}, was developed in the restaurant domain. The subsequent, data-driven version of the system \cite{Mairesse2011} takes as input a pragmatic goal and, like the system of \citeA{Paiva2005}, a list of real-valued style parameters, this time representing scores on the five personality traits. The system estimates generation parameters for stylistic features based on the input traits, using machine-learned models acquired from a dataset pairing sample utterances with human personality judgements. For example, an utterance reflecting high extraversion might be more verbose and involve more use of expletives (\ref{ex:mairesse1}), compared to a more introverted style, which might demonstrate more uncertainty, for example through the use of stammering and hedging (\ref{ex:mairesse2}).

\begin{examples}
\item \label{ex:mairesse1} Kin Khao and Tossed are bloody outstanding. Kin Khao just has rude staff. Tossed features sort of unmannered waiters, even if the food is somewhat quite adequate.
\item \label{ex:mairesse2}Err... I am not really sure. Tossed offers kind of decent food.  Mmhm... However, Kin Khao, which has quite ad-ad-adequate food, is a thai place. You would probably enjoy these restaurants.
\end{examples}
 
An interesting outcome of the evaluation with human subjects reported by \citeA{Mairesse2011} is that readers vary significantly in their judgements of what personality is actually reflected by a given text. This suggests that the relationship between such psychological features and their linguistic effects is far from straightforward. \citeA{Walker2011:Arboretum} compared the `Big 5' model incorporated in the rule-based version of {\sc personage}, to a corpus-based model drawn from character utterances in film scripts. These models were used to generate utterances for characters in an augmented reality game; their main finding was that modelling characters' style directly using corpora of utterances results in more specific and easily perceived traits than using a model based on personality traits, where the relationship between personality and individual style is more indirect. In another set of experiments on generating utterances for characters in a role-playing game, \citeA{Walker2011:Film} report the successful porting of {\sc personage} to the new domain by tuning some of its parameters on features identified in film dialogue. Models learned from film corpora were found to be close in style to the characters they were actually based on.

\subsection{Generating with Feeling: Affect and Politeness}\label{sec:affect}
Personality is usually thought of in terms of {\em traits}, which are relatively stable across time. However, language use may vary not only across individuals, as a function of their stable characteristics, but also within individuals across time, as a function of their more transient affective {\em states}. 
`Affective {\sc nlg}' \cite<a term due to>{Rosis2000} is concerned with variation that reflects emotional states which, unlike personality traits, are relatively transient. In this case, the goals can be twofold:  (i) to induce an emotional state in the receiver; or (ii) to reflect the emotional state of the producer. 

As in the case of personality, the relationship between emotion and language is far from clear, as noted by \citeA{Belz2003}. For one thing, it isn't clear whether only surface linguistic choices need be affected. Some authors have argued that a text's affective 
impact impinges on content selection; this stance has been adopted, for example, in some applications in e-health where reporting of health-related issues should be sensitive to their potential emotional impact \cite{DiMarco2007,Mahamood2011}.

Most work on affective {\sc nlg} has however focussed on tactical choices \cite<e.g.>{Hovy1988,Fleischman2002,Strong2007,VanDeemter2008,Keshtkar2011}. Various linguistic features that can have emotional impact have been identified, from the increased use of redundancy to enhance understanding of emotionally laden messages \cite{Walker1995,Rosis2000}, to the increased use of first-person pronouns and adverbs, as well as sentence ordering to achieve emphasis or reduce adverse emotional impact \cite{Rosis2000}. 

This research on affective {\sc nlg} relies on models of emotion of various degrees of complexity and cognitive plausibility. The common trend underlying all these approaches however is that emotional states should impact lexical, syntactic and other linguistic choices. The question then is to what extent such choices are actually perceived by readers or users of a system.

In an empirical study, \citeA{Sluis2010} reported on two experiments investigating the effect of various tactical decisions on the emotional impact of text on readers. In one experiment, texts gave a (fake) report to participants on their performance on an aptitude test, with manually induced variations, such as these:

\begin{examples}
\item {\em Positive slant}: On top of this you also outperformed most people in your age group with your exceptional scores for Imagination and Creativity (7.9 vs 7.2) and Logical- Mathematical Intelligence (7.1 vs. 6.5).  
\item  {\em Neutral/factual slant}: You did better than most people in your age group with your scores for Imagination and Creativity (7.9 vs 7.2) and Logical-Mathematical Intelligence (7.1 vs. 6.5). 
\end{examples}

Evaluation of these texts showed that the extent to which affective tactical decisions influence hearer's emotional states is dependent on a host of other factors, including the degree to which the reader is directly implicated in what the text says (in the case of an aptitude test, the reader would be assumed to feel the outcomes have personal relevance).
An important question raised by this study is how affect should be measured: \citeA{Sluis2010} used a standardised self-rating questionnaire to estimate changes in affect before and after reading a text, but the best way to measure emotion remains an open question. 

The emotional slant in the language used by an author or speaker may have implications for the degree to which the listener or reader may feel `impinged upon'. This becomes particularly relevant in interactive systems, where {\sc nlg} components are generating language in the context of dialogue. Consider, for example, the difference between these requests:

\begin{examples}
\item\label{ex:direct}{\em Direct strategy}: Chop the tomatoes! 
\item\label{ex:approval}{\em Approval strategy}: Would it be possible for you to chop the tomatoes? 
\item\label{ex:autonomy}{\em Autonomy strategy}: Could you possibly chop the tomatoes? 
\item\label{ex:indirect}{\em Indirect strategy}: The tomatoes aren't chopped yet. 
\end{examples}

The four strategies exemplified above come across as having varying degrees of politeness which, according to one influential account \cite{BrownLevinson1987}, depends on {\em face}. {\em Positive face} reflects the speaker's desire that some of her goals be shared with her interlocutors; {\em negative face} refers to the speaker's desire not to have her goals impinged upon by others. 
The connection with affect that we suggested above hinges on these distinctions: different degrees of politeness reflect different degrees of `threat' to the listener; hence, generating language based on the right face strategy could be seen as a branch of affective {\sc nlg}.

In an early, influential proposal, \citeA{Walker1997a} proposed an interpretation of the framework of \citeA{BrownLevinson1987} in terms of the four dialogue strategies exemplified in (\ref{ex:direct} -- \ref{ex:indirect}) above. Subsequently, \citeA{Moore2004} used this framework in the generation of tutorial feedback, where a discourse planner used a Bayesian network to inform linguistic choices compatible with the target politeness/affect value in a given context \cite<see>[for a related approach]{Johnson2004}. 

\citeA{Gupta2007} also used the four dialogue strategies identified by \citeA{Walker1997a} in the {\sc p}o{\sc lly} system, which used {\sc strips}-based planning to generate a plan distributed among two agents in a collaborative task \cite<see also>{Gupta2008}. An interesting finding in their evaluation is that perception of face-threat depends on the speech act; for example, requests can be more threatening. \citeA{Gupta2007} also note possible cultural differences in perception of face threat (in this case, between {\sc uk} and Indian participants).

\subsection{Stylistic Control as a Challenge for Neural {\sc nlg}}
In the past few years, stylistic -- and especially affective -- {\sc nlg} has witnessed renewed interest by researchers working on neural approaches to generation. The trends that can be observed here mirror those outlined in our general overview of deep learning approaches (Section \ref{sec:deep-learning}).

A number of models focus on {\em response generation} (in the context of dialogue, or social media exchanges), where the task is to generate a response, given an utterance. Thus, these models fit well within the {\em seq2seq} or Encoder-Decoder framework (see Section \ref{sec:deep-learning}  for discussion). Often, these models exploit social media data, especially from Twitter, a trend which goes back at least to \citeA{Ritter2011}, who adapted a Phrase-Based Machine Translation model to response generation. For example \citeA{Li2016} proposed a persona-based model in which the decoder {\sc lstm} is conditioned on embeddings obtained from tweets pertaining to individual speakers/authors. An alternative model conditions on both speaker and addressee profiles, with a view to incorporating not only the `persona' of the generator, but its variability with respect to different interlocutors. \citeA{Herzig2017}, also working on Twitter data, condition their decoder on personality features extracted from tweets based on the `Big Five' model, rather than on speaker-specific embeddings. This has the advantage of not enabling the generator to be tuned to specific personality settings, without re-training to adapt to a particular speaker style. While their personality-based model does not beat \citeauthor{Li2016}'s model, a human evaluation showed that judges were able to identify high-trait responses as more expressive than low-trait responses, suggesting that the conditioning was having a noticeable impact on style. In a dialogue context, \citeA{Ashgar2017} proposed to achieve affective responses on three levels: (a) by augmenting word embeddings with data from an affective dictionary; (b) by decoding with an affect-sensitive beam search; and (c) by training with an affect-sensitive loss function.

On the other hand, a number of models condition an {\sc lstm} on attributes reflecting affective or personality traits, with a view to generating strings that express such traits. \citeA{Ghosh2017} used {\sc lstm}s trained on speech corpora conditioned on affect category and emotional intensity to drive lexical choice. \citeA{Hu2017} used variational auto-encoders and attribute discriminators, to control the stylistic parameters of generated texts individually. They experimented on controlling sentiment and tense, but restricted the generation to sentences of up to 16 words. By contrast, \citeA{Ficler2017} extend the range of parameters used to condition the {\sc lstm}, with two content-related attributes (sentiment and theme) and four stylistic parameters (length, whether the text is descriptive, whether it has a personal voice, and whether the style is professional). Their generator is trained on a corpus of movie reviews. Similarly, \citeA{Dong2017} propose an attribute-to-sequence model for product review generation based on a corpus of Amazon user reviews \cite<see also>[for neural models for product review generation]{Lipton2016,Tang2016}. The conditioning includes the reviewer {\sc id}, reminiscent of the persona-based response model of \citeA{Li2016}; however, they also include the rating, which functions to modulate the affect in the output. Their model incorporates an attentional mechanism to concentrate on different parts of the input encoding when predicting the next word during decoding. For example, for a specific reviewer and a specific product, changing the input rating from 1 to 5 yields the following difference:

\begin{examples}
\item (Rating: 1) i’m sorry to say this was a very boring book. i didn’t finish it. i’m not a new fan of the series, but this was a disappointment
\item (Rating: 5) this was a very good book. i enjoyed the characters and the story line. i’m looking forward to reading more in this series.
\end{examples}

\subsection{Style and Affect: Concluding Remarks}
Controlling stylistic, affective and personality-based variation in {\sc nlg} is still in a rather fledgling state, with several open questions of both theoretical and computational import. Among these is the question of how best to model complex, multi-dimensional constructs such as personality or emotion; this question speaks both to the cognitive plausibility of the models informing linguistic choices, and to the practical viability of different machine learning strategies that could be leveraged for the task (for example, linear, additive models versus more `global' models of personality or style). Also important here is the kind of data used to inform generation strategies: as we have seen above, a lot of affective {\sc nlg} work relies on ratings by human judges. However, some recent work in affective computing has questioned the use of ratings, comparing them to ranking-based and physiological methods \cite<e.g.>{Martinez2014,Yannakakis2015}. This and similar research is probably of high relevance to {\sc nlg} researchers. Some recent work relied on automatic extraction of personality features using tools such as {\sc ibm}'s Personality Insights \cite{Herzig2017}. As such tools \cite<another example of which is Lingustic Inquiry and Wordcount or {\sc liwc}, >{Pennebaker2007} become more reliable and widely available, we may see a turn towards less reliance on human elicitation.  

A second important question is which linguistic choices truly convey the intended variation to the reader or listener. While current systems use a range of devices, from aggregation strategies to lexical choice, it is not clear which ones are actually perceived as having the desired effect. 

A third important research avenue, which is especially relevant to interactive systems, is adaptivity, that is, the way speakers (or systems) alter their linguistic choices as a result of their interlocutors' utterances \cite{Clark1996a,Niederhoffer2002,Pickering2004}, a theme that has also begun to be explored in {\sc nlg} \cite{Isard2006,Herzig2017}.
\section{Generating Creative and Entertaining Text}\label{sec:creativity}
`Good' writers not only present their ideas in coherent and well-structured prose. They also succeed in keeping the attention of the reader through narrative techniques, and in occasionally surprising the reader, for example, through creative language use such as small jokes or well-placed metaphors \cite<see e.g., among many others, >{flower1981cognitive,nauman2011makes,veale2015distributed}. The {\sc nlg} techniques and applications discussed so far in this survey arguably do not simulate good writers in this sense, and as a result automatically generated texts can be perceived as somewhat boring and repetitive. 

This lack of attention to creative aspects of language production within {\sc nlg} is not due to a general lack of scholarly interest in these phenomena. Indeed, computational research into creativity has a long tradition, with roots that go back to the early days of {\sc ai} \cite<as>[notes, the first story generation algorithm on record, Novel Writer, was developed by Sheldon Klein in 1973]{Gervas2013}. However, it is fair to say that, so far, there has been  little interaction between researchers from the computational creativity and {\sc nlg} communities respectively, even though both groups in our opinion could learn a lot from each other. In particular, {\sc nlg} researchers stand to benefit from insights into what constitutes creative language production, as well as structural features of narrative that have the potential to improve {\sc nlg} output even in data-to-text systems \cite<see>[for an argument to this effect in relation to a medical text generation system]{Reiter2008}. At the same time, researchers in computational creativity could also benefit from the insights provided by the {\sc nlg} community where the generation of fluent language is concerned since, as we shall see, a lot of the focus in this research, especially where narrative is concerned, is on the generation of plans and on content determination. 

In what follows, we give an overview of automatic approaches to creative language production, starting from relatively simple jokes and metaphors to more advanced forms, such as narratives.

\subsection{Generating Puns and Jokes}\label{sec:puns}

Consider:

\begin{example}
What's the difference between money and a bottom?\\
{\it One you spare and bank, the other you bare and spank.}
\end{example}

\begin{example}
What do you call a weird market?\\
{\it A bizarre bazaar.}
\end{example}

These two (pretty good!) {\it punning riddles}\/ were automatically generated by the {\sc jape} system developed by \citeA{Binsted1994,Binsted1997a}. Punning riddles form a specific joke genre and have received considerable attention in the context of computational humor, presumably because they are relatively straightforward to define, often relying on spelling or word sense ambiguities. Many good, human-produced examples have been collected in joke books and sites and may thus act as a source of inspiration or training data.

Simplifying somewhat, {\sc jape} (Joke Analysis and Production Engine) relies on a template-based {\sc nlg} system, combining fixed text ({\em What's the difference between X and Y?} or {\em What do you call X?}) with slots, which are the source of the riddle. Various standard lexical resources are used for joke production, including a British pronunciation dictionary (to find different words with a similar pronunciation, such as `bizarre' and `bazaar') and WordNet \cite[to find words with a similar meaning, such as {\em bazaar} and {\em market}]{Miller1995}. {\sc jape} uses various techniques to create the punning riddles, such as juxtaposition, in which related words are simply placed next to each other and treated as a normal construction, while making sure that the combination is novel (i.e., not in the {\sc jape} database already). It is interesting to observe that in this way {\sc jape} may automatically come up with existing jokes (a quick Google search reveals that many bizarre bazaars, as well as bazaar bizarres, exist).

Following the seminal work of Binsted and Ritchie, various other systems have been developed which can automatically generate jokes, including for example the {\sc haha}cronym system of \citeA{Stock2005}, which produces humorous acronyms, and the system of \citeA{Binsted2003}, which focusses on the generation of referential jokes (``It was so cold, I saw a lawyer with his hands in his own pockets."). 

\citeA{petrovic2013unsupervised} offer an interesting, unsupervised alternative to this earlier work, which does not require labelled examples or hard-coded rules . Like their predecessors, \citeauthor{petrovic2013unsupervised} also start from a template -- in their case {\em I like my X like I like my Y, Z} -- where $X$ and $Y$ are nouns (e.g., {\em coffee} and {\em war}) and $Z$ is an attribute (e.g., {\em cold}). Clearly, linguistic realisation is not an issue, but content selection -- finding `funny' triples $X$, $Y$ and $Z$ --  is  a challenge. Interestingly, the authors postulate a number of guiding principles for `good' triples. In particular, they hypothesize that (a) the joke is funnier if the attribute $Z$ can be used to describe both nouns $X$ and $Y$; (b) the joke is funnier if attribute $Z$ is both common and ambiguous;and (c) the joke is funnier the more dissimilar $X$ and $Y$ are. These three statements can be quantified relying on standard resources such as Wordnet and the Google n-gram corpus \cite{Brants2006}, and using these measures their system outputs, for example:

\begin{example}
I like my relationships like I like my source, open.
\end{example}

It is probably fair to say that computational joke generation research to date has mostly focussed on laying bare the basic structure of certain relatively simple puns and exploiting these to good effect \cite<e.g.,>{Ritchie2009}. However, many other kinds of jokes exist, often requiring sophisticated, hypothetical reasoning. Presumably, many of the central problems within {\sc ai} need to be solved first before generation systems will be capable of producing these kinds of advanced jokes. 

\subsection{Generating Metaphors and Similes}\label{sec:metaphor}
Whether you think something is funny or not may be subjective, but in any case insights from joke generation can be useful as a stepping stone towards a better understanding of creative language use, including metaphor, simile and analogy. 
In all of these, a mapping is made between two conceptual domains, in such a way that terminology from the source domain is used to say something about the target domain, typically in a nonliteral fashion, which can be helpful in computer-generated texts to illustrate complex information. For example, \citeA{hervas2006cross} study analogies in narrative contexts, such as {\em Luke Skywalker was the King Arthur of the Jedi Knights}, which immediately clarifies an important aspect of Luke Skywalker for those not in the know. In a simile, the two domains are compared (A `is like' B); in a metaphor they are equated. Jokes and metaphors/similes are related: the automatically generated jokes of \citeauthor{petrovic2013unsupervised} are comparable to similes, while \citeA{kiddon2011thats}, for example, frame the problem of identifying double entendre jokes as a type of metaphor identification. Nevertheless, one could argue that generating jokes is more complex because of the extra funniness constraint. 

Like computational humor, the automatic recognition and interpretation of metaphorical, non-literal language has received considerable attention since the early days of {\sc ai} \cite<see>[for an overview]{Shutova2013}. \citeA{Martin1990,Martin1994}, for example, focussed on the recognition of metaphor in the context of Unix Support, as in the following examples:

\begin{examples}
\item How can I kill a process? 
\item How can I enter lisp? 
\end{examples}

The first one, for example, makes a mapping between `life' (source) and `processes' (target), and is by now so common that is almost a dead metaphor, but this was not the case in the early days of Unix. Clearly, understanding of the metaphors is a prerequisite for automatically answering these questions. Early research on the computational interpretation of metaphor already recognised that metaphors rely on semantic conventions that are exploited (`broken') to express new meanings. A system for metaphor understanding, as well as one for metaphor generation, therefore requires knowledge about what literal meanings are, and how these can be stretched or translated into metaphoric meanings \cite<e.g.,>{Wilks1978,Fass1991}.

Recent work by Veale and Hao \citeyear{Veale2007,Veale2008} has shown that this kind of knowledge can be acquired from the web, and used for the generation of new metaphors and similes (comparisons). Their system, called Sardonicus, is capable of generating  metaphors for user-provided targets ({\sc t}), such as the following, expressing that Paris Hilton (``the person, not the hotel, though the distinction is lost on Sardonicus'', Veale \& Hao, 2007, p. 1474) is skinny:

\begin{example}
Paris Hilton is a stick
\end{example}

Sardonicus searches the web for nouns ({\sc n}) that are associated with skinniness, which are included in a case-base and range from {\em pole, pencil}, and {\em stick} to {\em snake} and {\em stick insect}. Inappropriate ones (like {\em cadaver}) are ruled out, based on the theory of category-inclusion of \citeA{Glucksberg2001}. This list of potential similes is then used to create Google queries, inspired by the work of \citeA{Hearst1992}, of the form {\em {\sc n}-like {\sc t}} (e.g., {\em stick insect-like Paris Hilton}, which actually occurs on the web), giving a ranking of the potential similes to be generated.

A comparable technique is used by \citeA{Veale2013} to generate metaphors with an affective component, as in `Steve Jobs was a great leader, but he could be such a tyrant'. The Google $n$-gram corpus  is used to find stereotypes suitable for simile generation (e.g., `lonesome as a cowboy'), a strategy reminiscent of the use of web-scale $n-$gram data to smooth the output of image-to-text systems (see Section\ref{sec:image}). Next, an affective dimension is added, based on the assumption that properties occurring in a conjunction (`as {\it lush and green}\/ as a jungle') are more likely to have the same affect than properties that do not. Using positive (e.g., `happy', `wonderful') and negative (e.g., `sad', `evil') seeds, coordination queries (e.g., `happy and X') are used to collect positive and negative labels for stereotypes, indicating, for instance, that babies are positively associated with qualities such as `smiling' and `cute', and negatively associated with `crying' and `sniveling'. This enables the automatic generation of positive (`cute as a baby') and negative (`crying like a baby') similes. Veale even points out that by collecting, for example, a number of negative metaphors for Microsoft being a monopoly, and using these in a set of predefined tropes, it becomes possible to automatically generate a poem such as the following:

\begin{quote}
{\bf No Monopoly Is More Ruthless}\\
Intimidate me with your imposing hegemony \\
No crime family is more badly organized, or controls more ruthlessly\\
Haunt me with your centralized organization\\
Let your privileged security support me\\
O Microsoft, you oppress me with your corrupt reign
\end{quote}

In fact, automatic generation of poetry is an emerging area at the crossroads of computational creativity and natural language generation \cite<see for example>[for variations on this theme]{Lutz1959,Gervas2001,Wong2008,Netzer2009,Greene2010,Colton2012,Manurung2012,zhang2014chinese}. See the recent review by \citeA{Oliveira2017}.

\subsection{Generating Narratives}\label{sec:narrative}
Computational narratology is concerned with computational models for the generation and interpretation of narrative texts \cite<e.g.,>{Gervas2009,Mani2010,Mani2013}. The starting point for many approaches to narrative generation is a view of narrative coming from classical narratology, a branch of literary studies with roots in the Formalist and Structuralist traditions \cite<e.g.,>{Propp1968,Genette1980,Bal2009}. This field has been concerned with analysing both the defining characteristics of narrative, such as plot or character, and more subtle features, such as the handling of time and temporal shifts, focalisation (that is, the ability to convey to the reader that a story is being recounted from a specific point of view), and the interaction of multiple narrative threads, in the form of sub-plots, parallel narratives, etc. An important recent development is the interest, on the part of narratologists, in bringing to bear insights from Cognitive Science and {\sc ai} on their literary work, making this field ripe for multi-disciplinary interaction  \cite<see especially>[for programmatic statements to this effect, as well as theoretical contributions]{Herman1997,Herman2007,Meister2003}.

Classical narratology makes a fundamental distinction between the `story world' and the text that narrates the story. In line with the formalist and structuralist roots of this tradition, the distinction is usually articulated as a dichotomy between {\em fabula} (or story) and {\em suzjet} (or discourse). There is a parallel between this distinction and that between a text plan in {\sc nlg}, versus the actual text which articulates that plan. However, the crucial difference is that in producing a plan for a narrative, a story generation system typically does not use input data of the sort required by most of the {\sc nlg} systems reviewed thus far, since the story is usually fictional. On the other hand, narratological tools have also been successfully applied to real-world narratives, including oral narratives of personal experience \cite<e.g.,>{Herman2001,Labov2010}.  

The focus of most work on narrative generation has been on the pre-linguistic stage, that is, on generating plans within a story world for fictional narratives, usually within a specific genre whose structural properties are well-understood, for example, fairy tales or Arthurian legends \cite<see>[for a review]{Gervas2013}. There are however links between the techniques used for such stories and those we have discussed above in relation to {\sc nlg} (see especially Section \ref{sec:planning}). Prominent among these are planning and reasoning techniques to model the creative process as a problem-solving task. For example, {\sc minstrel} \cite{Turner1992} uses reasoning to model creativity from the author's perspective, producing narrative plans based on authorial goals, such as the goal of introducing drama into a narrative, while ensuring thematic consistency.

 More recently, {\sc brutus} \cite{Bringsjord1999} used a knowledge base of story schemas, from which one is selected and elaborated using planning techniques to link causes and effects \cite<see also>[among others, for recent examples of the use of planning techniques to model the creative process in narrative generation]{Young2008,Riedl2010}.

\begin{figure}[!h]
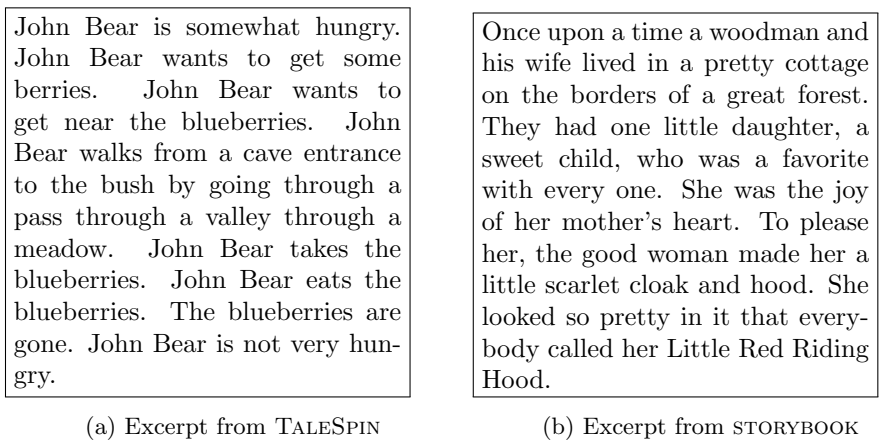

\begin{subfigure}[b]{0.5\textwidth} 
\fbox{\begin{minipage}{0.85\textwidth}
John Bear is somewhat hungry. John Bear wants to get some berries. John Bear wants to get near the blueberries. John Bear walks from a cave entrance to the bush by going through a pass through a valley through a meadow. John Bear takes the blueberries. John Bear eats the blueberries. The blueberries are gone. John Bear is not very hungry.
\end{minipage}}
\caption{Excerpt from {\sc TaleSpin}} 
\label{fig:talespin} 
\end{subfigure}
~\hfill
\begin{subfigure}[b]{0.5\textwidth} 
\fbox{\begin{minipage}{0.85\textwidth}
Once upon a time a woodman and his wife lived in a
pretty cottage on the borders of a great forest. They had one little daughter, a sweet child, who was a favorite with every one. She was the joy of her mother's heart. To please her, the good woman made her a little scarlet cloak and hood. She looked so pretty in it that everybody called her Little Red Riding Hood.
\end{minipage}}
\caption{Excerpt from {\sc storybook}} 
\label{fig:storybook} 
\end{subfigure}
\caption{Examples of automatically generated narratives. The left panel shows an excerpt from a story produced by {\sc TaleSpin} \cite{Meehan1977}; the right panel is an excerpt from the {\em Little Red Riding Hood} fairy-tale, generated by the {\sc storybook} system \cite{Callaway2002}.}
\end{figure}

As \citeA{Gervas2010} notes, the focus on planning story worlds and modelling creativity has often implied a sidelining of linguistic issues, so that rendering a story plan into text has often been viewed as a secondary consideration. For example Figure \ref{fig:talespin} shows an excerpt of a story produced by the {\sc talespin} system \cite{Meehan1977}: here, the emphasis is on using problem-solving techniques to produce a narrative in which events follow from each other in a coherent fashion, rather than on telling it in a fluent way. An important exception to this trend is the work of \citeA{Callaway2002}, who explicitly addressed the gap between computational narratology and {\sc nlg}. Their system took a narrative plan as a starting point, but focussed on the process of rendering the narrative in fluent English, handling time shifts, aggregation, anaphoric {\sc np}s and many other linguistic phenomena, as the excerpt in Figure \ref{fig:storybook} shows. It is worth noting that this system has since been re-used in the context of generating interactive text for a portable museum guide by \citeA{Stock2007}. 

In addition, there have been a number of contributions from the generation community on more specific issues related to narrative, such as how to convey the temporal flow of narrative discourse \cite{Oberlander1992,Dorr1995,Elson2010}. This is a problem that deserves more attention in {\sc nlg}, since texts with a complex narrative structure often narrate events in a different order from which they occurred. For example, a narrative or narrative-like text may recount events in order of importance rather than in temporal order, even when they are grounded in real-world data \cite<e.g.>{Portet2009}. This makes the use of the right choices for tense, aspect and temporal adverbials crucial to ensure clarity for the reader. This type of complexity in narrative structure also emerges in interactive narrative fiction \cite<for example, in games; cf.,>{montfort2007ordering}.

Beyond the focus on specific linguistic issues, there has also been some work that leverages data-driven techniques to generate stories. For example, \citeA{Mcintyre2009} propose a story generation system whose input is a database of entities and their interactions, extracted from a corpus of stories by parsing them, retrieving grammatical dependencies, and building chains of events in which specific entities play a role. The outcome is a graph encoding a partial order of events, with edges weighted by mutual information to reflect the degree of association between nodes. Sentence planning then takes place using template-like grammar rules specifying verbs with subcategorisation information, followed by realisation using {\sc realpro} \cite{Lavoie1997}. One of the most interesting features of this work is the coupling of the generation model with an interest model to predict which stories would actually be rated as interesting by readers. This was achieved by training a kernel-based classifier on shallow lexical and syntactic features of stories, a novel take on an old problem in narratology, namely, what makes a story `tellable', thereby distinguishing it from a mere report \cite<e.g.,>{Herman1997,Norrick2005,Bruner2011}.

Most story generation work is restricted to (very) short stories. It is certainly true that planning a book-length narrative along the lines sketched above is extremely challenging, but researchers have recently started exploring the possibilities, for instance in the context of  NaNoGenMon (National Novel Generation Month), in which participants write a computer program capable of generating a 'novel'. Perhaps the best known example is {\it World Clock}\/ \cite{montfort2013world} which describes 1440 (24 $\times$ 60) events taking place around the world, one randomly selected minute at a time. These are the first two:

\begin{quote}
It is now exactly 05:00 in Samarkand. In some ramshackle dwelling a person who is called Gang, who is on the small side, reads an entirely made-up word on a box of breakfast cereal. He turns entirely around.

It is now right about 18:01 in Matamoros. In some dim yet decent structure a man named Tao, who is no larger or smaller than one would expect, reads a tiny numeric code from a recipe clipping. He smiles a tiny smile.
\end{quote}

The book was fully generated by 165 lines of Python code, written by the author in a few hours, and later published (together with the software) by Harvard Book Store press. There is even a Polish translation (by Piotr Marecki), created by translating the terms and phrases used in the Python implementation of the original algorithm.

\subsection{Generating Creative Language: Concluding Remarks}
In this section we have highlighted recent developments in the broad area of creative language generation, a topic which is rather understudied in {\sc nlg}. Nevertheless, we would like to argue that {\sc nlg} researchers can improve the quality of their output by taking insights from computational creativity on board.

Work that exploits corpora and other lexical resources for the automatic generation of jokes, puns, metaphors and similes has revealed different ways in which words are related and can be juxtaposed to form unexpected and possibly even `funny'  or `poetic' combinations. Given that, for example, metaphor is pervasive in everyday language \cite<as argued, for example, by>{Lakoff1980}, not just in overtly creative uses,  {\sc nlg} researchers interested in enhancing the readability -- and especially the variability -- of the text-generating capability of their models would benefit from a closer look at work in poetry, joke and metaphor generation.

In a similar vein, work on narratology is rich in insights on the interaction of multiple threads in a single narrative, and how the choice of events and their ordering can give rise to interesting stories \cite<e.g.,>{Gervas2012}. These insights are valuable, for example, in the development of more elaborate text planners in domains where time and causality play a role. Similarly, narratological work on character and focalisation can also help in the development of better {\sc nlg} techniques to vary output according to specific points of view, an area that we touched on in Section \ref{sec:style}, 

We have deferred discussion of evaluation of creative {\sc nlg} to Section \ref{sec:evaluation}, which deals with evaluation in general. 
Anticipating some of that discussion, it is worth noting that evaluation of creative language generation remains something of a bottleneck. In part, this is because it is not always easy to determine the `right' question to ask in an evaluation of creative text. For instance, in the case of joke and poetry generators, demonstrating genre compatibility and recognition (`Is this a joke?') is arguably already an achievement, insofar as it suggests that a system is producing artefacts that conform to normative expectations (this is discussed further in Section \ref{sec:genre-eval} below). In other types of creative language generation, evaluation is more challenging because it is difficult to carry out without ensuring quality at {\em all} levels of the generation process, from planning to realisation. In the case of narrative generation, for example, if the emphasis is placed entirely on story planning, the perceived quality of the narrative will be compromised if story plans are rendered using a an excessively simple realisation strategy (as is the case in Figure \ref{fig:talespin}). This is an area where the consensus in the field is that much further research effort is required \cite<see>[for a recent argument to this effect]{Zhu2012}. It is also an area in which {\sc nlg} can potentially offer much to computational creativity researchers, including in the use of techniques to render text fluently and consistently, facilitating the evaluation of generated artefacts with human subjects.

\section{Evaluation}\label{sec:evaluation}
Though we have touched on the subject of evaluation at various points, it deserves a full discussion as a topic which has become a central methodological concern in {\sc nlg}. A factor that contributed to this development was the establishment of a number of {\sc nlg} shared tasks, launched in the wake of an {\sc nsf-}funded workshop held in Virginia in 2007 \cite{Dale2007}. These tasks have focussed on referring expression generation \cite{Belz2010,Gatt2010}; surface realisation \cite{Belz2011}; generation of instructions in virtual environments \cite{Striegnitz2011,Janarthanam2011}; content determination \cite{BouayadAgha2013,Banik2013}; and question generation \cite{Rus2011}. Recent proposals for new challenges extend these to narrative generation \cite{concepcion-EtAl:2016:INLG}, generation from structured web data \cite{colin-EtAl:2016:INLG}, and from pairs of meaning representations and text  \cite{novikova-rieser:2016:INLG,May2017}. In image captioning, shared tasks have helped the development of large-scale datasets and evaluation servers such as {\sc ms-coco}\footnote{\url{http://mscoco.org/dataset/\#captions-upload}} (cf. Section \ref{sec:im-data}). 

In general, however, {\sc nlg} evaluation is marked by a great deal of variety and it is difficult to compare systems directly. There are at least two reasons why this is the case.

\paragraph{Variable input} There is no single, agreed-upon input format for {\sc nlg} systems \cite{McDonald1993,Mellish1998a,evans2002nlg}. Typically, one can only compare systems against a common benchmark if the input is similar. Examples are the image-captioning systems described in Section \ref{sec:image}, or systems submitted to one of the shared tasks mentioned above. Even in case a common `standard' dataset is available for evaluation, comparison may not be straightforward due to input variation, or due to implicit biases in the input data. For example, \citeA{Rajkumar2014} observe that, despite many realisers being evaluated against the Penn Treebank, they make different assumptions about the input format, including how detailed the pre-syntactic input representation is, a problem also  observed in the first Surface Realisation shared task \cite{Belz2011}.  As \citeA{Rajkumar2014} note, a comparison of realisers on the basis of scores on the Penn Treebank shows that the highest-ranking is the {\sc fuf/surge} realiser (which is second in terms of coverage), based on experiments by \citeA{Callaway2005}. However, these experiments required painstaking effort to extract the input representations at the level of detail needed by {\sc fuf/surge}; other realisers support more underspecified input. In a related vein, image captioning evaluation studies have shown that many datasets contain a higher proportion of nouns than verbs, and few abstract concepts \cite{Ferraro2015}, making systems that generate descriptions emphasising objects more likely to score better. The relevance of this observation is shown by \citeA{Elliott2015}, who note that the ranking of their image captioning system based on visual dependency grammar depends in part on the data it is evaluated on, with better performance on data containing more images depicting actions (we return to this study below).  

\begin{figure}[t]
\centering
\includegraphics[width=\textwidth]{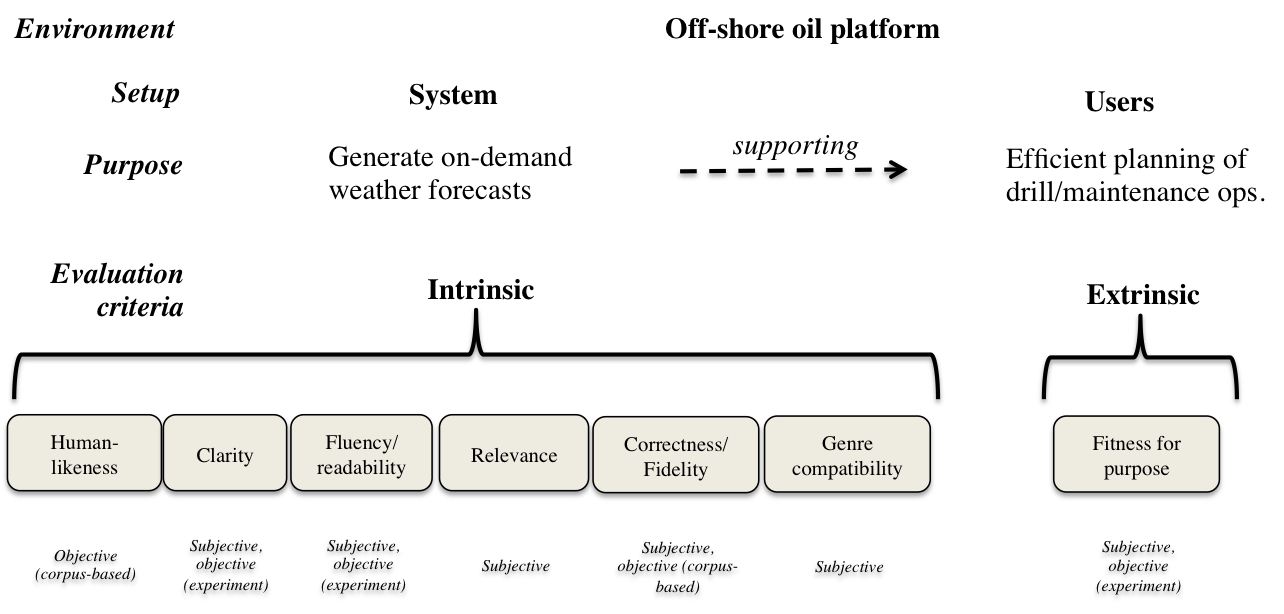}
\caption{Hypothetical evaluation scenario: a weather report generation system embedded in an offshore oil platform environment. Possible evaluation methods, focussing on different questions, are highlighted at the bottom, together with the typical methodological orientation (subjective/objective) adopted to address them.}
\label{fig:eval-scenario}
\end{figure}

\paragraph{Multiple possible outputs} Even for a single piece of input and a single system, the range of possible outputs is open-ended, a problem that arguably holds for any {\sc nlp} task involving textual output, including machine translation and summarisation. Corpora often display a substantial range of variation and it is often unclear, without an independent assessment, which outputs are to be preferred \cite{ReiterSripada2002}. In the image captioning literature, authors who have framed the problem in terms of retrieval have motivated the choice in part based on this problem, arguing that `since there is no consensus on what constitutes a good image description, independently obtained human assessments of different caption generation systems should not be compared directly' \cite[p. 580]{Hodosh2013}. While capturing variation may itself be a goal \cite<e.g.,>{Belz2008,Viethen2010,Hervas2013,castro2016towards}, as we also saw in our discussion of style in Section \ref{sec:style}, this is not always the case. Thus, in a user-oriented evaluation, the S{\sc um}T{\sc ime}-{\sc mousam} system weather forecasts were preferred by readers over those written by forecasters because the latter's lexicalisation decisions were susceptible to apparently arbitrary variation \cite{Reiter2005}; similar outcomes were more recently reported for statistical {\sc nlg} systems trained on the S{\sc um}T{\sc ime} corpus \cite{Belz2008,Angeli2010}.\\ 
\ \\ 
Rather than give an exhaustive review of {\sc nlg} evaluation -- hardly a realistic prospect given the diversity we have pointed out -- the rest of this section will highlight some topical issues in current work. By way of an overview of these issues, consider the hypothetical scenario sketched in Figure \ref{fig:eval-scenario}, which is loosely inspired by work on various weather-reporting systems developed in the field. This {\sc nlg} system is embedded in the environment of an offshore oil-rig; the relevant features of the {\em setup} \cite<in the sense of>{SparckJones1996} are the system itself and its users, here a group of engineers. While the {\em task} of the system is to generate weather reports from numerical weather prediction data, its ultimate {\em purpose} is to facilitate users' planning of drilling and maintenance operations.
Figure \ref{fig:eval-scenario} highlights some of the common questions addressed in {\sc nlg} evaluation, together with a broad typology of the methods used to address them, in particular, whether they are objective -- that is measurable against an external criterion, such as corpus similarity or experimentally obtained behavioural data -- or subjective, requiring human judgements. 

A fundamental methodological distinction, due to \citeA{SparckJones1996}, is between {\em intrinsic} and {\em extrinsic} evaluation methods. In the case of {\sc nlg}, an intrinsic evaluation measures the performance of a system without reference to other aspects of the setup, such as the system's effectiveness in relation to its users. In our example scenario, questions related to text quality, correctness of output and readability qualify as intrinsic, whereas the question of whether the system actually achieves its goal in supporting adequate decision-making on the offshore platform is extrinsic.

\subsection{Intrinsic Methods}\label{sec:intrinsic}
Intrinsic evaluation in {\sc nlg} is dominated by two methodologies, one relying on human judgements (and hence subjective), the other on corpora. 

\subsubsection{Subjective (Human) Judgements}
Human judgements are typically elicited by exposing naive or expert subjects to system outputs and getting them to rate them on some criteria. Common criteria include:

\begin{itemize}
\item {\em Fluency} or {\em readability}, that is, the linguistic quality of the text \cite<e.g.,>[{\em inter alia}]{Callaway2002,Mitchell2012,Stent2005a,Lapata2006,Cahill2009,Espinosa2010};
\item {\em Accuracy}, {\em adequacy}, {\em relevance} or {\em correctness} relative to the input, reflecting the system's rendition of the content \cite<e.g.>{Lester1997,Sripada2005,Hunter2012}, a criterion often used in subjective evaluations of image-captioning systems as well \cite<e.g.>{Kulkarni2011,Mitchell2012,Kuznetsova2012,Elliott2013}.
\end{itemize}

Though they are the most common, these two sets of criteria do not exhaust the possibilities. For example, subjective ratings have also been elicited for argument effectiveness in a system designed to generate persuasive text for prospective house buyers \cite{Carenini2006}. In image captioning, at least one system was evaluated by asking users to judge the creativity of the generated caption, with a view to assessing the contribution of web-scale n-gram language models to the captioning quality \cite{Li2011}. Below, we also discuss judgements of genre compatibility (Section \ref{sec:genre-eval}). In the case of fictional narrative, some evaluations have elicited judgments on qualities such as novelty \cite<e.g.,>{Perez2011} or believability of characters \cite<e.g.,>{Riedl2005a}.

The use of scales to elicit judgements raises a number of questions. One has to do with the nature of the scale itself. While discrete, ordinal scales are the dominant method, a continuous scale -- for example, one involving a visually presented slider \cite{Gatt2010,Belz2011a} -- might give subjects the possibility of giving more nuanced judgements. For example, a text generated by our hypothetical weather report system might be judged so disfluent as to be given the lowest rating on an ordinal scale; if the following text is judged as being worse, a subject would have no way of indicating this. A related question is whether subjects find it easier to {\em compare} items rather than judge each one in its own right. This question has begun to be addressed in the {\sc nlp} evaluation literature, usually with binary comparisons, for example between the outputs of two {\sc mt} systems \cite<see>[for discussion]{Dras2015}. In a recent study evaluating causal connectives produced by an {\sc nlg} system, \citeA{Siddharthan2012a} used Magnitude Estimation, whereby subjects are not given a predefined scale, but are asked to choose their own and proceed to make comparisons of each item to a `modulus', which serves as a comparison point throughout the experiment \cite<see>{Bard1996}.\footnote{The modulus is an item -- a text, or a sentence -- which is selected in advance and which subjects are asked to rate first. All subsequent ratings or judgements are performed in comparison to this modus item. Though subjects are able to use any scale they choose, this method allows all judgements to be normalised by the judgement given for the modulus. Typically, normalised judgements are analysed on a logarithmic scale.} \citeA{Belz2010a} compared a preference-based paradigm to a standard rating scale to evaluate systems from two different domains (weather reporting and {\sc reg}), and found that the former was more sensitive to differences between systems, and less susceptible to variance between subjects.

An additional concern with  subjective evaluations is inter-rater reliability. Multiple judgements by different evaluators may exhibit  high variance, a problem that was encountered in the case of Question Generation \cite{Rus2011}. Recently, \citeA{Godwin2016} suggested that such variance can be reduced by an iterative method whereby training of judges is followed by a period of discussion, leading to the updating of evaluation guidelines. This, however, is more costly in terms of time and resources. 

It is probably fair to state that, these days, subjective, human evaluations are often carried out via online platforms such as Amazon Mechanical Turk
and CrowdFlower,
though this is probably more feasible for widely-spoken languages such as English. A seldom-discussed issue with such platforms concerns their ethical implications \cite<for example, they involve large groups of poorly paid individuals; see>{Fort2011} as well as the reliability of the data collected, though measures can be put in place to ensure, for instance, that contributors are fluent in the target language \cite<see e.g.,>{goodman2013data,mason2012conducting}.

\afterpage{
\begin{table}[!h]
\small
\centering
\begin{tabular}{llp{9cm}l}
\hline 
& Metric & Description & Origins \\
\hline
\multirow{7}{*}[-60pt]{\rotatebox[origin=c]{90}{N-gram overlap}} &  {\sc bleu} & Precision score over variable-length n-grams, with a length penalty \cite{Papineni2002} and, optionally, smoothing \cite{Lin2004}. & {\sc mt}\\
& {\sc nist} & A version of {\sc bleu} with higher weighting for less frequent $n-$grams and a different length penalty \cite{Doddington2002}. & {\sc mt}\\
& {\sc rouge} & Recall-oriented score, with options for comparing non-contiguous $n-$grams and longest common subsequences \cite{Lin2003}. & {\sc as}\\
& {\sc meteor} & Harmonic mean of unigram precision and recall, with options for handling (near-synonymy) and stemming \cite{Lavie2007}. & {\sc mt}\\
& {\sc gtm} & General Text Matcher. F-Score based on precision and recall, with greater weight for contiguous matching spans \cite{Turian2003} &  {\sc mt} \\
& {\sc cide}r & Cosine-based n-gram similarity score, with n-gram weighting using {\sc tf-idf}  \cite{Vedantam2015}. & {\sc ic} \\		
& {\sc wmd} & Word-Mover Distance, a similarity score between texts, based on the (semantic) distance between words in the texts \cite{Kusner2015}. For {\sc nlp}, distance is operationalised using normalised bag of words ({\sc nbow}) representations \cite{Mikolov2013}. & {\sc ds}; {\sc ic} \\
\hline

\multirow{3}{*}[-20pt]{\rotatebox[origin=c]{90}{String distance}} &  Edit distance & Number of insertions, deletions, substitutions and, possibly, transposition required to transform the candidate into the reference string \cite{Levenshtein1966}. & {\sc n/a}\\
& {\sc ter} & Translation edit rate, a version of edit distance \cite{Snover2006}. & {\sc mt} \\
& {\sc terp} & Version of {\sc ter} handling phrasal substitution, stemming and synonymy \cite{Snover2006}. & {\sc mt} \\
& {\sc terpa} & Version of {\sc ter} optimised for correlations with adequacy judgements \cite{Snover2006}. & {\sc mt} \\
\hline
\multirow{4}{*}[-30pt]{\rotatebox[origin=c]{90}{Content overlap}} & Dice/Jaccard & Set-theoretic measures of overlap between two unordered sets (e.g. of predicates or other content units) & {\sc n/a} \\
& {\sc masi} & Measure of agreement between set-valued items, a weighted version of Jaccard \cite{Passonneau2006} & {\sc as} \\
& {\sc pyramid} & Overlap measure relying on comparison of weighted Summarization Content Units (SCUs) \cite{nenkova2004,yang2016} & {\sc as} \\
& {\sc spice} & Measure of overlap between candidate and reference texts based on propositional content obtained by parsing the text into graphs representing objects and relations, by first parsing captions into scene graphs representing objects and relations \cite{Anderson2016} & {\sc ic}\\
\hline
\end{tabular}
\caption{Intrinsic, corpus-based metrics based on string overlap, string distance, or content overlap. The last column indicates the {\sc nlp} sub-discipline in which a metric originated, where applicable. Legend: {\sc mt} = Machine translation; {\sc as} = automatic summarisation; {\sc ic} = image captioning; {\sc ds} = document similarity.}
\label{table:intrinsic-metrics}
\end{table}
\clearpage
}

\subsubsection{Objective Humanlikeness Measures Using Corpora}
Intrinsic methods that rely on corpora can generally be said to be addressing the question of `humanlikeness', that is, the extent to which the system's output matches human output under comparable conditions. From the developer's perspective, the selling point of such methods is their cheapness, since they are usually based on automatically computed metrics. A variety of corpus-based metrics, often used earlier in related fields such as Machine Translation or Summarisation, have been used in {\sc nlg} evaluation.
 Some of the main ones are summarised in Table \ref{table:intrinsic-metrics}, which groups them according to their principal characteristics, and for each adds a key reference.

Measures of n-gram overlap or string edit distance, usually originating in Machine Translation or Summarisation \cite<with some exceptions, such as {\sc cide}r,>{Vedantam2015} are frequently used for evaluating surface realisation \cite<e.g.,>{White2007,Cahill2006,Espinosa2010,Belz2011} and occasionally also to evaluate short texts characteristic of data-driven systems in domains such as weather reporting \cite<e.g.>{Reiter2009a,Konstas2013} and image captioning \cite<see>{Bernardi2016,Kilickaya2017}. Edit distance metrics have been exploited for realisation \cite{Espinosa2010}, but also for {\sc reg}  \cite{Gatt2010}.

The focus of these metrics is on the output text, rather than its fidelity to the input. In a limited number of cases, surface-oriented metrics have been used to evaluate the adequacy with which output text reflects content \cite{Banik2013,Reiter2009a}. However, if content determination is the focus, a measure of surface overlap is at best a proxy, relying on an assumption of a straightforward correspondence between input and output. This assumption may be tenable if texts are brief and relatively predictable. In some cases, it has been possible to use metrics to measure content determination directly, based on semantically annotated corpora. For instance, {\sc reg} algorithms have been evaluated in this fashion using set overlap metrics \cite{Viethen2007,Deemter2012}. Also relevant in this connection is the {\sc pyramid} method \cite{nenkova2004} for summarisation, which relies on the identification of the content units (which maximally correspond to clauses) in multiple human summaries. These are weighted and ordered by their frequency of mention by human summarises. A candidate summary is scored according to the ratio between the weight of the content units it includes, compared to the weight of an ideal summary bearing the same number of content units \cite<see>[for discussion]{Nenkova2011}.

Direct measurements of content overlap between generated and candidate outputs will likely increase, as automatic data-text alignment techniques make such `semantically transparent' corpora more readily available for end-to-end {\sc nlg} \cite<see e.g.,>[and the discussion in Section \ref{sec:datadriven}]{Chen2008,Liang2009}. An important development away from pure surface overlap is the use of semantic resources \cite<as in the case of {\sc meteor}, >{Lavie2007}, or word embeddings \cite<as in {\sc wmd, }>{Kusner2015}, to compute the proximity of output to reference texts beyond literal string overlap. In a comparative evaluation of metrics for image captioning, \citeA{Kilickaya2017} found an advantage for {\sc wmd} compared to other metrics.

\subsubsection{Evaluating Genre Compatibility and Stylistic Effectiveness}\label{sec:genre-eval}
A slightly different question that has occasionally been posed in evaluation studies asks whether the linguistic artefact produced by a system is a recognisable instance of a particular genre or style. As noted in Section \ref{sec:style}, it is difficult to ascertain to what extent readers actually perceive subtle stylistic variation. Thus, \citeA{Mairesse2011} found inconsistent perceptions of personality in the evaluation of {\sc personage}, which was complicated by the fact that stylistic features interact and may cancel each other out. 

Genre perception is a central question for approaches to generating creative language (see Section \ref{sec:creativity}). For example, \citeA{Hardcastle2008} describe an evaluation of a generation system for cryptic crossword clues based on a Turing test in which the objective was to determine whether the system's outputs were recognisably different from human-authored clues. In a related vein, when evaluating the {\sc jape} joke generation system, \citeA[see Section \ref{sec:puns}]{Binsted1997} presented 120 8-11 year old children with a number of punning riddles, some automatically generated by {\sc jape} and some selected from joke books. They also included a number of non-joke controls, such as:

\begin{example}
What do you get when you cross a horse and a donkey? \\
{\it A mule}
\end{example}

For each stimulus that they were exposed to, children were asked to indicate whether they thought it was a joke, and how funny they considered it. The results revealed that computer generated riddles were recognised as jokes, and considered funnier than non-jokes. Interestingly, the joke children rated highest was automatically generated by {\sc jape} (we urge the reader to inspect the original paper), although in general, human-produced jokes were considered funnier by children than automatically generated ones. In this evaluation study, therefore, an extrinsic aspect of the generated text, concerning its efficacy (here, its `funniness') was found to be correlated with its recognisability as an instance of the target genre. 

\citeA{petrovic2013unsupervised} evaluated their unsupervised approach to joke generation by harvesting human-written jokes from Twitter, conforming to the {\em I like my X \textellipsis} template used by their system. Blind ratings by human judges of human-written and automatically generated jokes showed that their best-performing model was rated as funny in 16\% of cases, compared to 33\% of the human jokes (itself a relatively low rate).

While the questions posed in these studies clearly have an intrinsic orientation (`Is the text compatible with the expected genre conventions?'), they also have a bearing on extrinsic factors, since the ability to recognise an artefact as an instance of a genre or as exhibiting a certain style or personality is arguably one of the sources of its impact, which in turn includes judgments of whether a text is funny or interesting, for example.
 
Of course, the intention behind variation in style, personality or affect may well be to ultimately increase effectiveness in achieving some ulterior goal. Indeed, any {\sc nlg} system intended to be embedded in a specific environment will need to address stylistic and genre-based issues. For example, our hypothetical weather report generator might use a very brief, technical style given its professional pool of target users \cite<as was the case with S{\sc um}T{\sc ime}>{Reiter2005}; in contrast, weather reports intended for public consumption, such as those in the W{\sc eather}G{\sc ov} corpus,  would probably be longer and less technical \cite{Angeli2010}. 

However, there is a difference between evaluating whether genre constraints or stylistic variation help contribute to a goal, and evaluating whether the text actually exhibits the desired variation. For example, \citeA{Mairesse2011} evaluated the {\sc personage} system (see Section \ref{sec:style}) by asking users to judge personality traits as reflected in generated dialogue fragments (rather than, say, measuring whether users were more likely to eat at a restaurant if this was recommended by a configuration of the system with a high degree of extraversion). This is similar in spirit to the question about jokehood asked by \citeA{Binsted1997}, in contrast to the more explicitly extrinsic evaluation of the {\sc standup} joke generator by \citeA{Waller2009}, which asked whether the system actually helped users improve their interactions with peers.

\subsection{Extrinsic Evaluation Methods}\label{sec:extrinsic}
In contrast to intrinsic methods, extrinsic evaluations measure effectiveness in achieving a desired goal. In the example scenario of Figure \ref{fig:eval-scenario}, such an evaluation might address the impact on planning by the engineers who are the target users of the system. Clearly, `effectiveness' is dependent on the application domain and purpose of a system. Examples include:

\begin{itemize}
\item persuasion and behaviour change, for example, through exposure to personalised smoking cessation letters \cite{Reiter2003};
\item purchasing decision after presentation of arguments for and against options on the housing market based on a user model \cite{Carenini2006};
\item engagement with ecological issues after reading blogs about migrating birds \cite{Siddharthan2012};
\item decision support in a medical setting following the generation of patient reports \cite{Portet2009,Hunter2012};
\item enhancing linguistic interaction among users with complex communication needs via the generation of personal narratives \cite{Tintarev2016};
\item enhancing learning efficacy in tutorial dialogue \cite{Dieugenio2005,Fossati2015,Boyer2011,Lipschultz2011,Chi2014}
\end{itemize}

While questionnaire-based or self-report studies can be used to address extrinsic criteria \cite<e.g.,>{Hunter2012,Siddharthan2012,Carenini2006}, in many cases evaluation relies on some objective measure of performance or achievement. This can be done with the target users {\sc in situ}, enhancing the ecological validity of the study, but can also take the form of a task that models the scenarios for which the {\sc nlg} system has been designed. Thus, in the {\sc give} Challenge \cite{Striegnitz2011}, in which {\sc nlg} systems generated instructions for a user to navigate through a virtual world, a large-scale task-based evaluation was carried out by having users play the {\sc give} game online, while various indices of success were logged, including the time it took a user to complete the game.  {\sc reg} algorithms whose goal was to generate identifying descriptions of objects in visual domains, were evaluated in part based on the time it took readers to identify a referent based on a generated description, as well as their error rate \cite{Gatt2010}. {\sc skillsum}, a system to generate feedback reports from literacy assessments, was evaluated by measuring how user's self-assessment of their own literacy skills improved after reading generated feedback, compared to control texts \cite{Williams2008}. 

A potential drawback of extrinsic studies, in addition to time and expense, is a reliance on an adequate user base (which can be difficult to obtain when users have to be sampled from a specific population, such as the engineers in our hypothetical scenario in Figure \ref{fig:eval-scenario}) and the possibility of carrying out the study in a realistic setting. Such studies also raise significant design challenges, due to the need to control for intervening and confounding variables, comparing multiple versions of a system (e.g. in an ablative design; see Section \ref{sec:glassbox} below), or comparing a system against a gold standard or baseline. For example, \citeA{Carenini2006} note that evaluating the effectiveness of arguments presented in text needs to take into account aspects of a user's personality which may impact how receptive they are to arguments in the first place.

An example of the trade-off between design and control issues and ecological validity is provided by the BabyTalk family of systems. A pilot system called {\sc bt-45} \cite{Portet2009}, which generated patient summaries from 45-minute spans of historical patient data, was evaluated in a task involving nurses and doctors, who chose from among a set of clinical actions to take based on the information given. These were then compared to `ground truth' decisions by senior neonatal experts. This evaluation was carried out off-ward; hence, subjects took clinical decisions in an artificial environment without direct access to the patient. On the other hand, in the evaluation of {\sc bt-nurse}, a successor to {\sc bt-45} which summarised patient data collected over a twelve-hour shift \cite{Hunter2012}, the system was evaluated on-ward using live patient data, but ethical considerations precluded a task-based evaluation. For the same reasons, comparison to `gold standard' human texts was also impossible. Hence, the evaluation elicited judgements, both on intrinsic criteria such as understandability and accuracy and on extrinsic criteria such as perceived clinical utility \cite<see>[for a similarly indirect extrinsic measure of impact, this time in an ecological setting]{Siddharthan2012}.

\subsection{Black Box Versus Glass Box Evaluation}\label{sec:glassbox}
With the exception of evaluations of specific modules or algorithms, as in the case of {\sc reg} or surface realisers, most of the evaluation studies discussed so far would be classified as `black box' evaluations of `end-to-end', or complete, {\sc nlg} systems. In a `glass box' evaluation, on the other hand, it is the contribution of individual components that is under scrutiny, ideally in a setup where versions of a system with and without a component are evaluated in the same manner. Note that the distinction between black box and glass box evaluation is orthogonal to the question of which methods are used. 

An excellent example of a glass-box evaluation is that by \citeA{Callaway2002}, who used an ablative design, eliciting judgements of the quality of the output of their narrative generation system based on different configurations that omitted or included key components. In a related vein, \citeA{Elliott2013} compared image-to-text models that included fine-grained dependency representations of spatial as well as linguistic dependencies, to models with a coarser-grained image representation, finding an advantage for the former.

However, exhaustive component-wise comparisons are sometimes difficult to make and may result in a combinatorial explosion of configurations, with a concomitant reduction in data points collected per configuration (assuming subjects are limited and need to be divided among different conditions) and a reduction in statistical power. Alternatives do exist in the literature. \citeA{Reiter2003} elicited judgements on weather forecasts using human and machine-generated texts, together with a `hybrid' version where the content was selected by forecasters, but the language was automatically generated. This enabled a comparison of human and automatic content selection. \citeA{Angeli2010} used corpus-based and subjective measures to assess linguistic quality, coupled with precision and recall-based measures to assess content determination of their statistical system against human-annotated texts. In {\sc bt-nurse} \cite{Hunter2012}, nurses were prompted for free text comments (in addition to answering a questionnaire targeting extrinsic dimensions), which were then manually annotated and analysed to determine which elements of the system were potentially problematic.

\subsection{On the Relationship Between Evaluation Methods}
To what extent are the plethora of methods surveyed -- from extrinsic, task-oriented to intrinsic ones relying on automatic metrics or human judgements -- actually related? It turns out that multiple evaluation methods seldom give converging verdicts on a system, or on the relative ranking of a set of systems under comparison. 

\subsubsection{Metrics Versus Human Judgements}
Although corpus-based metrics used in {\sc mt} and summarisation are typically validated by demonstrating their correlation with human ratings, meta-evaluation studies in these fields have suggested that the correspondence is somewhat weak \cite<e.g.,>{Dorr2004,Callison-Burch2006,Caporaso2008}. Similarly, shared task evaluations on referring expression generation showed that corpus-based, judgement-based and experimental or task-based methods frequently do not correlate \cite{Gatt2010}. In their recent review \citeA{Bernardi2016} note a similar issue in image captioning system evaluation. Thus, \citeA{Kulkarni2013} found that their image description system did not outperform two earlier methods \cite{Farhadi2010,Yang2011} on {\sc bleu} scores; however, human judgements indicated the opposite trend, with readers preferring their system \cite<similar observations are made by>{Kiros2014}. \citeA{Hodosh2013} compared the agreement (measured by Cohen's $\kappa$) between human judgements and {\sc bleu} or {\sc rouge} scores for retrieved captions, finding that outputs were not ranked similarly by humans and metrics, unless the retrieved captions were identical to the reference captions. 

On occasion, the correlation between a metric and human judgements appears to differ across studies, suggesting that metric-based results are highly susceptible to variation due to generation algorithms and datasets. For instance, \citeA{Konstas2013} (discussed in Section \ref{sec:parsing} above) find that on corpus-based metrics, the best-performing version of their model does not outperform that of \citeA{Kim2010} on the {\sc robocup} domain, or that of \citeA{Angeli2010} on their weather corpus ({\sc weathergov}), though it performs better than \citeauthor{Angeli2010}'s on the noisier {\sc atis} travel dataset. However, an evaluation of fluency and semantic correctness, based on human judgements, showed that the system outperformed, by a small margin, both \citeauthor{Kim2010}'s and \citeauthor{Angeli2010}'s on both measures in all domains with the exception of {\sc weathergov}, where \citeauthor{Angeli2010}'s system did marginally better.

In a related vein, \citeA{Elliott2015} compare their image captioning system, based on visual dependency relations, to the Bidirectional {\sc rnn} developed by \citeA{Karpathy2015}, on two different datasets. The two systems were close to each other on the {\sc vlt2k} dataset, but not on Pascal1k, a result that the authors claim is due to {\sc vlt2k} containing more pictures involving actions. As for the relationship between metrics and human judgements, \citeA{Elliott2013} concluded that {\sc meteor} correlates better than {\sc bleu} \cite<see>[for a systematic comparison of automatic metrics in this domain]{Elliott2014}, a finding also confirmed in their later work \cite{Elliott2015}, as well as in the {\sc ms-coco} Evaluation Challenge, which found that {\sc meteor} was more robust. However, work by \citeA{Kuznetsova2014a} showed variable results; their highest-scoring method as judged by humans, involving tree composition, was ranked higher by {\sc bleu} than by {\sc meteor}. In the {\sc ms-coco} Evaluation Challenge, some systems outperformed a human-human upper bound when compared to reference texts using automatic metrics, but no system reached this level in an evaluation based on human judgements \cite<see>[for further discussion]{Bernardi2016}.

Some studies have explicitly addressed the relationship between methods as a research question in its own right. An important contribution in this direction is the study by \citeA{Reiter2009a}, which addressed the validity of corpus-based metrics in relation to human judgements, within the domain of weather forecast generation \cite<a similar study has recently been conducted on image captioning; see>{Elliott2014}. In a first experiment, focussing on linguistic quality, the authors found a high correlation between expert and non-expert readers' judgements, but the correlation between human judgements and the automatic metrics varied considerably (from $0.3$ to $0.87$), depending on the version of the metric used and whether the reference texts were included in the comparison by human judges. The second experiment evaluated both linguistic quality, by asking human judges to rate clarity/readability; and content determination, by eliciting judgements of accuracy/appropriateness (by comparing texts to the raw data). The automatic metrics correlated significantly with judgements of clarity, but far less with accuracy, suggesting that they were better at predicting the linguistic quality than correctness. 

Other studies have yielded similarly inconsistent results. In a study on paraphrase generation, \citeA{Stent2005a} found that automatic metrics correlated highly with judgements of adequacy (roughly akin to accuracy), but not fluency. By contrast, \citeA{Espinosa2010} found that automatic metrics such as {\sc nist}, {\sc meteor} and {\sc gtm} correlate moderately well with human fluency and adequacy judgements of English surface realisation quality, while \citeA{Cahill2009} reported only a weak correlation for German surface realisation. \citeA{Wubben2012}, comparing text simplification strategies, found low, but significant correlations between {\sc bleu} and fluency judgements, and a very low, {\em negative} correlation between {\sc bleu} and adequacy. These contrasting findings suggest that the relationship between metrics may depend on  purpose and genre of the text under consideration; for example, \citeA{Reiter2009a} used weather reports, while \citeA{Wubben2012} used Wikipedia articles.

Various factors can be adduced to explain the inconsistency of these meta-evaluation studies:

\begin{enumerate}
\item Metrics such as {\sc bleu} are sensitive to the length of the texts under comparison. With shorter texts, n-gram based metrics are likely to result in lower scores.
\item The type of overlap matters: for example, many evaluations in image captioning rely on {\sc bleu-1} \cite[was among the first to experiment with longer n-grams]{Elliott2013,Elliott2014}, but longer n-grams are harder to match, though they capture more syntactic information and are arguably better indicators of fluency.
\item Semantic variability is an important issue. Generated texts may be similar to reference texts, but differ on some near-synonyms, or subtle word order variations. As shown in Table \ref{table:intrinsic-metrics}, some metrics are designed to partially address these issues.
\item Many intrinsic corpus-based metrics are designed to compare against multiple reference texts, but this is not always possible in {\sc nlg}. For example, while image captioning datasets typically contain multiple captions per image (typically, around 5), this is not the case in other domains, like weather reporting or restaurant recommendations.
\end{enumerate}
The upshot is that {\sc nlg} evaluations increasingly rely on multiple methods, a trend that is equally visible in other areas of {\sc nlp} , such as {\sc mt} \cite{Callison-Burch2007,Callison-Burch2008}.

\subsubsection{Using Controlled Experiments}
A few studies have validated evaluation measures against experimental data. For example, \citeA{Siddharthan2012a} compared the outcomes of their magnitude estimation judgement study (see Section \ref{sec:intrinsic} above) to the results from a sentence recall task, finding that the results from the latter are largely consistent with judgements and concluding that they can substitute for task-based evaluations to shed light on breakdowns in comprehension at sentence level. A handful of studies have also used behavioral experiments and compared `online' processing measures, such as reading time of referring expressions, to corpus-based metrics \cite<e.g.>{Belz2010}. Correlations with automatic metrics are usually poor. A somewhat different use of reading times was made by \citeA{Lapata2006}, who used them as an objective measure against which to validate Kendall's $\tau$ as a metric for assessing information ordering in text (an aspect of text stucturing).  In a recent study, \citeA{Zarriess2015} compared generated texts to human-authored and `filler' texts (which were manually manipulated to compromise their coherence). They found that reading-time measures were more useful to distinguish these classes of texts than offline measures based on elicited judgements of fluency and clarity.

\subsection{Evaluation: Concluding Remarks}
Against the background of this section, three main conclusions can be drawn:

\begin{enumerate}
\item There is a widespread acceptance of the necessity of using multiple evaluation methods in {\sc nlg}. While these are not always consistent among themselves, they are useful in shedding light on different aspects of quality, from fluency and clarity of output, to adequacy of semantic content and effectiveness in achieving communicative intentions. The choice of method has a direct impact on the way in which results can be interpreted.

\item Meta-evaluation studies have yielded conflicting results on the relationship between human judgements, behavioural measures and automatically computed metrics. The correlation among them varies depending on task and application domain. This is a subject of ongoing research, with plenty of studies focussing on the reliabilty of metrics and their relationship to other measures, especially human judgements. 

\item A question that remains under-explored concerns the dimensions of quality that are themselves the object of inquiry. (In this connection, it is worth noting that some kindred disciplines have sought to de-emphasise their role on the grounds that they are inconsistent; see \citeauthor{Callison-Burch2008}, 2008, among others). For example, what are people judging when they judge fluency or adequacy and how consistently do they do so? It is far from obvious whether these judgements should really be expected to correlate with other measures, given that the latter are {\em producer-}oriented, focussing on output, while judgements are themselves often {\em receiver-}oriented, focussing on how the output is read or processed \cite<for a related argument, see>{Oberlander1998}.  Furthermore, while meta-linguistic judgements can be expected to reflect the impact of a text on its readers, there is nevertheless the possibility that behavioural, online methods designed to directly investigate aspects of processing would yield a different picture, a result that has been obtained in some psycholinguistic studies \cite<e.g.>{Engelhardt2006}.
\end{enumerate}

In conclusion, our principal recommendation to {\sc nlg} practitioners, where evaluation is concerned, is to err in favour of diversity, by using multiple methods, as far as possible, and reporting not only their results, but also the correlation between them. Weak correlations need not imply that the results of a particular method are invalid. Rather, they may indicate that measures focus on different aspects of a system or its output. 

\section{Discussion and Future Directions}\label{sec:discussion}

Over the past two decades, the field of {\sc nlg} has advanced considerably, and many of these recent advances have not been covered in a comprehensive survey yet. This paper has sought to address this gap, with the following goals:

\begin{enumerate}
\item to give an update of the core tasks and architectures in the field, with an emphasis on recent data-driven techniques;
\item to briefly highlight recent developments in relatively new areas, incuding vision-to-text generation and the generation of stylistically varied, engaging or creative texts; and
\item to extensively discuss the problems and prospects of evaluating {\sc nlg} applications.
\end{enumerate}

Throughout this survey, various general, related themes have emerged. Probably the central theme has been the gradual shift away from traditional, rule-based approaches to statistical, data-driven ones, which, of course, has been taking place in {\sc ai} in general. In {\sc nlg}, this has had substantial impact on how individual tasks are approached (e.g., moving away from domain-dependent to more general, domain-independent approaches, relying on available data instead) as well as on how tasks are combined in different architectures (e.g., moving away from modular towards more integrated approaches).  The trade-off between output quality of the generated text and the efficiency and robustness of an approach is becoming a central issue: data-driven approaches are arguably more efficient than rule-based approaches, but the output quality may be compromised, for reasons we have discussed. Another important theme has been the increased interplay between core {\sc nlg} research and other disciplines, such as computer vision (in the case of vision-to-text) and computational creativity research (in the case of creative language use).

At the conclusion of this comprehensive survey of the state of the art in {\sc nlg}, and given the fast pace at which developments occur both in industry and academia, we feel it is useful to point to some potential future directions, as well as to raise a number of questions which recent research has brought to the fore.

\subsection{Why (and How) Should NLG be Used?}
Towards the beginning of their influential survey on {\sc nlg}, \citeA{Reiter2000} recommended to the developer that she pose this question before embarking on the design and implementation of a system. Can {\sc nlg} really help in the target domain? Does a cheaper, more standard solution exist and would it work just as well? From the perspective of an engineer or a company, these are obviously relevant questions. As recent industry-based applications of {\sc nlg} show, this technology is typically valuable whenever information that needs to be presented to users is relatively voluminous, and comes in a form which is not easily consumed and does not afford a straightforward mapping to a more user-friendly modality without considerable transformation. This is arguably where {\sc nlg} comes into its own, offering a battery of techniques to select, structure and present the information.

However, the question whether {\sc nlg} is worth using in a specific setting should also be accompanied by the question of {\em how} it should be used. Our survey has focussed on techniques for the generation of text, but text is not always presented in isolation. Other important dimensions include document structure and layout, an under-studied problem \cite<but see>{Power2003}. They also include the role of graphics in text, an area where there is the potential for further interaction between the {\sc nlg} and visualisation communities, addressing such questions as which information should be rendered textually and which can be made more accessible in a graphical modality \cite<e.g.,>{demir2012}. These questions are of great relevance in some domains, especially those where accurate information delivery is a precursor to decision-making in fault-critical situations \cite<for some examples, see>{Elting1999,Law2005,Meulen2007}.

\subsection{Does NLG Include Text-to-Text?}
In our introductory section, we distinguished text-to-text generation from data-to-text generation; this survey has focussed primarily on the latter. The two areas have distinguishing characteristics, not least the fact that {\sc nlg} inputs tend to vary widely, as do the goals of {\sc nlg} systems as a function of the domain under consideration. In contrast, the input in text-to-text generation, especially Automatic Summarisation, is comparatively homogeneous, and while its goals can vary widely, the field has also been successful at defining tasks and datasets (for instance, through the {\sc duc} shared tasks), which have set the standard for subsequent research.

Yet, a closer look at the two types of generation will show more scope for convergence than the above characterisation suggests. To begin with, if {\sc nlg} is concerned with going from data to text, then surely textual input should be considered as one out of broad variety of forms in which input data might be presented. Some recent work, such as that of \citeA{Kondadadi2013} (discussed in Section \ref{sec:datadriven}) and \citeA{Mcintyre2009} (discussed in Section \ref{sec:creativity}) has explicitly focussed on leveraging such data to generate coherent text. Other approaches to {\sc nlg}, including some systems that conform to a standard, modular, data-to-text architecture \cite<e.g.,>{Hunter2012}, have had to deal with text as one out of a variety of input types, albeit using very simple techniques. Generation from heterogeneous inputs which include text as one type of data is a promising research direction, especially in view of the large quantities of textual data available, often accompanied by numbers or images. 

\subsection{Theories and Models in Search of Applications?}
In their overview of the status of evaluation in {\sc nlg} in the late 1990s, \citeA{Mellish1998a} discussed, among the possible ways of evaluating a system, its theoretical underpinnings and in particular whether the theoretical model underlying an {\sc nlg} system or one of its components is adequate to the task and can generalise to new domains. Rather than evaluating an {\sc nlg} system as such, this question targets the theory itself, and suggests that we view {\sc nlg} as a potential testbed for such theories or models. But what are the theories that underlie {\sc nlg}?

The prominence of theoretical models in {\sc nlg} tends to depend on the task under consideration. For instance, many approaches to realisation discussed in Section \ref{sec:lr} are based on a specific theory of syntactic structure; research on {\sc reg} has often been based on insights from pragmatic theory, especially the Gricean maxims \cite{Grice1975}; and much research on text structuring has been inspired by Rhetorical Structure Theory \cite{Mann1988}. Relatively novel takes on various sentence planning tasks -- especially those concerned with style, affect and personality -- tend to have a theoretical inspiration, in the form of a model of personality \cite{John1999} or a theory of politenes \cite{BrownLevinson1987}, for example.

More often than not, such theories are leveraged in the process of formalising a particular problem to achieve a tractable solution. Treating their implementation in an {\sc nlg} system as an explicit test of the theory, as \citeA{Mellish1998a} seem to suggest, happens far less often. This is perhaps a reflection of a division between `engineering-oriented' and `theoretically-oriented' perspectives in the field: the former perspective emphasises workable solutions, robustness and output quality; the latter emphasises theoretical soundness, cognitive plausibility and so forth. However, the theory/engineering dichotomy is arguably a false one. While the goal of {\sc nlg} research is often different from, say, that of cognitive modelling (for example, few {\sc nlg} systems seek to model production errors explicitly), it is also true that theory-driven implementations are themselves worthy contributions to theoretical work. 

Recently, some authors have argued that {\sc nlg} practitioners should pay closer attention to theoretical and cognitive models. The reasons marshalled in favour of this argument are twofold. First, psycholinguistic results and theoretical models can actually help to improve implemented systems, as \citeA{Rajkumar2014} show for the case of realisation. Second, as argued for example by \citeA{VanDeemter2012a}, theoretical models can benefit from the formal precision that is the bread-and-butter of computational linguistic research; a concrete case in point in {\sc nlp} is provided by \citeA{Poesio2004}, whose implementation of Centering Theory \cite{Grosz1995} shed light on a number of underspecified parameters in the original model and subsequent modifications of it. Our argument here is that {\sc nlg} has provided a wealth of theoretical insights which should not be lost to the broader research community; similarly, {\sc nlg} researchers would undoubtedly benefit from an awareness of recent developments in theoretical and experimental work.

\subsection{Where do We Go from Here?}
Finally, we conclude with some speculations on some further directions for future research for which the time seems ripe.

Within the field of Natural Language Processing as a whole, a remarkable recent developments is the explosion of interest in social media, including online blogs, micro-blogs such as Twitter feeds, and social platforms such as Facebook. In one respect, interest in social media could be seen as a natural extension of long-standing topics in {\sc nlp}, including the desire to deal with language `in the wild'. However, social media data has given more impetus to the exploration of non-canonical language \cite<e.g.>{Eisenstein2013}; the impact of social and demographic factors on language use \cite<e.g.>{Hovy2015,Johannsen2015}; the prevalence of paralinguistic features such as affect, irony and humour \cite{Pang2008,Lukin2013}; and other variables such as personality \cite<e.g.>{Oberlander2006,Farnadi2013,Schwartz2013a}. Social media feeds are also important data streams for the identification of topical and trending events  \cite<see>[for a recent review]{Atefeh2015}. There is as yet little work on generating textual or multimedia summaries of such data \cite<but see, for example,>{Wang2014} or generating text in social media contexts \cite<exceptions include>{Ritter2011,Cagan2014}. Since much of social media text is subjective and opinionated, an increased interest in social media on the part of {\sc nlg} researchers may also give new impetus to research on the impact of style, personality and affect on textual variation (discussed in Section \ref{sec:style}), and on non-literal language (including some of the phenomena discussed in Section \ref{sec:creativity}).

A second potential growth area for {\sc nlg} is situated language generation. The term {\em situated} is usually taken to refer to language use in physical or virtual environments where production choices explicitly take into account perceptual and physical properties. Research on situated language processing has advanced significantly in the past several years, with frameworks for language production and understanding in virtual contexts \cite<e.g.,>{Kelleher2005}, as well as a number of contributions within {\sc nlg}, especially for the generation of language in interactive environments \cite{Kelleher2006,Stoia2006,Garoufi2013,Dethlefs2015}. The popular {\sc give} Challenge added further impetus to this research \cite{Striegnitz2011}. Clearly, this work is also linked to the enterprise of grounding generated language in the perceptual world, of which the research discussed in Section \ref{sec:image} constitutes one of the current trends. However, there are many fields where situatedness is key, in which {\sc nlg} can still make novel contributions. One of these is gaming. With the exception of a few endeavours to enhance the variety of linguistic expressions used in virtual environments \cite<e.g.,>{Orkin2007}, {\sc nlg} technology is relatively unrepresented in research on games, despite significant progress on dynamic content generation in game environments \cite<e.g.,>{Togelius2011}.
This may be due to the perception that linguistic interaction in games is predictable and can rely on `canned' text.  However, with the growing influence of gamification as a strategy for enhancing a variety of activities beyond entertainment, such as pedagogy, as well as the development of sophisticated planning techniques for varying the way in which game worlds unfold on the fly, the assumption of predictability where language use is concerned may well be up for revision. 

Third, there is a growing interest in applying {\sc nlg} techniques to generation from structured knowledge bases and ontologies \cite<e.g.>[some of which were briefly discussed in Section \ref{sec:parsing}]{Ell2012,Duma2013,Gyawali2014,Mrabet2016,Sleimi2016}. The availability of knowledge bases such as {\sc dbp}edia, or folksonomies such as Freebase, not only constitute input sources in their own right, but also open up the possibility of exploring alignments between structured inputs and text in a broader variety of domains than has hitherto been the case.

Finally, while there has been a significant shift in the past few years towards data-driven techniques in {\sc nlg}, many of these have not been tested in commercial or real-world applications, despite the growth in commercialisation of text generation services noted in the introductory section. Typically, the arguments for rule-based systems in commercial scenarios, or in cases where input is high-volume and heterogeneous, are that (1) their output is easier to control for target systems; or (2) that data is in any case unavailable in a given domain, rendering the use of statistical techniques moot; or (3) data-driven systems have not been shown to be able to scale up beyond experimental scenarios \cite<some of these arguments are made, for instance, by>{Harris2008}. A response to the first point depends on the availability of techniques which enable the developer to `look under the hood' and understand the statistical relationships learned by a model. Such techniques are, for example, being developed to investigate or visualise the representations learned by deep neural networks. The second point calls for more investment in research on data acquisition and data-text alignment. Techniques for generation which rely on less precise alignments between data and text are also a promising future direction. Finally, scalability remains an open challenge. Many of the systems we have discussed have been developed within research environments, where the aim is of course to push the frontiers of {\sc nlg}  and demonstrate feasibility or correctness of novel approaches. While in some cases, research on data-to-text has addressed large-scale problems -- notably in some of the systems that summarise numerical data -- a greater concern with scalability would also focus researchers' attention on issues such as the time and resources required to collect data and train a system and the efficiency of the algorithms being deployed. Clearly, developments in hardware will alleviate these problems, as has happened with some statistical methods that have recently become  more feasible.

\section{Conclusion} 
Recent years have seen a marked increase in interest in automatic text generation. Companies now offer {\sc nlg} technology for a range of applications in domains such as journalism, weather, and finance. The huge increase in available data and computing power, as well as rapid developments in machine-learning, have created many new possibilities and motivated {\sc nlg} researchers to explore a number of new applications, related to, for instance, image-to-text generation, while applications related to social media seem to be just around the corner, as witness, for instance, the emergence of {\sc nlg}-related techniques for automatic content-creation as well as {\sc nlg} for twitter and chatbots \cite<e.g.,>{Dale2016}. With developments occurring at a steady pace, and the technology also finding its way into industrial applications, the future of the field seems bright. In our view, research in {\sc nlg} should be further strengthened by more collaboration with kindred disciplines. It is our hope that this survey will serve to highlight some of the potential avenues for such multi-disciplinary work.

\section*{Acknowledgements}
We thank the four reviewers for their detailed and constructive comments. In addition, we have greatly benefitted from discussions with and comments from Grzegorz Chrupala, Robert Dale, Raquel Herv\'as, Thiago Castro Ferreira, Ehud Reiter, Marc Tanti, Mari\"et Theune, Kees van Deemter, Michael White and Sander Wubben. EK received support from RAAK-PRO SIA (2014-01-51PRO) and The Netherlands Organization for Scientific Research (NWO 360-89-050), which is gratefully acknowledged.

\vskip 0.2in
\bibliographystyle{theapa}
\bibliography{jair}
\end{document}